%% file: main.tex
\documentclass[english, 12pt, a4paper, sci, utf8, a-2b, online]{aaltothesis}

\setupthesisfonts

\usepackage{graphicx}
\usepackage{longtable}
\usepackage[type={CC}, modifier={by-nc-sa}, version={4.0}]{doclicense}
\usepackage{overpic}
\usepackage{color}
\usepackage{multirow}
\usepackage{colortbl}
\usepackage{rotating}
\usepackage{hyperref}

\def\ie{\emph{i.e.}}
\def\eg{\emph{e.g.}}
\def\etc{\emph{etc}}
\def\etal{{\em et al.~}}

\newcommand{\figref}[1]{Figure~\ref{#1}}
\newcommand{\tabref}[1]{Table~\ref{#1}}
\newcommand{\eqnref}[1]{(Equation~\ref{#1})}
\newcommand{\secref}[1]{Section~\ref{#1}}

\makeatletter
\newcommand{\thickhline}{%
    \noalign {\ifnum 0=`}\fi \hrule height 1pt
    \futurelet \reserved@a \@xhline
}

\def\ourmodel{\textit{HICOME}}
\def\ourdataset{\textit{CoSINe}}  
\def\NumDLCoSODComp{12}
\definecolor{bblue}{rgb}{0,150,230}
\definecolor{mygray}{gray}{.92}

\newcommand{\tbd}[1]{{\textcolor{blue}{#1}}}

\input{cover}

\begin{document}
\makecoverpage
\makecopyrightpage
\clearpage

\input{head}    

\section{Introduction}
\label{sec:introduction}
\begin{figure*}[t!]
    \centering
    \small{}
    \begin{overpic}[width=.98\textwidth]{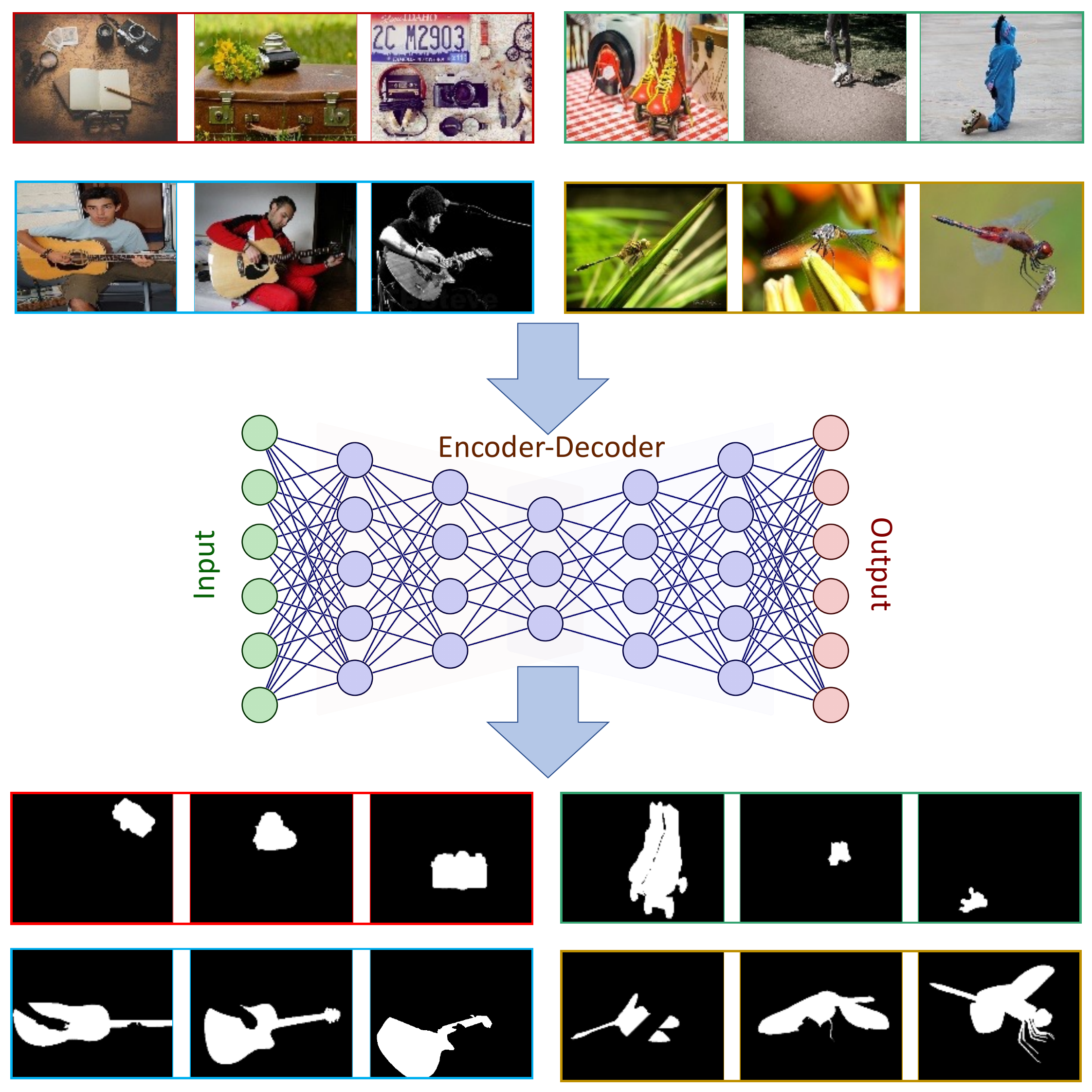}
    \end{overpic}
    \caption{\textit{The pipeline of CoSOD task.} The inputs (image groups) are fed into the network (CoSOD models), which generates the output (co-saliency maps).}
    \label{fig:cosod_procedure}
\end{figure*}

\begin{figure*}[t!]
    \centering
    \small{}
    \begin{overpic}[width=.98\textwidth]{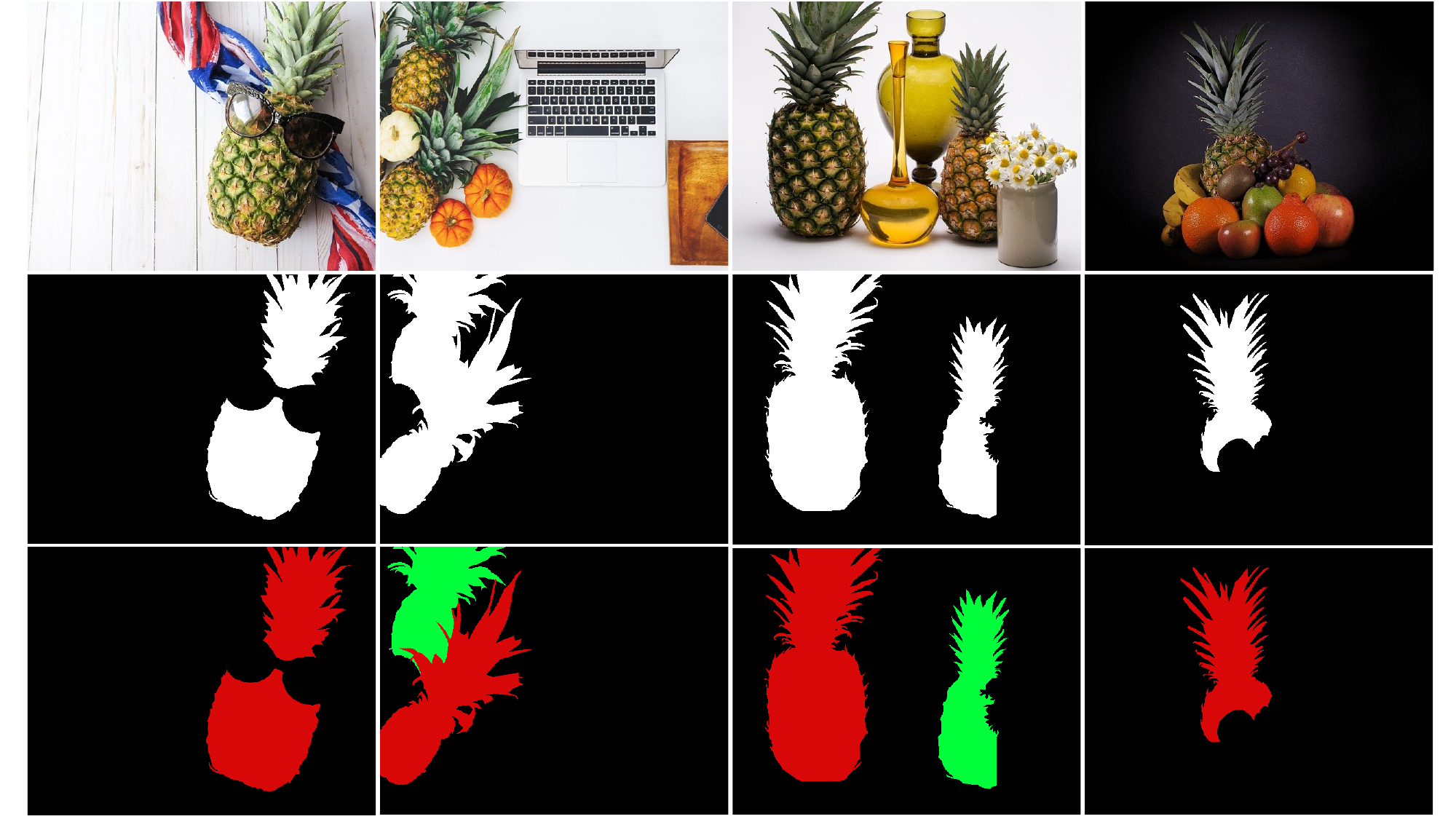}
        \put(-1,43){\rotatebox{90}{Source}}
        \put(-1,23){\rotatebox{90}{Binary}}
        \put(-1,3){\rotatebox{90}{Instance-level}}
    \end{overpic}
    \caption{\textit{Samples of pineapple group in CoCA dataset.} The images, binary ground-truth maps, and instance-level ground-truth maps.}
    \label{fig:coca_samples}
\end{figure*}

Co-salient object detection (CoSOD) aims at detecting the most common and salient objects among a given group of images, as illustrated in~\figref{fig:cosod_procedure}. Given a group of images as the input, the CoSOD models generate the masks of the most common objects among the salient ones (as shown in~\figref{fig:cosod_procedure}). It has arisen a wide range of interest in the field of Computer Vision (CV). Compared with Salient Object Detection (SOD), which is designed for finding the most attention-grabbing objects in a single image, CoSOD contains multiple goals and is more challenging to distinguish co-occurring distractors (salient objects with non-target classes) and to detect target objects in a set of images (as examples in the CoSOD test set shown in~\figref{fig:coca_samples}). Due to the existence of distractors, \textbf{intra-class consistency} and \textbf{inter-class separation} are both vital factors in terms of learning a qualified co-salient object detector~\cite{GCoNet,GCoNet+}. With the increase of deep-learning techniques, deep CoSOD models~\cite{CoSOD3k,GCoNet,GCoNet+,MCCL} have surpassed traditional CoSOD methods~\cite{CBCS,CoSOD_trad_align} and show their great performance on different benchmarks~\cite{CoSal2015,CoSOD3k,GICD}. 

Because of its potential and increasing accuracy, CoSOD has arisen a wide range of attention in both academic research and practical applications. CoSOD can be utilized as a pre-processing step for other CV tasks~\cite{hsu2019deepco3,wang2019no} to extract the masks of co-occurring objects, such as detecting foreground objects in videos to help video object segmentation and generating affinity maps for object co-localization. In addition, it can be also applied in many related practical applications~\cite{GCoNet,CoSal2015}, including query-based object retrieval and content-aware object co-segmentation.

\subsection{Motivation}
\label{sec:motivation_solution}
\textbf{Dataset.} Hundreds of works on CoSOD have been published in the last few years\footnote{A collection of representative methods and datasets before 2021 can be found at https://github.com/DengPingFan/CoSOD3k~\cite{CoSOD3k}.}. To establish a fair and open evaluation of these methods, many CoSOD test sets have been proposed, \eg{}, CoSal2015~\cite{CoSal2015}, CoSOD3k~\cite{CoSOD3k}, and CoCA~\cite{GICD}. However, there is still a weird problem unsolved in CoSOD,~\ie{}, the lack of a powerful and efficient dataset employed as the standard training dataset. Existing CoSOD datasets for training mainly suffer from the following problems:

\textit{Wrong ground-truth maps.} There are many wrong ground-truth maps in existing CoSOD training sets due to a lack of instance-level annotations to eliminate objects of wrong classes. For example, DUTS\_class~\cite{GICD} kept all the binary ground-truth maps from DUTS~\cite{DUTS}. 

\textit{Absence of saliency.} In COCO-SEG~\cite{COCO_SEG} and COCO-9k~\cite{GWD}, images and ground-truth maps are borrowed from the COCO dataset~\cite{COCO}, and then divided into groups by class. Apart from the above problem, the selection of objects does not take the saliency into account. In that case, many objects in the background that attract little human attention are also put into the ground-truth maps, which is a fault.

\textit{Small number of groups.} CoSOD is a class-agnostic task and can be applied in scenes where objects are never seen. Therefore, a larger number of groups can help the model to get overfitted in certain classes of too many training samples.

\textit{Balance of groups.} Similar to the above, a balanced data set can help the models trained on it avoid leaning to certain classes which occur too many times. For example, the \textit{person} group takes almost half of the images in the entire COCO-SEG dataset.

\bigskip{}
\noindent{}\textbf{Methodology.} Apart from the dataset, existing CoSOD methods also have not truly hit the key of the CoSOD task, though they have achieved a kind of satisfying performance on existing benchmarks. Existing works in the early years tried to extract the within-group common features~\cite{GWD,CoADNet,CADC}. In~\cite{GCoNet,MCCL}, consensus mining is extended from a single group into multiple groups with a Siamese network to achieve interaction between groups. All these efforts are proposed to better exploit the two vital cues in CoSOD,~\ie{}, intra-group compactness and inter-group separability.


\subsection{Research Questions}
\label{sec:research_questions}
As the problems mentioned above still exist and there is still a lot of room to grow, several concerns are presented here to gain a clearer understanding of the progress of current CoSOD research. 

\textbf{First}, since the existing CoSOD training sets suffer from these problems, how to deal with the lack of a standard, effective, and efficient training set to train CoSOD models in a better and faster way? If a new dataset is proposed, what advantages should it provide to outperform the previous ones?

\textbf{Second}, though previous works have made progress on making use of the intra-class consistency and inter-class separation to guide CoSOD models to learn more discriminative features, many proposed components bring a lot of extra computation and slow down the inference. On the other hand, surrounding distractors are also obstacles for CoSOD models to learn the features of the right objects. Hence, the question is how to eliminate the disturbing of distractors and learn the accurate and discriminative features without introducing extra computation to inference.

\textbf{Finally}, if most of the defects mentioned above have been greatly improved, what is the next potential way for CoSOD to go further?

\subsection{Contributions}
\label{sec:contributions}
To tackle the aforementioned problems and concerns, a new CoSOD dataset and the corresponding novel method are proposed in this thesis to jointly deal with the pain points of the CoSOD area today. The contributions of this work can be summarized as follows:

\begin{itemize}\setlength\itemsep{0.3em}
    \item[\textbf{1.}] 
    \textbf{A new CoSOD training set (\ourdataset{}) is introduced to overcome the inherent defects of existing ones.} The proposed ~\ourdataset{} is so far the largest CoSOD training set in terms of the number of groups. It has 22,978 images in 919 groups in total, where each image may contain multiple positive instances. A large number of groups is very suitable for the nature of CoSOD, avoiding models from finding common objects by remembering the features of certain classes. This dataset is thought to push forward CoSOD research not only at the performance level, but also at the design of CoSOD methods.

    \item[\textbf{2.}] 
    \textbf{A novel CoSOD model named Hierarchical Instance-aware CONsensus MinEr (\ourmodel{}) is proposed to make full use of the proposed~\ourdataset{} dataset and to get closer to the essence of CoSOD. Several tricks specially designed for CoSOD model training are also proposed and sorted out.}
    Three new components are proposed in this thesis which significantly improve the performance of the proposed~\ourmodel{},~\ie{}, Hierarchical Consensus Fusion (\textbf{HCF}), Spatial Increment Attention (\textbf{SIA}), and Instance-Aware Contrastive Consensus Learning (\textbf{IACCL}) to make good use of the characteristics of~\ourdataset{} and strengthen the ability to detect co-salient objects. 1) Hierarchical Consensus Fusion (HCF): To make the proposed model more robust and to be able to find the common objects of different sizes in a more accurate way, a hierarchical consensus fusion module is employed to capture the cross-scale consensus. 2) Spatial Increment Attention (SIA): a spatial increment module is added to the original multi-head attention (MHA) to enhance the key and value of it. SIA is placed at each single decoder block for a larger receptive field and feature enhancement. 3) Instance-Aware Contrastive Consensus Learning (IACCL): Contrastive learning is applied to the consensus of objects that are identified as targets to eliminate the distractors. IACCL brings a stronger ability to discriminate objects of different classes to the proposed~\ourmodel{}. These three components are organically combined in the proposed~\ourmodel{} and make it achieve great performance on all existing datasets in the experiments.


    \item[\textbf{3.}]
    \textbf{Comprehensive experiments are conducted on both the proposed~\ourdataset{} dataset and~\ourmodel{} method to show their superior performance.}\footnote{The benchmark will be publically available on \url{https://github.com/ZhengPeng7/CoSINe}.}
    Although the development of Co-Salient Object Detection is rapid, there are three commonly used datasets for training,~\ie{}, DUTS\_class~\cite{GICD}, COCO-9k~\cite{GWD}, and COCO-SEG~\cite{COCO_SEG}, while there is no standard for choosing the training sets for this task. Due to existing works where the used training sets are not the same, more comprehensive experiments are conducted with all combinations of these three training sets for fair experimental comparisons. More experiments are also conducted with the proposed~\ourdataset{} dataset as the only training set. As shown in the experiments, existing CoSOD models and~\ourmodel{} can show much greater performance when using the proposed~\ourdataset{} dataset as the training set instead of the existing training sets.
    With the combinations of the newly proposed components in this thesis, as mentioned above in novel approaches,~\ourmodel{} achieves great improvement compared with previous works~\cite{GCoNet+,MCCL}, achieving the new SoTA performance among all publicly available CoSOD models~\cite{CoSOD3k}\footnote{Public leaderboard of CoSOD models: \url{https://paperswithcode.com/task/co-saliency-detection}.}. Both the proposed~\ourdataset{} and~\ourmodel{} show their superiorities when compared with existing datasets and methods, respectively.

    \item[\textbf{4.}]
    \textbf{Finally, the potential of applying CoSOD in practical applications is investigated.} Some general discussions are also given on how to promote the development of CoSOD and how to extend existing CoSOD techniques in other areas to help researchers in the CoSOD area for further study. The online demo of detecting co-salient objects can be accessed at \url{https://huggingface.co/spaces/ZhengPeng7/HICOME_demo}.

\end{itemize}

\subsection{Outline of the Thesis}
\label{sec:outline_of_the_thesis}
This thesis consists of 1) an overall review of the Co-Salient Object Detection (CoSOD) task, 2) existing methods, datasets, and their remaining problems, 3) the proposed novel method (\ourmodel{}) and dataset (\ourdataset{}), and 4) future directions of CoSOD and the final conclusions. Specifically, in Section 2, a comprehensive study of the existing CoSOD methods and datasets is provided to give an analysis of them from different perspectives. In Section 3, a detailed description of the proposed new CoSOD method~\ourmodel{} is provided, including the novel components, training tricks, objective function, and the comparison with previous CoSOD methods. In Section 4, the proposed CoSOD dataset~\ourdataset{} is introduced, and a comparison with the previous ones is given. Section 5 consists of extensive experiments conducted, including the datasets, evaluation protocol, implementation details, ablation studies, and comparison experiments with existing competing methods and datasets. In Section 6, some practical applications based on CoSOD are introduced. In Section 7, the conclusions and potential directions for the future are provided.

\subsection{List of Publications}
\label{sec:list_of_publications}
This work is a follow-up work of my previous works on Co-Salient Object Detection, which are listed below. Sections 2 and 6 refer in part to these two works.

\begin{itemize}\setlength\itemsep{0.3em}
    \item[\textbf{I}] \textbf{Peng Zheng}, Huazhu Fu, Deng-Ping Fan, Qi Fan, Jie Qin, Yu-Wing Tai, Chi-Keung Tang, and Luc Van Gool. "GCoNet+: A Stronger Group Collaborative Co-Salient Object Detector." \textit{IEEE Transactions on Pattern Analysis and Machine Intelligence (TPAMI)}, vol. 45, no. 9, pp. 10929-10946, 2023.
    \item[\textbf{II}] \textbf{Peng Zheng}, Jie Qin, Shuo Wang, Tian-Zhu Xiang, and Huan Xiong. "Memory-aided Contrastive Consensus Learning for Co-salient Object Detection." \textit{Thirty-Seventh AAAI Conference on Artificial Intelligence (AAAI)}, Washington DC, United States, vol. 37, no. 3, pp. 3687-3695, 2023.
\end{itemize}

\begin{itemize}\setlength\itemsep{0.3em}
    \item[]\textbf{In Publication I}, Zheng is responsible for all ideas, experiments, the overall writing of the paper, and the online demo. Many thanks to Dr. Deng-Ping Fan, Dr. Huazhu Fu, and Prof. Jie Qin for their detailed valuable advice on writing this article and polishing it up. Thanks to Prof. Ling Shao and Prof. Jie Qin for providing me with the opportunity to do this research at IIAI, UAE.
    
    \item[]\textbf{In Publication II}, Zheng has the responsibility for all ideas, experiments, paper writing, and the online demo. Many thanks to Prof. Jie Qin for helping me in polishing up the paper and improving the responses in the rebuttal. Thanks to Prof. Huan Xiong for offering a position for me to work as a research assistant at MBZUAI, UAE.
\end{itemize}

\thispagestyle{empty}

\clearpage

\section{Related Work}
\label{sec:related_work}

In this section, a comprehensive study of related techniques and existing CoSOD methods is presented for the analysis of them from different perspectives. Then, existing CoSOD datasets (training sets and test sets) are listed and described with a benchmark. The remaining problems of existing CoSOD methods and datasets raise the necessity of the proposed novel method and dataset proposed in section~\secref{sec:methodology}.

\subsection{Salient Object Detection}
\label{sec:SOD}
Salient Object Detection (SOD) has been a traditional computer vision task for many decades, aiming to mimic the human vision system to capture objects that attract the most attention in single images. Progress has been made in detecting salient objects in raw RGB images~\cite{Cheng2011GlobalCB,SOD_att1,SOD_trad1,SOD_op1}, and RGB-D images~\cite{VST,SOD_RGB-D}. In the early years, hand-crafted features (\eg{}, super-pixels~\cite{SOD_patch1,SOD_patch2,SOD_patch3,SOD_patch4}, object proposals~\cite{SOD_op1,SOD_op2,SOD_op3}) were the most widely used computational units, which are extracted to produce the saliency maps with further processing, such as clustering~\cite{SOD_clustering}, contrasting~\cite{Cheng2011GlobalCB,SOD_trad3}, graphing~\cite{grabcut},~\etc{}. These methods have achieved relatively good results in their era, but most of them are rather time-consuming in extracting all the computational units and establishing the relationships between them with traditional algorithms~\cite{SOD_trad1,SOD_trad2,SOD_trad3}. With the rapid development of deep learning and the huge success of fully convolutional networks (FCNs) in segmentation tasks, recent SOD researchers follow the FCN~\cite{FCN} to employ encoder-decoder architectures in their proposed networks to produce pixel-wise prediction of saliency maps~\cite{EGNet,BASNet,PoolNet}.

In terms of the network architecture of SOD models, they mainly have two categories, \ie{}, the CNN-based architecture, and the transformer-based architecture. Based on UNet~\cite{UNet} and FPN~\cite{FPN}, many SOD networks with components designed for extracting and fusing multi-scale or hierarchical features are proposed~\cite{EGNet,PoolNet,SOD1} to aggregate features of objects of different sizes. After the vision transformers show their strong performance on fundamental vision tasks~\cite{ViT,swin_v1,PVTv1}, the corresponding transformer architecture is also adopted in SOD, as the transformer + CNN model~\cite{integrity} and the pure transformer model~\cite{VST}.

In terms of supervision, it is divided into different supervision sources and different supervision styles. At the output of SOD models, apart from the original ground-truth maps, additional information such as boundary~\cite{BASNet}, edge~\cite{EGNet}, and gradient~\cite{SOD_gradient} is also introduced to the SOD networks for auxiliary supervision. Famous attention mechanisms (\eg{}, channel attention~\cite{SENet}, CBAM~\cite{CBAM}, and dual attention~\cite{DANet}) are also tried in many SOD works~\cite{SOD_att1,SOD_att2,SOD_att3,SOD_att4} and show some improvement. At the input of SOD models, signals in other modalities like depth, thermal, and light field are also used as a part to form the RGB-D (3D)~\cite{SOD_RGB-D,SOD_RGB-D1,SOD_RGB-D2}, RGB-T (3D)~\cite{SOD_RGB-T1,SOD_RGB-T2}, and RGB-LF (4D)~\cite{SOD_RGB-LF1,SOD_RGB-LF2} data to achieve more accurate detection. Most of the SOD models are trained in a fully-supervised manner~\cite{integrity,SOD_att1,SOD_att4,SOD_gradient,SOD_RGB-D}, while semi-supervised~\cite{SOD_semi1,SOD_semi2}, weakly-supervised~\cite{SOD_semi1,SOD_semi2}, self-supervised~\cite{SOD_self1,SOD_self2}, and unsupervised~\cite{SOD_clustering,SOD_unsupervised1} SOD approaches also show their effectiveness with less or no training data.

In terms of the quality of produced saliency maps, Zheng~\etal{}~\cite{GCoNet+} introduce the Confidence Enhancement Module where differential binarization is employed to generate more binarized predictions in and use adversarial learning to check the integrity of predicted maps implicitly in~\cite{MCCL}. In~\cite{integrity}, the authors propose the integrity channel enhancement component to guide their model to learn better integrity.

\subsection{Image Co-Segmentation}
\label{sec:CoSegmentation}
Image co-segmentation is a fundamental computer vision task which targets segmenting objects of the same category in a group of images. It has shown the effectiveness in many related tasks, such as few-shot learning~\cite{coseg_fss1,coseg_fss2}, semantic segmentation~\cite{coseg_ss1,coseg_ss2}, co-salient object detection~\cite{CoSOD3k,iCoseg,GWD},~\etc{}. Before the deep-learning era, traditional methods often utilized handcrafted features like energy~\cite{coseg_rw1} and color histograms~\cite{coseg_rw2} to run the comparison between the visual features of each image group. With the development of deep-learning techniques, co-segmentation methods tend to employ a Siamese network to find the co-occurring features within the input image pairs~\cite{coseg_fss1,coseg2} in an end-to-end way. Specifically, in terms of network design, Fan~\etal{}~\cite{CoSOD3k} and Wei~\etal{}~\cite{GWD} employ the co-attention mechanism to generate the consensus features of each group to localize the common objects. Zheng~\etal{}~\cite{coseg_lstm1} and Li~\etal{}~\cite{coseg_lstm2} apply LSTM~\cite{lstm} modules for the information exchange between two image groups and enhance the group-level representation. In terms of training strategy, weakly-supervised methods~\cite{coseg_weakly} and unsupervised methods~\cite{coseg_rw1} also show their effectiveness in co-segmentation.

\subsection{Co-Salient Object Detection}
\label{sec:CoSOD}
CoSOD is a combination of two upstream tasks,~\ie{}, Salient Object Detection (SOD), and Common Object Segmentation (Co-Segmentation). Specifically, CoSOD aims to segment the most common and salient objects in a group of relevant images. This area has attracted widespread attention from the community with hundreds of related papers published in the past few years, which have made a great contribution and promotion to it.


\subsubsection{CoSOD Methods}
\label{sec:CoSOD_methods}
In the early years, researchers split images into computational units (\eg{}, superpixels~\cite{SLIC}, pixel clusters~\cite{CBCS}) for further processing. With these computational units, typical high-level features of them are extracted to generate the semantic representation, such as color~\cite{CBCS}, contrast~\cite{Cheng2011GlobalCB}, and contour~\cite{trad_CoSOD_contour}. These handcrafted features are often employed to determine the common regions with the correspondence model, which can be obtained by similarity ranking, clustering guidance, and translation alignment~\cite{CoSOD_trad_align}. Metric learning~\cite{MetricCoSOD,MetricCoSOD1} and statistics of histograms~\cite{Cheng2011GlobalCB} are also found useful in formulating reasonable semantic attributes.

With the beginning of the deep-learning era, many deep CoSOD methods have been proposed to construct high-level semantic features by learning in an end-to-end manner. These methods have achieved great improvement compared with previous traditional methods.
Wei~\etal{}~\cite{GWD} attempt to find the common objects by distilling the common information within each single group.
In~\cite{CoSOD3k}, a simple co-attention module is placed into an SOD model to find the objects with the most common class in the image group.
In~\cite{GCoNet}, common information is mined with a group affinity module and put into other groups as the contrast for more robust learning.
Dynamic convolution is employed in~\cite{CADC} to capture image-specific cues and the group-wise common knowledge.
In~\cite{MetricCoSOD}, a metric learning regularization term is embedded into support vector machine (SVM) training for assisting in detecting co-salient objects.
In~\cite{GCoNet+}, Zheng~\etal{} strengthen the discriminating ability of the model for more accurate localization of targets and improve the quality of produced saliency maps. This is also the first method to apply metric learning in their deep CoSOD model to learn the more discriminative inter-group features of different classes.
To make better use of the learned consensus, metric learning on the consensuses of different groups is conducted in~\cite{MCCL}, to confirm larger distances among consensuses of different groups.

In addition to the development of CoSOD itself, CoSOD methods also benefit from the boom of upstream CV tasks, such as image classification and object detection. Among the existing CoSOD methods,~\cite{GCoNet,CoADNet,GICD,ren2022adaptive,GCoNet+} build their CoSOD models with powerful CNN models (\eg{}, VGGNet~\cite{VGG}, ResNet~\cite{ResNet}, and Inception~\cite{Inceptionv2v3}). Since the booming of Vision Transformers in 2021 (\eg{}, ViT~\cite{ViT}, Swin~\cite{swin_v1,swin_v2}, and PVT~\cite{PVTv1,PVTv2}), many CoSOD models employ the transformer components as their backbone~\cite{MCCL} or intermediate block~\cite{CoSformer}.
Apart from the network architecture, various styles of supervision are used in existing CoSOD methods, including unsupervised learning~\cite{FASS,qian2022co,GONet,hsu2018co,coseg_rw1}, weakly-supervised learning~\cite{zeng2019joint,qian2022co}, and semi-supervised learning~\cite{FASS}. These methods do not need full annotation, but can still achieve acceptable performance.

\subsubsection{CoSOD Datasets}
\label{sec:CoSOD_datasets}

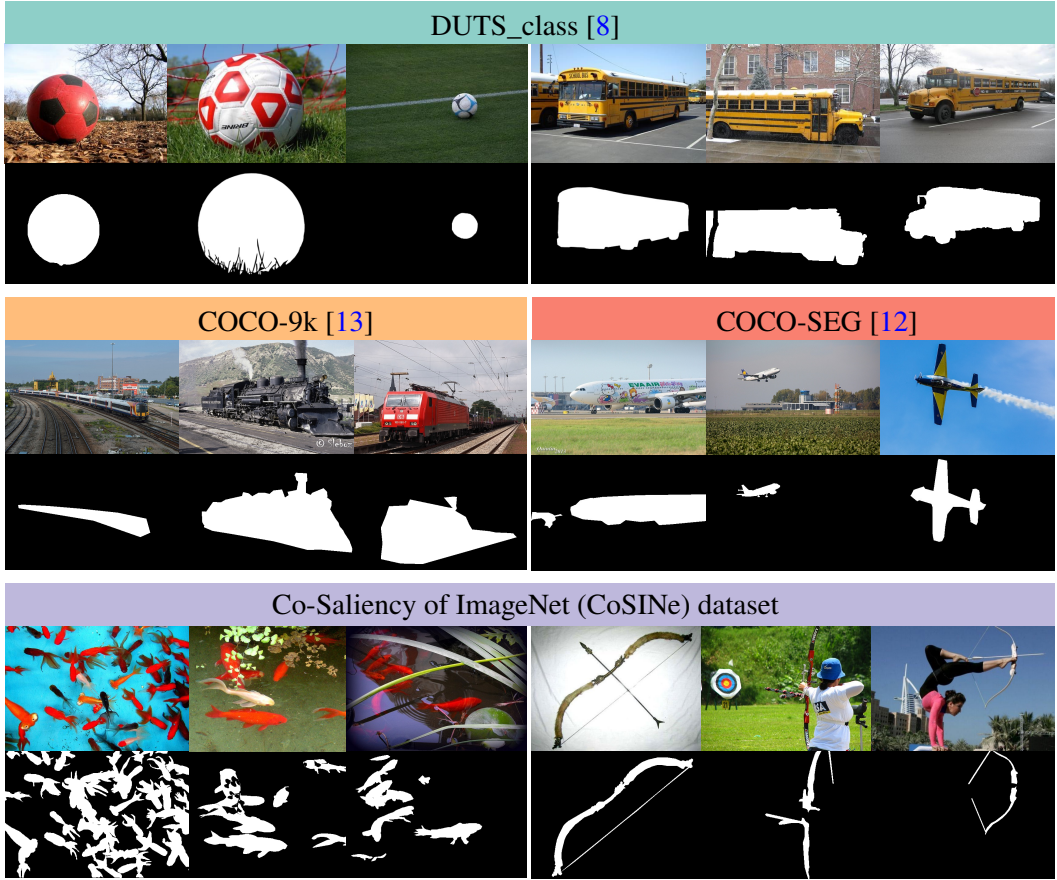
\begin{figure*}[t!]
    \centering
    \small{}
    \begin{overpic}[width=.98\textwidth]{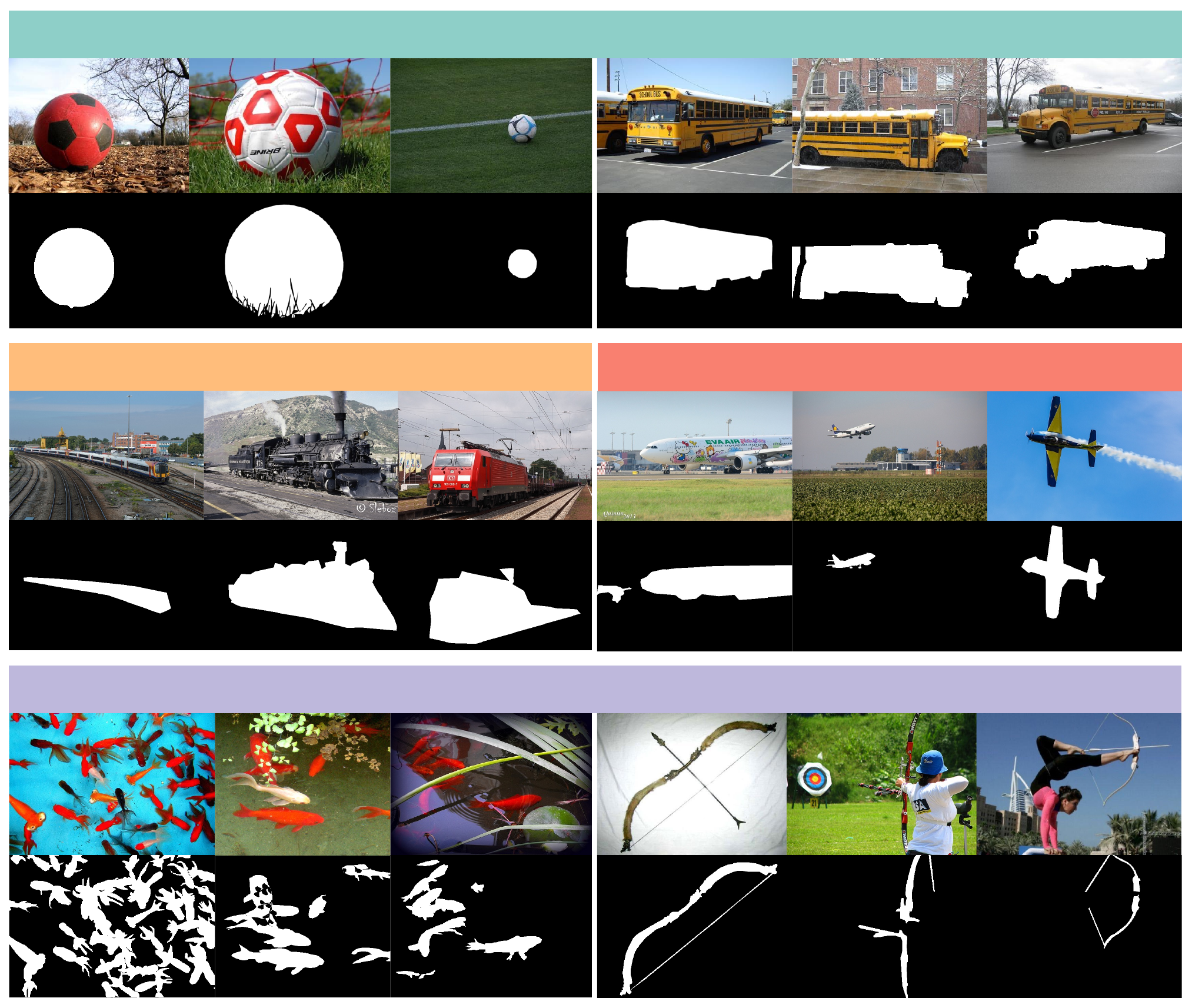}
        \put(41,81.2){DUTS\_class~\cite{GICD}}
        \put(19,53){COCO-9k~\cite{GWD}}
        \put(68,53){COCO-SEG~\cite{COCO_SEG}}
        \put(26,26){Co-Saliency of ImageNet (CoSINe) dataset}
    \end{overpic}
    \caption{\textit{Examples of the widely used CoSOD training sets and the proposed~\ourdataset{} dataset.} Compared with previous training sets,~\ourdataset{} tends to have a larger number of target objects and more distractors in the image, which enables the model to learn the better ability to discriminate target objects surrounded by noise objects in more complicated contexts.}
    \label{fig:dataset_comp}
\end{figure*}

In the past year, only a few CoSOD datasets have been proposed for training~\cite{GWD,GICD,COCO_SEG} and evaluations~\cite{MSRC,iCoseg,li2011co,CoSal2015,CoSOD3k,GICD}.

Some CoSOD \textbf{test sets} are proposed to evaluate the CoSOD methods from different perspectives. Among these datasets, \textit{MSRC}~\cite{MSRC} includes 8 image groups and 240 images for recognizing classes of foreground objects. The \textit{iCoSeg}~\cite{iCoseg} dataset was introduced by Batra~\etal{} in 2010 with 38 groups and 643 images in total. \textit{Image Pair}~\cite{li2011co} is a dataset of images in pairs, specifically designed to evaluate the CoSOD performance in pairs of images, consisting of 210 images and 105 groups in total.

Since 2015, some larger test sets have been proposed to give a more challenging and robust evaluation of the deep CoSOD models. In 2016, Zhang~\etal{}~\cite{CoSal2015} introduced their \textit{CoSal2015} dataset for the evaluation of the CoSOD model, which contains much more images and a larger number of groups. However, most images in \textit{CoSal2015} only contain one salient object in the image, making the evaluation far from the scenarios in practical applications where many surrounding distractors exist. In~\cite{CoSOD3k}, Fan~\etal{} propose the \textit{CoSOD3k} dataset, which contains not only the images divided by group and their corresponding binary ground-truth maps but also the bounding box and segmentation map of every single instance. Compared with \textit{CoSal2015}, \textit{CoSOD3k} contains more objects in each image which play the role of potential distractors, making it more challenging and closer to real-world application. Additionally, the instance-level annotations provide further potential for instance-level CoSOD approaches. In~\cite{GICD}, \textit{CoCA} is the latest CoSOD test set proposed to evaluate the common category object segmentation. \textit{CoCA} is so far the most challenging test set in many aspects, including the number of objects in each image, the various backgrounds, distractors, \etc{}, as shown in \figref{fig:coca_samples}.

In terms of the \textbf{training sets} in CoSOD, there are now three widely used datasets,~\ie{}, DUTS\_class~\cite{GICD}, COCO-9k~\cite{GWD}, and COCO-SEG~\cite{COCO_SEG}. DUTS\_class is modified from the saliency detection dataset DUTS~\cite{DUTS}, where a classifier is employed to divide the images into different groups by the class of the main object in the image. Due to the direct use of ground-truth maps from DUTS~\cite{DUTS}, where the ground-truth maps of all instances in single images may contain instances of different classes,~\ie{}, masks of these instances are wrong annotations which should not exist. In COCO-9k~\cite{GWD} and COCO-SEG~\cite{COCO_SEG}, images and ground-truth maps are directly borrowed from the COCO~\cite{COCO} dataset. The pairs of images and ground-truth maps are divided into different groups by the selected instances with GT maps in the image. Although these training sets have promoted the development of CoSOD greatly, a standard and large training set is still in high demand for further research in CoSOD.

\clearpage

\section{Methodology}
\label{sec:methodology}

This section contains the description of the proposed~\ourmodel{} CoSOD model, including the baseline network and the three novel proposed components,~\ie{}, Hierarchical Consensus Fusion (HCF), Spatial Increment Attention (SIA), and Instance-Aware Contrastive Consensus Learning (IACCL). Several tricks that are specifically useful to train better CoSOD models are also given here, with extensive experiments.

\subsection{Overview}
\label{sec:overview}

\begin{figure*}[t!]
    \centering
    \small{}
    \begin{overpic}[width=.98\textwidth]{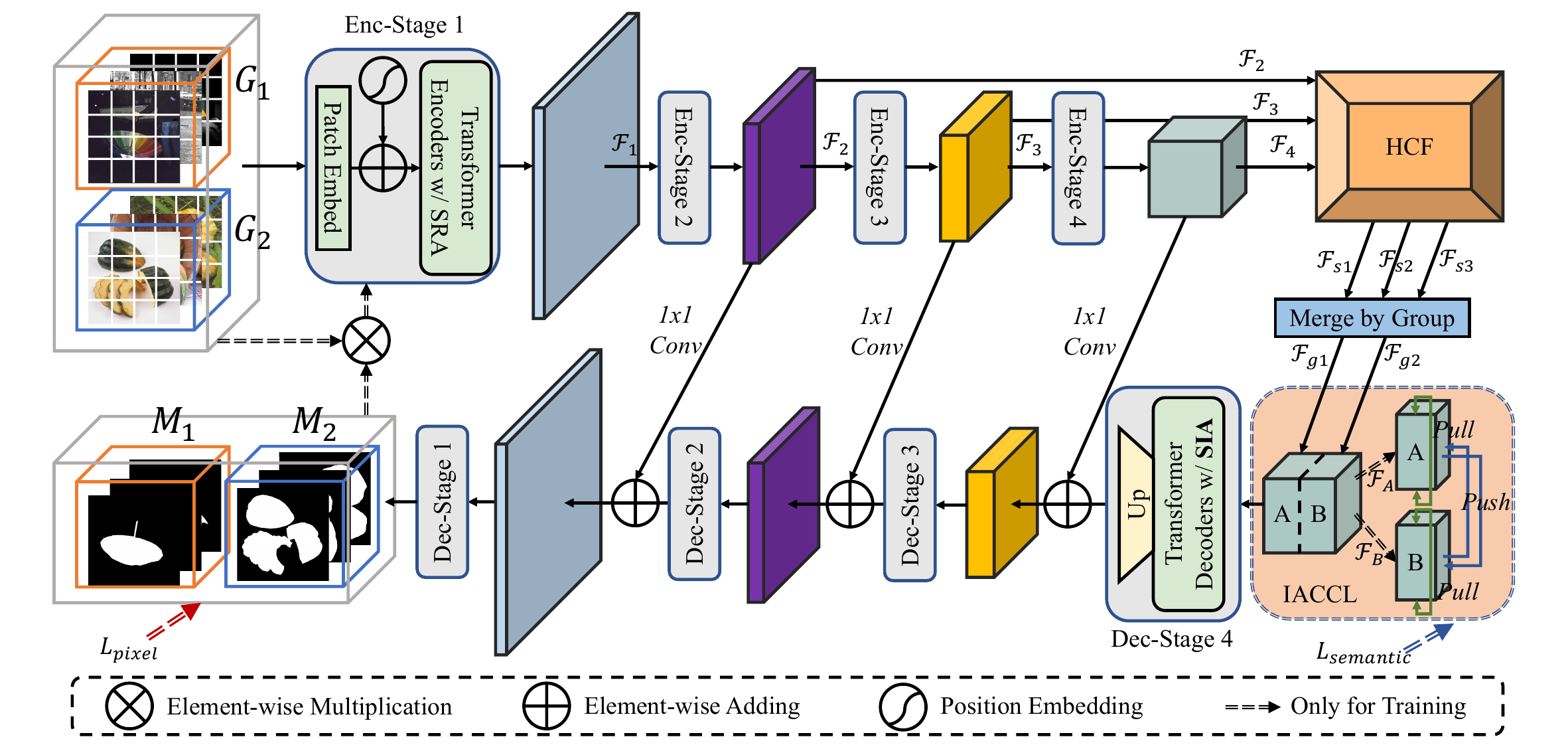}
    \end{overpic}
    \caption{\textit{Overall architecture of the proposed~\ourmodel{}.}}
    \label{fig:ourmodel}
\end{figure*}

\begin{figure*}[t!]
    \centering
    \small{}
    \begin{overpic}[width=.5\textwidth]{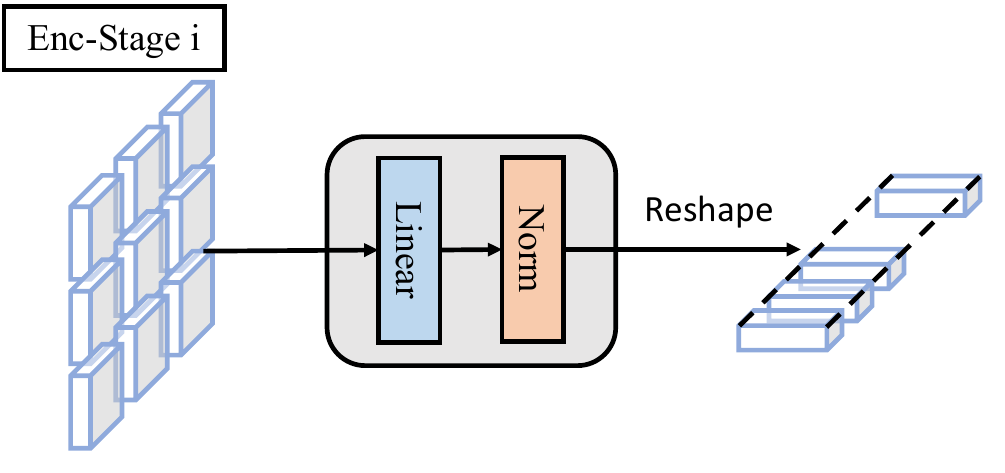}
        \put(12,2){$\frac{H_{i-1}}{P_{i}} \times{} \frac{W_{i-1}}{P_{i}} \times{} (P_{i}^{2}C_{i-1})$}
        \put(70,6){$\frac{H_{i-1} \times{} W_{i-1}}{{P_{i}}^2} \times{} C_{i-1}$}
    \end{overpic}
    \caption{\textit{The composition of the Patch Embedding block.}}
    \label{fig:patch_embedding}
\end{figure*}

The flow chart of~\ourmodel{} is illustrated in~\figref{fig:ourmodel}.
The basic framework of the proposed~\ourmodel{} follows the encoder-decoder structure with the Pyramid Vision Transformer v2~\cite{PVTv2} as the backbone.
First,~\ourmodel{} simultaneously takes two groups of raw images ${G_1, G_2}$ as input. After the concatenation of the two given groups, the backbone extracts the features of different stages, which are denoted as $\mathcal{F}_1, \mathcal{F}_2, \mathcal{F}_3, \mathcal{F}_4$ from shallow to deep. Each PVTv2 backbone consists of a patch embedding layer and a transformer encoder block, as shown in~\figref{fig:patch_embedding} and~\figref{fig:trans_enc_dec}, respectively. Features from different stages of the backbone will be gathered to feed into the hierarchical consensus fusion (HCF) module to obtain further processing. The hierarchical consensus features obtained from the HCF module are generated into groups of different scales,~\ie{}, $\mathcal{F}_{s1}, \mathcal{F}_{s2}, \mathcal{F}_{s3}$, as shown in~\figref{fig:HCF}. Then, these features are reorganized into different units according to their group,~\ie{}, $\mathcal{F}_{g1}, \mathcal{F}_{g2}$. In the normal training process, the consensus features of different groups are concatenated and fed into the following decoder stages.
The decoder block shares an architecture similar to that of the encoder blocks, as shown in~\figref{fig:trans_enc_dec},~\ie{}, the encoder and the decoder both consist of stacked transformer blocks with spatial reduction or spatial increment attention, multi-head attention, and feed forward network, sequentially. Following FPN~\cite{FPN}, the backbone features are sent to the parallel decoder stage with only an $1\times1$ convolution layer. After all stages of the decoder, the final predictions $\mathcal{M}_1, \mathcal{M}_2$ are generated.
In the training procedure, the predicted maps ($\mathcal{M}_1, \mathcal{M}_2$) will be multiplied by the raw images to suppress the role of non-target objects. Then, the multiplication results are fed back into the backbone and the HCF again to generate a more precise representation of the consensus of ${G_1, G_2}$. In the instance-aware contrastive consensus learning module (IACCL), the consensus features of different groups are seen as negative pairs, and those of the same group are treated as positive pairs. A contrastive loss will be used to validate semantic consistency within the same group and semantic discrepancy between different groups. This process is only conducted during training and does not cause additional computation during inference. Finally,~\ourmodel{} is supervised with two types of losses,~\ie{}, saliency loss and semantic loss, which is illustrated in~\secref{sec:objective_function} in detail.

\subsubsection{Hierarchical Consensus Fusion}
\label{sec:HCF}

\begin{figure*}[t!]
    \centering
    \small{}
    \begin{overpic}[width=.98\textwidth]{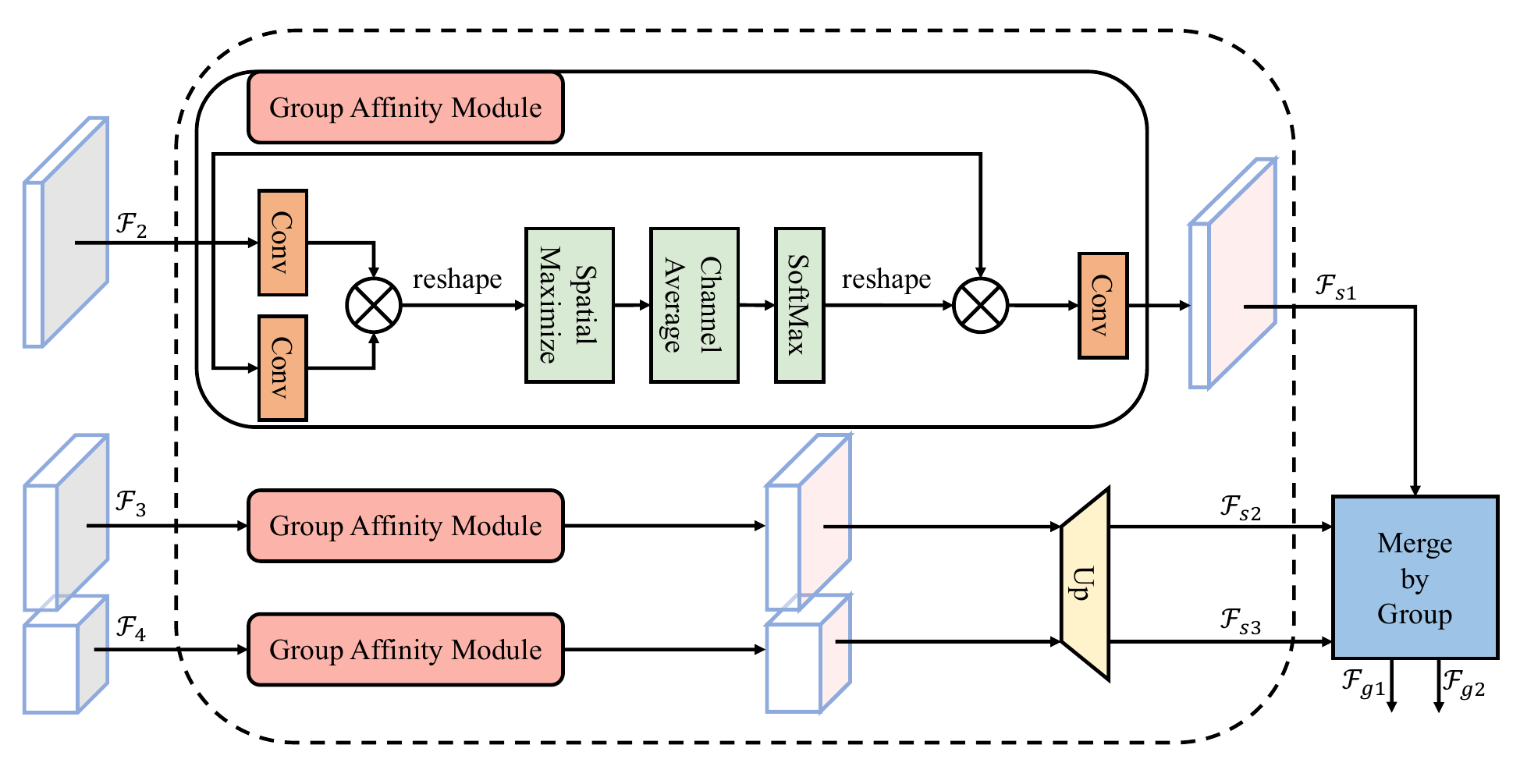}
        \put(30,-0.5){Hierarchical Consensus Fusion}
    \end{overpic}
    \caption{\textit{The structure of the proposed Hierarchical Consensus Fusion (HCF) module.} Features at different scales, from shallow to deep, are transferred into the HCF module. Each feature is used to compute the intra-group consensus.}
    \label{fig:HCF}
\end{figure*}

As discussed in~\secref{sec:research_questions}, the sizes of objects in the wild vary a lot, leading to a worse performance of existing CoSOD models when objects of the same class vary a lot in their sizes. Specifically, we employ the group affinity module~\cite{GCoNet} as a consensus feature extractor on a single scale. As shown in~\figref{fig:HCF}, the backbone features $\mathcal{F}_{2}, \mathcal{F}_{3}, \mathcal{F}_{4}$ are fed into the HCF module. Each feature is processed with a group affinity module to generate a consensus feature at each scale. The consensus features $\mathcal{F}_{s2}, \mathcal{F}_{s3}$ are resized to the size of $\mathcal{F}_{s1}$. Finally, the consensus features $\mathcal{F}_{s1}, \mathcal{F}_{s2}, \mathcal{F}_{s3}$ are split and merged by group.

\subsubsection{Spatial Increment Attention}
\label{sec:SIA}

\begin{figure*}[t!]
    \centering
    \small{}
    \begin{overpic}[width=.98\textwidth]{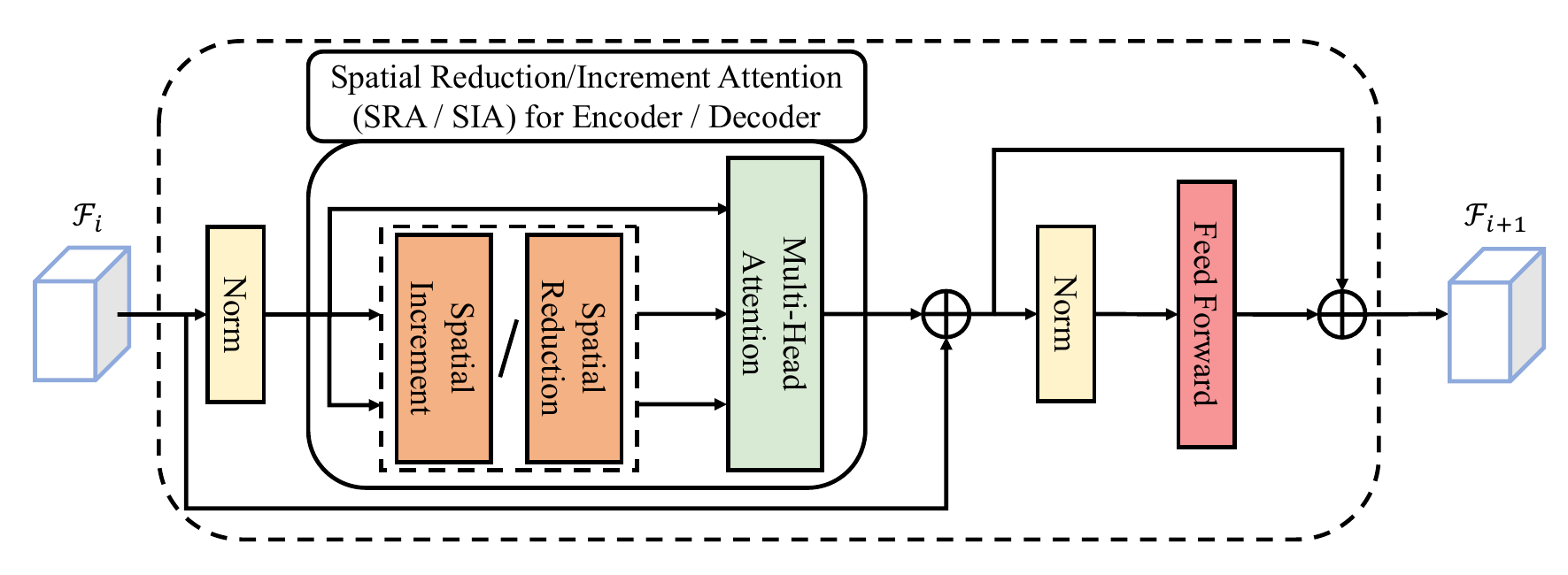}
        \put(30,-0.5){Transformer Encoder / Decoder $\times{}~N$}
    \end{overpic}
    \caption{\textit{The composition of the Transformer Encoder~\cite{PVTv2} and Transformer Decoder.} The number of Transformer Encoders and Decoders ($N$) in each block is kept consistent with the Pyramid Vision Transformer v2~\cite{PVTv2}.}
    \label{fig:trans_enc_dec}
\end{figure*}

Unlike existing segmentation models using transformer blocks only for the encoder, the proposed~\ourmodel{} uses PVTv2~\cite{PVTv2} as the backbone and similar stacked transformer blocks in the decoder. As shown in~\figref{fig:trans_enc_dec}, different from the transformer blocks in the encoder, which aim at squeezing features to lower-resolution ones, the transformer blocks in the decoder replace the spatial reduction attention with the spatial increment attention.

\subsubsection{Instance-Aware Contrastive Consensus Learning}
\label{sec:IACCL}

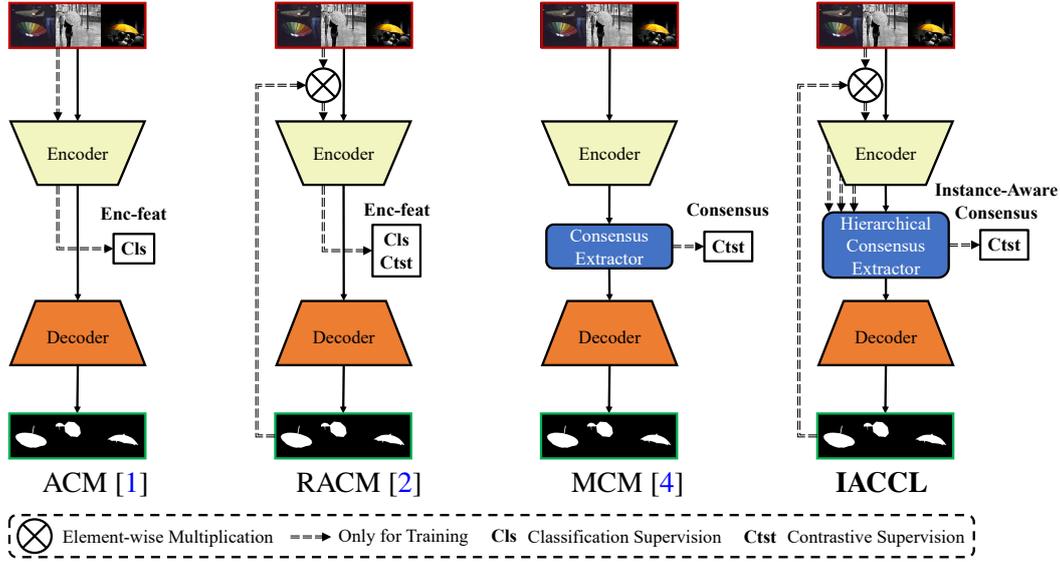
\begin{figure*}[t!]
    \centering
    \small{}
    \begin{overpic}[width=.98\textwidth]{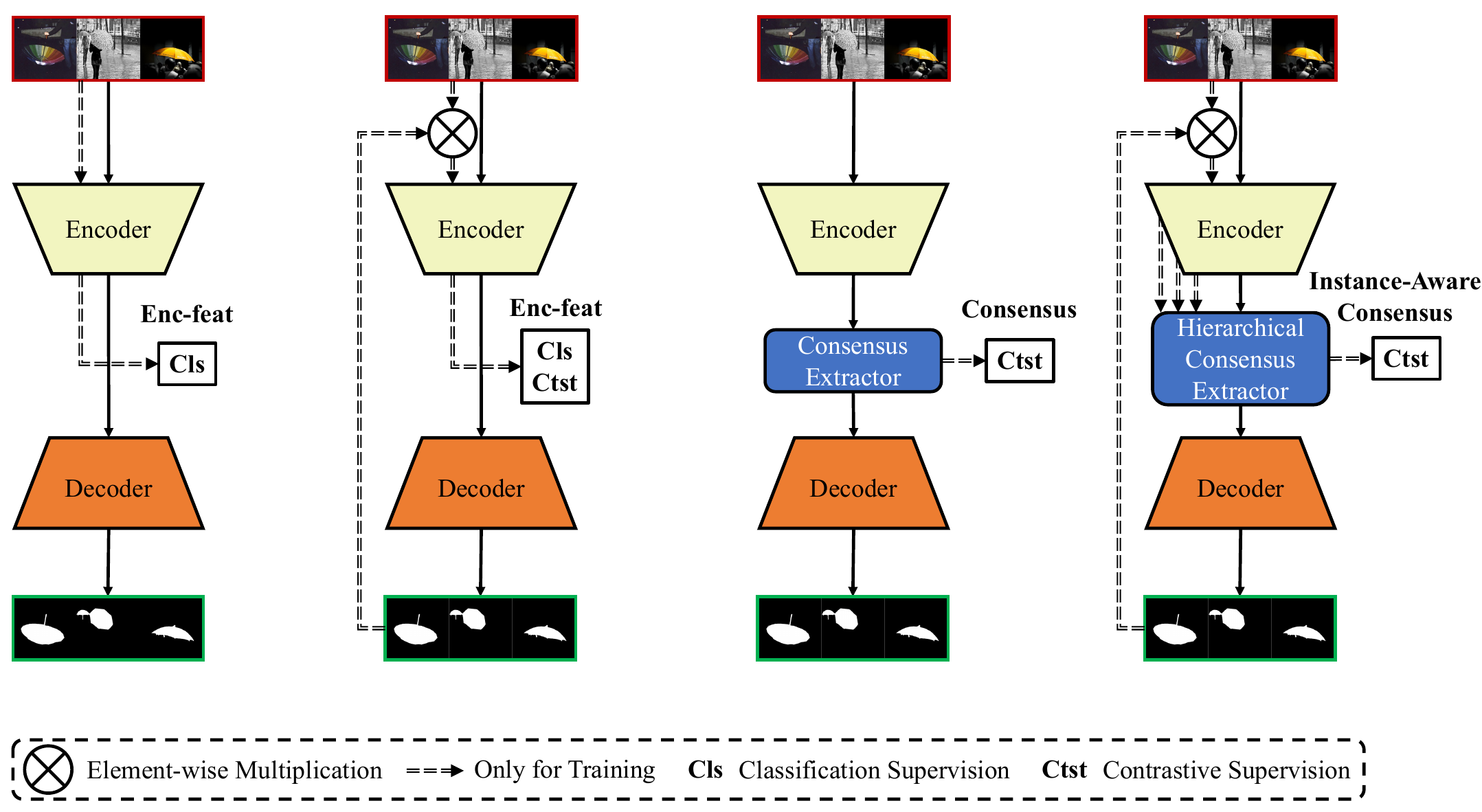}
        \put(4,7){ACM~\cite{GCoNet}}
        \put(28,7){RACM~\cite{GCoNet+}}
        \put(54,7){MCM~\cite{MCCL}}
        \put(79,7){\textbf{IACCL}}
    \end{overpic}
    \caption{\textit{Comparison among previous similar classification / contrast methods and the proposed IACCL.} The hierarchical consensus extractor used here is the proposed HCF in~\secref{sec:HCF}.}
    \label{fig:IACCL}
\end{figure*}

Distractors around target objects are difficult for CoSOD models to eliminate. In previous CoSOD methods, discrimination on semantic features has been used and shown to be effective. In~\cite{GCoNet}, an auxiliary classification module is applied to improve the ability to represent semantics. In~\cite{GCoNet+}, the multiplication between the predicted maps and the original images is fed to the encoder, from which the output is utilized with classification and the proposed group-based symmetric triplet (GST) loss to improve the discrimination ability of the model. In~\cite{MCCL}, momentum consensus features are used to calculate a contrastive loss. As shown in~\figref{fig:IACCL}, in this thesis, an instance-aware contrastive consensus learning (IACCL) is proposed to use hierarchical consensus for contrastive learning. Compared to previous methods, the hierarchical consensus contains not only more consensus at different scales but also fewer noise objects. Benefitting from the mentioned advantages, IACCL can successfully make the generated consensus of different class objects more discriminative to each other.
Given two groups as input,~\ie{}, $group-1$ and $group-2$, the consensus features of them $\mathcal{F}^1, \mathcal{F}^2$ can be obtained as $HCF(Enc(G_1\otimes{}M_1)), HCF(Enc(G_2\otimes{}M_2))$. Similarly to GST loss, $\mathcal{F}^1, \mathcal{F}^2$ are divided into two features of equal length $\mathcal{F}^1_0, \mathcal{F}^1_1$ and $\mathcal{F}^2_0, \mathcal{F}^2_1$. The IACCL loss can be represented as follows:
\begin{equation}
  {L}_\text{IACCL} = {L}_\text{Tri}(~\mathcal{F}^{1}_1,\mathcal{F}^{2}_0,~\mathcal{F}^{2}_1)~+~{L}_\text{Tri}(~\mathcal{F}^{2}_0,\mathcal{F}^{1}_0,~\mathcal{F}^{1}_1),
  \label{eqn:loss_IACCL}
\end{equation}
where $L_{Tri}$ denotes the triplet loss~\cite{TripletLoss}. In each $L_{Tri}$, the first two elements are negative to each other and last two elements are positive to each other.

\subsection{Training Tricks}
\label{sec:tricks}

\textbf{Negative sampling for training.}
Compared to CoSOD model training with only intra-group information~\cite{GWD,CADC,CoADNet}, multiple groups have been shown to be useful in the CoSOD task~\cite{GCoNet,DCFM,GCoNet+,MCCL}. Setting negative samples with all zero ground truth maps is a simple but effective way to improve the robustness of the CoSOD models,~\eg{}, the group contrast module proposed in~\cite{GCoNet} and the group exchange masking introduced in~\cite{CoSOD_group_exchange}. In the implementation of the proposed~\ourmodel{}, we embed this group alternation strategy as a trick to train a more robust CoSOD model. Specifically, the initial image / label of the two given groups are $(G_1, GT_1), (G_2, GT_2)$. Then, the negative samples from other groups are added to the initial image groups of which the ground-truth maps are all zeros, to make the new image / label $(concat(G_1, G_{-1}), concat(GT_1, 0)), (concat(G_2, G_{-2}), concat(GT_2, 0))$, where $G_{-i}$ denotes images from groups other than $group-i$. The quantitative results of the negative sampling can be found in~\tabref{tab:ablation_tricks}.

\textbf{Stable batch padding.}
CoSOD models are often trained by taking part or the entire set of given groups in one iteration. However, as shown in~\tabref{tab:datasets}, the sizes of groups vary a lot in existing CoSOD datasets. Even if the variation is much smaller in the proposed~\ourdataset{}, models may still take two groups whose sizes are very different. In previous training of CoSOD models~\cite{GCoNet,GCoNet+}, models used to take the size of the smaller group as the batch size for the current iteration. Although this strategy can work well, it heavily decreases the batch size in training, which is important to train a good CoSOD model. Furthermore, changing batch sizes among different iterations may cause instability during training.

To address this problem, the proposed stable batch padding is employed here for more robust training of CoSOD models. Given the batch size $N$ and $(G_1, G_2)$ with $(N_1, N_2)$ group size, respectively, $max(N, N_i)$ samples are randomly selected from each group. If $(N_i < N)$, $N - N_i$ samples will be randomly selected from $G_i$ and performed with data augmentation methods, including flip, crop, enhancement,~\etc{}, and inserted into the original $G_i$, followed by shuffling. We also try the adaptive batch padding, where the final batch size is $max(size(G_1), size(G_2))$. After this modification, all training batches will be $N$ samples. As shown in~\tabref{tab:ablation_tricks}, this padding strategy can improve the training of the proposed~\ourmodel{} on the three benchmarks. The final results of the proposed~\ourmodel{} in~\tabref{tab:sota} are obtained in the fixed batch padding setting.

\subsection{Objective Function}
\label{sec:objective_function}
The objective function is a weighted combination of saliency loss and semantic loss (the instance-aware contrastive consensus loss proposed in~\secref{sec:IACCL}).
Saliency loss follows~\cite{GCoNet+} as a weighted combination of BCE loss and IoU loss, which are illustrated as follows:
\begin{equation}
  {L}_\text{BCE} = - \sum{[Y\log(\hat{Y}), (1 - Y)\log(1 - \hat{Y})]},
  \label{eqn:loss_BCE}
\end{equation}
\begin{equation}
  {L}_\text{IoU} = 1 - \frac{1}{N}\sum{\frac{|Y \cap \hat{Y}|}{|Y \cup \hat{Y}|}},
  \label{eqn:loss_IoU}
\end{equation}
where $Y$ is the ground truth and $\hat{Y}$ is the prediction.
With the semantic loss~\eqnref{eqn:loss_IACCL} applied, 
the final objective function is:
\begin{equation}
  {L} = \lambda_1 {L}_\text{BCE} + \lambda_2 {L}_\text{IoU} + \lambda_3 {L}_\text{IACCL},
  \label{eqn:loss_final}
\end{equation}
where $\lambda_1,~\lambda_2,~\lambda_3$ are respectively set to 30, 0.5, and 3 to keep all the losses on the same quantitative level at the beginning of training.

\clearpage

\section{\ourdataset{} Dataset}
\label{sec:dataset}

\begin{table*}[!t]
\centering
\scriptsize{}
\renewcommand{\arraystretch}{1.0}
\setlength\tabcolsep{4pt}
\caption{\textit{Comparison between existing CoSOD datasets.} All the existing CoSOD datasets are listed here, including both the training sets and test sets. The group size and resolution are represented as mean value~$\pm{}$~standard deviation.}
\resizebox{0.99\linewidth}{!}{
\begin{tabular}{l||c|c|cccc}
    \thickhline
    \textbf{Dataset} & \textbf{Pub. \& Year} & \textbf{Type} & \textbf{\#Image} & \textbf{\#Group} & \textbf{Group Size} & \textbf{Resolution (HxW)} \\
    \hline \hline
    \rowcolor{mygray}  
    MSRC~\cite{MSRC} & ICCV 2005 & Test & 240 & 8 & 33.3~$\pm{}$~8.0 & 224.9~$\pm{}$~33.7~$\times{}$~308.1~$\pm{}$~33.7 \\
    \hline
    iCoSeg~\cite{iCoseg} & CVPR 2010 & Test & 643 & 38 & 16.9~$\pm{}$~10.4 & 385.3~$\pm{}$~73.7~$\times{}$~462.2~$\pm{}$~64.9 \\
    \hline
    \rowcolor{mygray}  
    Image Pair~\cite{li2011co} & TIP 2011 & Test & 210 & 105 & 2 & 105.4~$\pm{}$~43.1~$\times{}$~131.1~$\pm{}$~28.9 \\
    \hline
    CoSal2015~\cite{CoSal2015} & IJCV 2016 & Test & 2,015 & 50 & 40.3~$\pm{}$~5.5 & 395.2~$\pm{}$~125.2~$\times{}$~475.3~$\pm{}$~144.4 \\
    \hline
    \rowcolor{mygray}  
    CoSOD3k~\cite{CoSOD3k} & TPAMI 2021 & Test & 3,316 & 160 & 20.7~$\pm{}$~7.8 & 423.1~$\pm{}$~162.7~$\times{}$~489.4~$\pm{}$~199.7 \\
    \hline
    CoCA~\cite{GICD} & ECCV 2020 & Test & 1,295 & 80 & 16.2~$\pm{}$~6.6 & 456.4~$\pm{}$~80.3~$\times{}$~617.3~$\pm{}$~63.2 \\
    \hline \hline

    \rowcolor{mygray}  
    DUTS\_class~\cite{GICD} & ECCV 2020 & Train & 8,250 & 291 & 28.4~$\pm{}$~25.6 & 319.3~$\pm{}$~51.2~$\times{}$~381.8~$\pm{}$~40.5 \\
    \hline
    \rowcolor{mygray}  
    COCO-9k~\cite{GWD} & IJCAI 2017 & Train & 9,213 & 65 & 141.7~$\pm{}$~108.8 & 480.2~$\pm{}$~93.2~$\times{}$~582.0~$\pm{}$~89.8 \\
    \hline
    COCO-SEG~\cite{COCO_SEG} & AAAI 2019 & Train & $\sim$200k & 78 & 2576.1~$\pm{}$~5576.5 & 489.8~$\pm{}$~95.3~$\times{}$~579.4~$\pm{}$~89.7 \\
    \hline
    \hline
    \rowcolor{mygray}
    \ourdataset{} & Sub. 2023 & Train & \textbf{22,978} & \textbf{919} & \textbf{25.0~$\pm{}$~5.9} & \textbf{421.3~$\pm{}$~171.4~$\times{}$~483.1~$\pm{}$~192.5} \\
    \thickhline
\end{tabular}
}
\label{tab:datasets}
\end{table*}

In this section, existing training sets for CoSOD and analyses of their drawbacks are provided. The comparison between existing training sets and the proposed~\ourdataset{} dataset from different aspects are also given to show the superiority of~\ourdataset{} in training a better deep CoSOD model.

\subsection{Existing CoSOD Training Sets}
\label{sec:existing_cosod_trainingsets}

These three training sets,~\ie{}, DUTS\_class~\cite{GICD}, COCO-9k~\cite{GWD}, and COCO-SEG~\cite{COCO_SEG} have pushed forward the area of CoSOD. However, the annotations of these two datasets are of low quality (\eg{}, polygon masks, rough contours, see~\figref{fig:dataset_comp}). Another problem of COCO-9k and COCO-SEG is that the numbers of groups in them are only 65 and 78, respectively. Suffering from these problems, a standard and effective training set of high quality and a large number of classes is in high demand.

\begin{figure}[t!]
    \begin{overpic}[width=.98\textwidth]{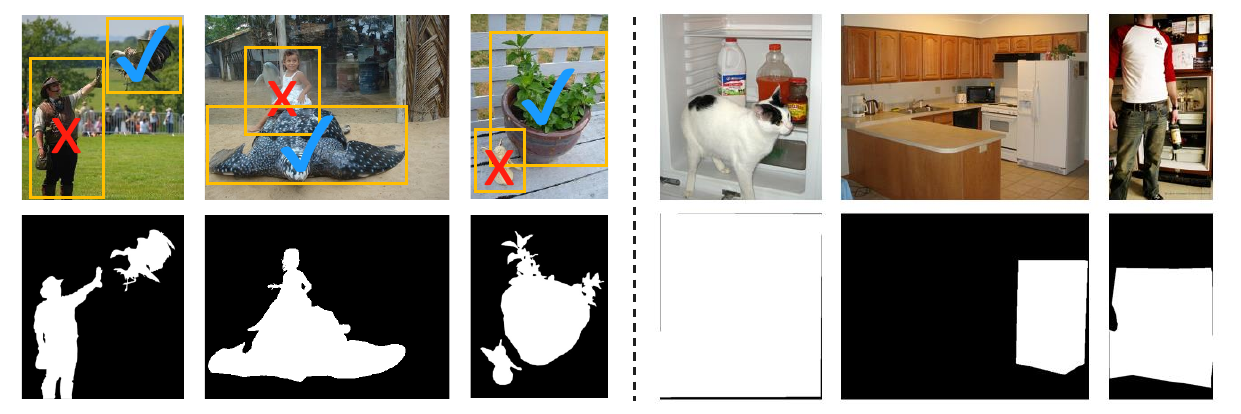}
        \put(18,-2){DUTS\_class~\cite{GICD}}
        \put(57,-2){COCO-9k/COCO-SEG~\cite{GWD,COCO_SEG}}
    \end{overpic}
\caption{\textit{GT maps with problems occur in DUTS\_class, COCO-9k, and COCO-SEG datasets as shown in Publication I.} Due to the lack of instance-level annotations or with them, these datasets do not correctly eliminate the objects with wrong features in their GT maps. DUTS\_class has the problem that objects with wrong classes exist in the GT maps. COCO-9k and COCO-SEG have the problem that objects which are not salient are included.}
\label{fig:problems_TRsets}
\end{figure}

Apart from the drawbacks of existing training sets (as shown in~\figref{fig:problems_TRsets}), there are still some improvements that were not achieved. \textbf{First}, the main task of CoSOD is to find objects of a common class in the group, which makes the ability of the models to discriminate the class of objects more important. Especially, CoSOD methods should work well on the group in which the category of objects may never be seen. Thus, a larger number of classes/groups in the training set should help the model a lot in learning the common information features of objects. \textbf{Second}, the number of image of all classes is better to be balanced to avoid leaning toward some dominant classes they learned, for example, persons.

Based on these two principles, the proposed \textit{\textbf{Co-Saliency of ImageNet}} (\ourdataset{}) dataset is introduced, which does a lot better on the points mentioned above and shows great improvement when used as the training set to the existing models and~\ourmodel{}. 
The proposed~\ourdataset{} is mainly based on the ImageNet-1k~\cite{ImageNet} and ImageNet-S~\cite{ImageNet-S} dataset, which provides the raw image and semantic segmentation labels.

\subsection{Dataset Features and Statistics}
\label{sec:dataset_statistics}
The features and some statistics of the proposed~\ourdataset{} are discussed as follows:

\begin{itemize}\setlength\itemsep{0.3em}
    \item
    \textit{Resolution distribution.} As described in~\tabref{tab:datasets}, the mean resolution of the images given in~\ourdataset{} is $421\times483$, which is higher than that of most of the existing CoSOD datasets and applicable to practical applications.

    \item
    \textit{Saliency of annotated objects.} To avoid the problem that some annotated objects are not salient, ground-truth maps of objects in~\ourdataset{} have been checked for their salientness for more accurate training of CoSOD models in order to detect salient objects.

    \item
    \textit{Large number of groups.} A large number of different groups contribute greatly to the model's ability to discriminate objects of different classes.~\ourdataset{} has the largest number of groups (919) among all CoSOD datasets.

    \item
    \textit{Balanced numbers of images in groups.} The number of images in different groups in~\ourdataset{} is much more balanced compared to other existing CoSOD training sets. In CoSOD model training, balanced groups help CoSOD models with a more stable training procedure.

\end{itemize}
In summary, as shown in~\tabref{tab:datasets}, the proposed CoSINE dataset has 22,978 images in 919 groups. Compared to existing training sets, ~\ourdataset{} has the largest number of groups. Additionally,~\ourdataset{} has the most stable group sizes. All these factors make the proposed~\ourdataset{} currently the best training set for training CoSOD models, which is confirmed by further experiments in~\secref{sec:competing_datasets}.

\clearpage

\section{Experiments}
\label{sec:experiments}
This section provides the guidelines and details of the basic and extensive experiments,~\ie{}, datasets, settings, evaluation protocol, and analysis in training and testing, respectively. Comprehensive ablation studies are provided to show the effectiveness of each proposed module in~\ourmodel{}. Comparison between existing datasets and the proposed~\ourdataset{} is also made to show the superiority of the proposed one and its value in promoting the development of CoSOD.

\subsection{Datasets}
\label{sec:datasets}
\textbf{Training Sets.}
On the one side, following the GICD~\cite{GICD}, DUTS\_class is used as the training set to design the experiments. After removing the noisy samples by Zhang~\etal{}~\cite{GICD}, the whole DUTS\_class is divided into 291 groups, which contain 8,250 images in total. The DUTS\_class dataset is the only training set used for evaluation in the ablation study. 
Nowadays, there is still a lack of a widely accepted training set. To make a fair comparison with up-to-date works~\cite{GWD,COCO_SEG,ICNet,CADC,CoADNet}, the widely-adopted COCO-9k~\cite{GWD}, a subset of the COCO~\cite{COCO} with 9,213 images of 65 groups, and the COCO-SEG~\cite{COCO_SEG} which is also a subset of the COCO~\cite{COCO} and contains 200k images (refer to~\tabref{tab:datasets} for more details of datasets), are employed to train the proposed~\ourmodel{}~as supplementary experiments.

On the other side, to deal with the absence of a standard and effective training set in CoSOD, the~\ourdataset{} dataset is proposed, which is applied as the training set for~\ourmodel{} and several representative and latest methods to provide a better evaluation on the performance of these models. Additionally, the generalization ability comparison among all these datasets is also conducted,~\ie{}, DUTS\_class~\cite{GICD}, COCO-9k~\cite{GWD}, COCO-SEG~\cite{COCO_SEG}, and the proposed~\ourdataset{}.

\textbf{Test Sets.}
To obtain a comprehensive evaluation of the proposed~\ourmodel{} and~\ourdataset{}, three widely used CoSOD datasets are employed for the test,~\ie{}, CoCA~\cite{GICD}, CoSOD3k~\cite{CoSOD3k}, and CoSal2015~\cite{CoSal2015}. Among these three datasets, CoCA is the most challenging one. 
It is of much higher diversity and complexity in terms of background, occlusion, illumination, surrounding objects,~\etc{}. Following the latest benchmark~\cite{CoSOD3k}, iCoseg~\cite{iCoseg} and MSRC~\cite{MSRC} are not used for evaluation, since only one salient object is given in most of their images. It is more convincing to evaluate CoSOD methods on images with more salient objects, which is closer to real-life applications.

\subsection{Evaluation Protocol}
Following the GCoNet~\cite{GCoNet}, the S-measure~\cite{Smeasure}, maximum F-measure~\cite{Fmeasure}, maximum E-measure~\cite{Emeasure}, mean E-measure~\cite{Emeasure}, mean F-measure~\cite{Fmeasure}, and mean absolute error (MAE) are employed to evaluate the performance in the experiments.\footnote{The evaluation toolbox can be referred to \url{https://github.com/ZhengPeng7/CoSINE}.}

\textbf{E-measure}~\cite{Emeasure} is designed as a perceptual metric to evaluate the similarity between the predicted maps and the ground-truth maps from both local and global views. 
E-measure is defined as:
\begin{equation}
  E_{\xi} = \frac{1}{W \cdot H}\sum_{x=1}^{W}\sum_{y=1}^{H}\phi_{\xi}(x, y),
  \label{eqn:Em}
\end{equation}
where $\phi_{\xi}$ indicates the \tbd{enhanced alignment matrix}. Similar to the F-measure, the max E-measure ($E_{\xi}^{max}$) is also adopted as the evaluation metric.

\textbf{S-measure}~\cite{Smeasure} is a structural similarity measurement between a saliency map and its corresponding ground truth map. The evaluation with $S_{\alpha}$ can be obtained at high speed without binarization. S-measure is computed as:
\begin{equation}
  S_{\alpha} = \alpha \cdot S_o + (1 - \alpha) \cdot S_{r},
  \label{eqn:Sm}
\end{equation}
where \tbd{$S_o$ and $S_r$} denote instance-aware and region-aware structural similarity, respectively, and $\alpha$ is set to 0.5 by default, as suggested by Fan~\etal{}in~\cite{Smeasure}.

\textbf{F-measure}~\cite{Fmeasure} is designed to evaluate the weighted harmonic mean value of precision and recall. The output of the saliency map is binarized with different thresholds to obtain a set of binary saliency predictions. The predicted saliency maps and ground-truth maps are compared for precision and recall values. The maximum F-measure score obtained with the best threshold for the entire dataset is defined as $F_{\beta}^{max}$. F-measure can be computed as:
\begin{equation}
  F_{\beta} = \frac{(1+\beta^2) Precision \cdot Recall}{\beta^2 (Precision + Recall)},
  \label{eqn:Fm}
\end{equation}
where $\beta^2$ is set to 0.3 to emphasize the precision over recall, following~\cite{SOD_review1}.

\textbf{MAE} $\epsilon$ is a simple pixel-level evaluation metric that measures the absolute difference between the predicted maps and the ground truth maps without binarization.
It is defined as:
\begin{equation}
  \epsilon = \frac{1}{W \times H}\sum_{x=1}^{W}\sum_{y=1}^{H}|\hat{Y}(x,y) - \text{GT}(x, y)|.
  \label{eqn:MAE}
\end{equation}

\subsection{Implementation Details}
\label{sec:imple}

In terms of network structure, following~\cite{MCCL}, PVTv2~\cite{PVTv2} is employed as the backbone network in~\ourmodel{}. The group affinity module proposed in~\cite{GCoNet} is used as a basic module for single-scale consensus extraction.

In terms of training, due to the different sizes of groups, the default batch size might be smaller than the size of the current group. Therefore, we employ the stable batch padding strategy to generate batches of static length (32) for more stable training (refer to~\secref{sec:tricks} for more details).
The images are resized to 256$\times{}$256 for training. Three data augmentation strategies are employed in the training process, \ie{}, horizontal flip, color enhancement, and rotation. The proposed~\ourmodel{} is trained for 120 epochs with the Adam optimizer~\cite{Adam}, and the initial learning rate $l_r$ is set to $3\cdot10^{-4}$, $\beta{}_1=0.9$, and $\beta{}_2=0.99$. The learning rate will decrease by 10 in the last 20th epoch. The training process needs $\sim$36GB GPU memory and takes around 13 hours. All the experiments are implemented on the PyTorch~\cite{PyTorch} framework, with a single NVIDIA A100 GPU.

In terms of testing, the batch size is set as the full size of the single given group, which is consistent with existing works~\cite{CADC,GCoNet,UFO,GCoNet+,MCCL}. The images are also resized to 256$\times{}$256 for the test. The proposed~\ourmodel{} makes the inference at 67.2 fps.

\subsection{Ablation Studies}
\label{sec:ablation}

\begin{figure*}[t!]
    \centering
    \scriptsize{}
    \begin{overpic}[width=.95\textwidth]{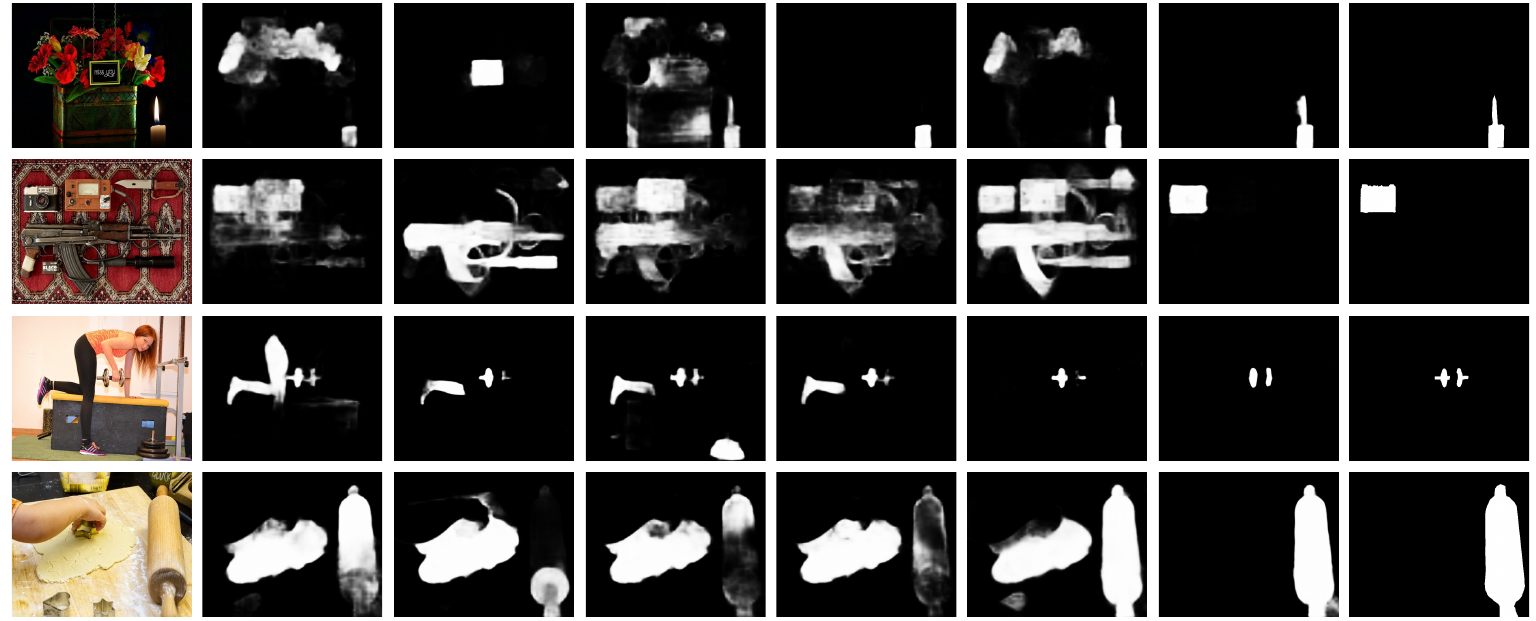}
        \put(-1.2,33){\rotatebox{90}{Candle}}
        \put(-1.2,22){\rotatebox{90}{Camera}}
        \put(-1.2,11){\rotatebox{90}{Dumbbell}}
        \put(-1.2,0.){\rotatebox{90}{Rolling Pin}}
        \put(5.5,-2){{(a)}}
        \put(18.5,-2){{(b)}}
        \put(30.5,-2){{(c)}}
        \put(42.5,-2){{(d)}}
        \put(55.5,-2){{(e)}}
        \put(67.5,-2){{(f)}}
        \put(80.5,-2){{(g)}}
        \put(93,-2){{(h)}}
    \end{overpic}
	\caption{\textit{Qualitative ablation studies of the proposed~\ourmodel{}~on different modules and their combinations.} (a) Source image; (b) Baseline; (c) SIA; (d) HCF; (e) HCF+IACCL; (f) HCF+SIA; (g) HCF+SIA+IACCL, the final version of the proposed~\ourmodel{}; (h) Ground-truth maps.}
	\label{fig:qualitive_ablation}
\end{figure*}
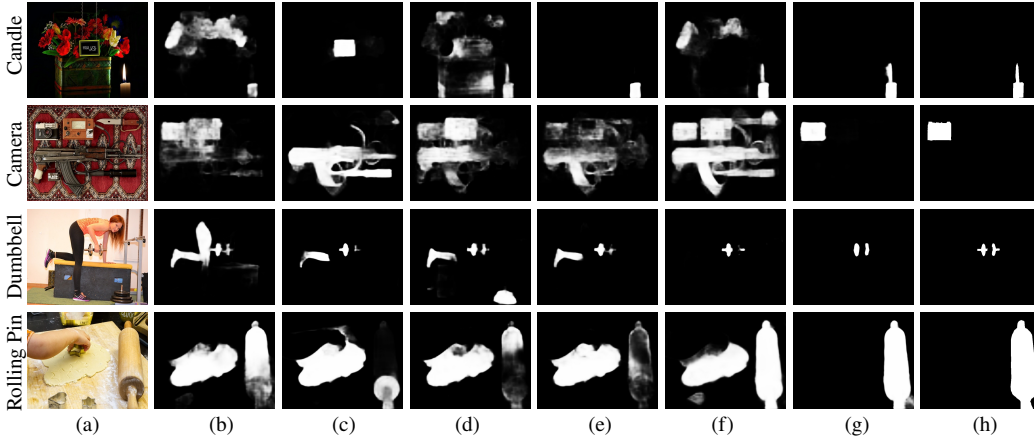

The effectiveness of each component proposed in the proposed~\ourmodel{} was examined,~\ie{}, HCF, SIA, and IACCL. Investigate is also made to show how these components improve the model and how much improvement is brought about by each of them. The qualitative results for each component are shown in~\figref{fig:qualitive_ablation}.

\begin{table*}[t!]
\begin{center}
\footnotesize
\renewcommand{\arraystretch}{1.2}
\setlength\tabcolsep{4pt}
\caption{\textit{Quantitative ablation studies of the proposed components in the proposed~\ourmodel{}.} The ablation studies of the proposed~\ourmodel{}~ are conducted on the effectiveness of the proposed components, including HCF (Hierarchical Consensus Fusion), SIA (Spatial Increment Attention), IACCL (Instance-Aware Contrastive Consensus Learning), and their combinations.}
\label{tab:ablation_modules}
\tiny{}
\begin{tabular}{c|ccc||cccc|cccc|cccc}
\hline
& \multicolumn{3}{c||}{Modules}  & \multicolumn{4}{c|}{CoCA~\cite{GICD}} & \multicolumn{4}{c|}{CoSOD3k~\cite{CoSOD3k}} & \multicolumn{4}{c}{CoSal2015~\cite{CoSal2015}} \\
ID &  \hspace{1.25mm}HCF\hspace{1.25mm} & SIA & IACCL & $E_{\xi}^\text{ max} \uparrow$ & $S_\alpha \uparrow$ & $F_\beta^\text{ max} \uparrow$ & $\epsilon \downarrow$ & $E_{\xi}^\text{ max} \uparrow$ & $S_\alpha \uparrow$ & $F_\beta^\text{ max} \uparrow$ & $\epsilon \downarrow$ & $E_{\xi}^\text{ max} \uparrow$ & $S_\alpha \uparrow$ & $F_\beta^\text{ max} \uparrow$ & $\epsilon \downarrow$ \\
\hline
1 &  &  &  & 0.784 & 0.720 & 0.594 & 0.092 & 0.936 & 0.888 & 0.883 & 0.042 & 0.949 & 0.905 & 0.916 & 0.037 \\
2 & \checkmark &  &  & 0.782 & 0.723 & 0.596 & 0.088 & 0.937 & 0.892 & 0.888 & 0.040 & 0.955 & 0.913 & 0.925 & 0.032 \\
3 &  & \checkmark &  & 0.796 & 0.716 & 0.583 & 0.091 & \textbf{0.944} & \textbf{0.904} & \textbf{0.902} & \textbf{0.035} & 0.956 & 0.917 & 0.929 & 0.030 \\
4 & \checkmark & \checkmark &  & 0.813 & 0.741 & 0.626 & \textbf{0.082} & 0.939 & 0.898 & 0.893 & 0.039 & 0.956 & 0.919 & 0.930 & 0.031 \\
4 & \checkmark &  & \checkmark & 0.786 & 0.728 & 0.598 & 0.083 & 0.935 & 0.892 & 0.885 & 0.040 & 0.953 & 0.913 & \textbf{0.923} & 0.033 \\
\hline
\rowcolor{mygray}
5 & \checkmark & \checkmark & \checkmark & \textbf{0.816} & \textbf{0.746} & \textbf{0.634} & 0.091 & \textit{0.943} & \textit{0.903} & \textit{0.899} & \textit{0.038} & \textbf{0.957} & \textbf{0.919} & \textit{0.929} & \textbf{0.030} \\
\hline
\end{tabular}
\end{center}
\end{table*}

\textit{Baseline}
The baseline network follows previous SoTA methods~\cite{GCoNet,GCoNet+,MCCL} to design the proposed~\ourmodel{} in a Siamese way for full supervision. First, following~\cite{GCoNet+}, all extensive modules and blocks in~\cite{MCCL} are removed to obtain a simple FPN architecture network after the accuracy-efficiency trade-off. Then, similarly to~\cite{MCCL}, the Pyramid Vision Transformer v2~\cite{PVTv2} is used as a replacement for the CNN backbone (\eg{}, VGG~\cite{VGG}, ResNet~\cite{ResNet}, InceptionNet~\cite{Inceptionv2v3}) to obtain robust features on multiple scales. In the encoder-decoder architecture, encoder features are transferred to the decoder in parallel with 1x1 convolutions. The supervision is only employed at the final prediction head with a hybrid loss of binary cross-entropy (BCE) loss and intersection over union (IoU) loss. Finally, a single group affinity module~\cite{GCoNet} is added to connect the end of the encoder and the beginning of the decoder, which has shown its effectiveness in previous works~\cite{GCoNet,GCoNet+}. On the basis of these modifications, a simple, efficient and strong baseline network is obtained, which has already shown decent results as shown in~\tabref{tab:ablation_modules} compared with existing methods in~\tabref{tab:sota}.

\textit{Effectiveness of \textbf{HCF}.} The HCF combines the features of different stages of the encoder to extract the consensus of objects on different scales. With feature fusion, the HCF can better capture objects of various scales and extract more consensus among objects of the same category but on different scales. As shown in the row of candles in~\figref{fig:qualitive_ablation}, models with HCF can better find the right targets among many objects of various scales. Also, HCF brings $\sim$0.5 Smeasure improvement on average on the three existing test sets.

\textit{Effectiveness of \textbf{SIA}.} SIA is added to the original MHA in each spatial increment module to enhance its key and value. SIA is introduced to bring larger receptive fields to the decoder blocks. As qualitative and quantitative results shown in~\figref{fig:qualitive_ablation} and~\tabref{tab:ablation_modules}, respectively, the SIA greatly improves the model in terms of general segmentation ability. It also helps the model achieve the highest performance on the CoSOD3k test set in the ablation study.

\textit{Effectiveness of \textbf{IACCL}.} Although contrastive learning has shown its effectiveness in the CoSOD task~\cite{MetricCoSOD,GCoNet+,MCCL}, the distraction of surrounding non-target objects may add noise to the feature which causes the failure of learning. Therefore, the accurate extraction of only the target object features is vital to the success of contrastive learning in the CoSOD framework. The IACCL is introduced to lead the model to eliminate the pixels of the wrong objects. In addition, the consensus of the right objects is computed to achieve more robust contrastive learning. As experiments show in~\tabref{tab:ablation_modules}, the IACCL can further improve the performance of the proposed network, while introducing zero extra computation during inference.

\begin{table*}[t!]
\begin{center}
\tiny{}
\renewcommand{\arraystretch}{1.2}
\setlength\tabcolsep{4pt}
\caption{\textit{Quantitative ablation studies of the tricks in the strategy of training the proposed~\ourmodel{}.} The ablation studies of the proposed~\ourmodel{}~ are conducted on the effectiveness of overall tricks on the framework, including the negative sampling, adaptive batch padding, and fixed batch padding, which are denoted by \textit{Neg-Sam}, \textit{BP-Adap}, and \textit{BP-Fix}, respectively.}
\label{tab:ablation_tricks}
\tiny{}
\begin{tabular}{c|ccc||cccc|cccc|cccc}
\hline
& \multicolumn{3}{c||}{Modules}  & \multicolumn{4}{c|}{CoCA~\cite{GICD}} & \multicolumn{4}{c|}{CoSOD3k~\cite{CoSOD3k}} & \multicolumn{4}{c}{CoSal2015~\cite{CoSal2015}} \\
ID & Neg-Sam & \hspace{-1.25mm}BP-Adap\hspace{-1.25mm} & BP-Fix & $E_{\xi}^\text{ max} \uparrow$ & $S_\alpha \uparrow$ & $F_\beta^\text{ max} \uparrow$ & $\epsilon \downarrow$ & $E_{\xi}^\text{ max} \uparrow$ & $S_\alpha \uparrow$ & $F_\beta^\text{ max} \uparrow$ & $\epsilon \downarrow$ & $E_{\xi}^\text{ max} \uparrow$ & $S_\alpha \uparrow$ & $F_\beta^\text{ max} \uparrow$ & $\epsilon \downarrow$ \\
\hline
1 &  &  &                       & 0.816 & 0.746 & 0.634 & 0.091 & 0.943 & 0.903 & 0.899 & 0.038 & 0.957 & 0.919 & 0.929 & 0.030  \\
2 & \checkmark &  &             & 0.815 & 0.746 & 0.638 & 0.090 & 0.942 & 0.901 & 0.897 & 0.038 & 0.956 & 0.918 & 0.928 & 0.031  \\
3 &  & \checkmark &             & 0.814 & 0.749 & 0.641 & 0.085 & 0.943 & 0.902 & 0.898 & 0.037 & 0.955 & 0.916 & 0.929 & 0.031  \\
3 &  &  & \checkmark            & 0.820 & 0.750 & 0.640 & \textbf{0.082} & 0.941 & 0.901 & 0.895 & 0.037 & \textbf{0.960} & \textbf{0.923} & \textbf{0.933} & \textbf{0.027}  \\
4 & \checkmark & \checkmark &   & 0.817 & 0.746 & 0.641 & 0.088 & 0.942 & 0.900 & 0.901 & 0.038 & 0.955 & 0.919 & 0.928 & 0.030  \\
\hline
\rowcolor{mygray}
5 & \checkmark &  & \checkmark  & \textbf{0.820} & \textbf{0.752} & \textbf{0.647} & 0.086 & \textbf{0.945} & \textbf{0.904} & \textbf{0.903} & \textbf{0.036} & \textit{0.957} & 0.917 & 0.927 & \textit{0.030}  \\
\hline
\end{tabular}
\end{center}
\end{table*}

\textit{Effectiveness of \textbf{training tricks}.} In the experiments, some training tricks that are specifically useful for training CoSOD models are provided, including negative sampling, adaptive batch padding, and fixed batch padding. Negative sampling is similar to the group contrast module proposed in~\cite{GCoNet}, where part of the images of different groups are put into the current batch as noise, with the all-zero maps as their ground-truth maps. The adaptive batch padding and fixed batch padding are used to deal with the problem that the sizes of groups in the training set vary a lot. As a consequence, the model cannot receive a stable and large size of the given batches during training. In addition, models with siamese training~\cite{GCoNet,GCoNet+,MCCL} take samples from multiple groups, where the inconsistent group sizes in a single batch are not acceptable. Therefore, it can either pad the smaller groups with randomly selected images from themselves to keep consistent with the larger ones, or extend both of them to a fixed size (\eg{}, 32) in the same way. As the experiments given in~\tabref{tab:ablation_tricks} show, these three training tricks can improve the performance of the proposed~\ourmodel{} to varying degrees.

\begin{figure*}[t!]
    \centering
    \footnotesize{}
    \begin{overpic}[width=.95\textwidth]{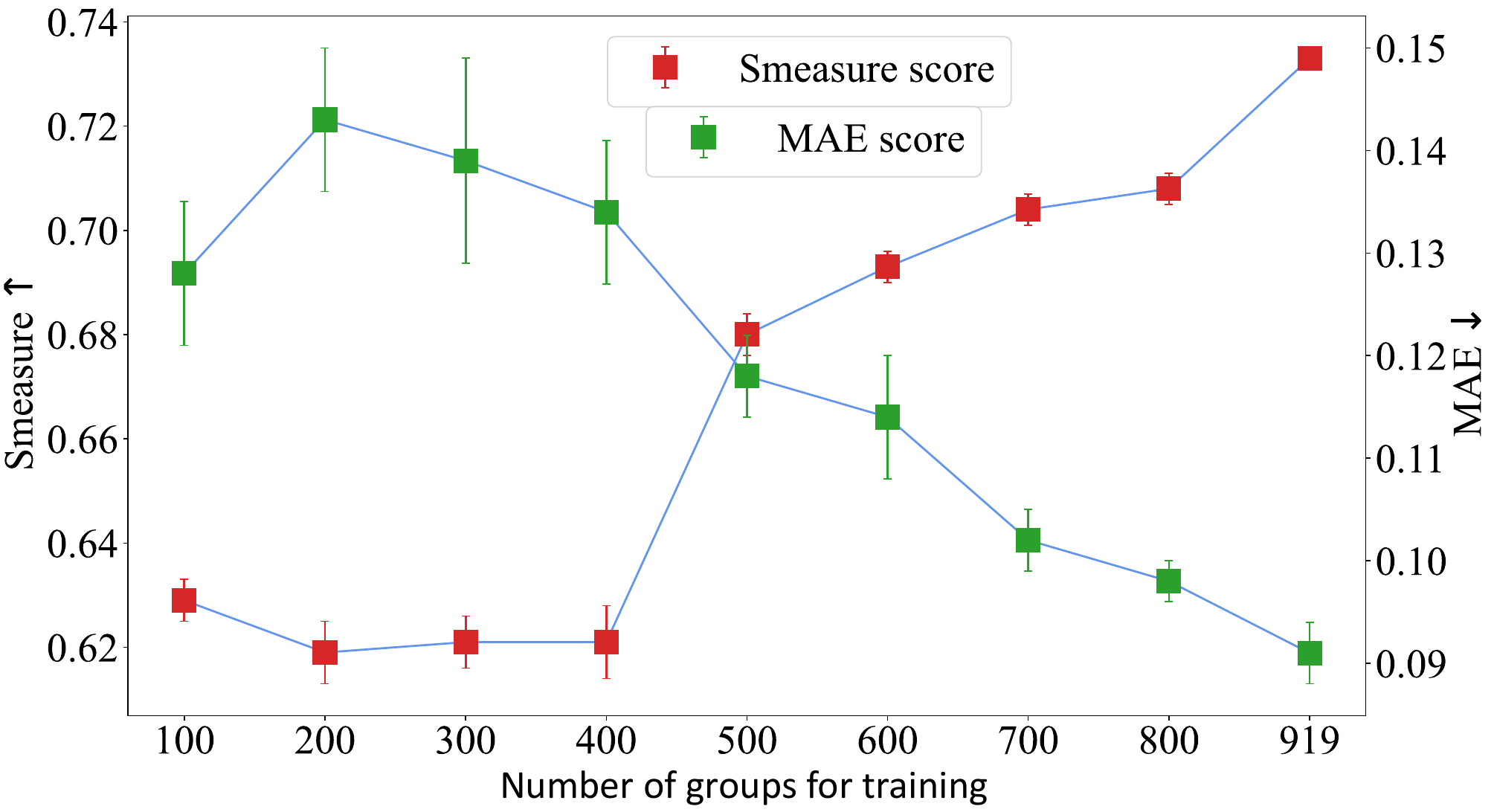}
    \end{overpic}
    \caption{\textit{Performance of the proposed model trained with different numbers of groups of the proposed~\ourdataset{} dataset.} The same model is trained without training tricks on different numbers of groups of~\ourdataset{} dataset. Smeasure and MAE scores are given with their standard deviation error bars for a clearer demonstration. ``$\uparrow$'' (``$\downarrow$'') means that the higher (lower) is better.}
    \label{fig:perf_different_groups}
\end{figure*}

\textit{Effectiveness of \textbf{large number of groups} to learn CoSOD.} To validate how important the number of groups is for models aimed at detecting co-salient objects, experiments of training the same model (the proposed~\ourmodel{}) with different numbers of groups of the complete~\ourdataset{} dataset. As shown in~\figref{fig:perf_different_groups}, the model maintains a relatively low performance level with small numbers of groups, but increases dramatically after 500 groups are used in training. The trend of the relationship between the number of groups and the performance is shown as expected. Multiple rounds of experiments are also conducted on each setting to provide error bars of the performance perturbation for a more robust evaluation.

\subsection{Competing Methods}
\label{sec:competing_methods}

\begin{table*}[t!]
\begin{center}
\renewcommand{\arraystretch}{1.2}
\renewcommand{\tabcolsep}{1.14mm}
\tiny{}
\caption{\textit{Quantitative comparisons between the proposed~\ourmodel{}~and other methods, trained with existing training sets.} Methods are attached with links to open-source codes or paper sources. Since there are several datasets used in the CoSOD task for training, all the training sets (denoted as TR) used in corresponding methods are listed here,~\ie{}, 1, 2, and 3 represent the DUTS\_class~\cite{GICD}, COCO-9k~\cite{GWD}, and COCO-SEG~\cite{COCO_SEG}, respectively.} 
\label{tab:sota}
\begin{tabular}{r||r|cccc|cccc|cccc}
\hline
 &  & \multicolumn{4}{c|}{CoCA~\cite{GICD}} & \multicolumn{4}{c|}{CoSOD3k~\cite{CoSOD3k}} & \multicolumn{4}{c}{CoSal2015~\cite{CoSal2015}} \\
Method & Pub. \& Year | Tr & $E_{\xi}^\text{max} \uparrow$ & $S_\alpha \uparrow$ & $F_\beta^\text{ max} \uparrow$ & $\epsilon \downarrow$ & $E_{\xi}^\text{ max} \uparrow$ & $S_\alpha \uparrow$ & $F_\beta^\text{ max} \uparrow$ & $\epsilon \downarrow$ & $E_{\xi}^\text{ max} \uparrow$ & $S_\alpha \uparrow$ & $F_\beta^\text{ max} \uparrow$ & $\epsilon \downarrow$ \\
\hline

\href{https://github.com/HzFu/CoSaliency_tip2013}{CBCS}~\cite{CBCS} & TIP 2013 | - & 0.641 & 0.523 & 0.313 & 0.180 & 0.637 & 0.528 & 0.466 & 0.228 & 0.656 & 0.544 & 0.532 & 0.233 \\
\hline

\href{https://github.com/zzhanghub/gicd}{GICD}~\cite{GICD} & ECCV 2020 | 1 & 0.715 & 0.658 & 0.513 & 0.126 & 0.848 & 0.797 & 0.770 & 0.079 & 0.887 & 0.844 & 0.844 & 0.071 \\
\href{https://github.com/DengPingFan/CoEGNet}{CoEGNet}~\cite{CoSOD3k} & TPAMI 2021 | 1 & 0.717 & 0.612 & 0.493 & 0.106 & 0.837 & 0.778 & 0.758 & 0.084 & 0.884 & 0.838 & 0.836 & 0.078 \\
\href{https://github.com/fanq15/GCoNet}{GCoNet} & CVPR 2021 | 1 & 0.760 & 0.673 & 0.544 & 0.105 & 0.860 & 0.802 & 0.777 & 0.071 & 0.887 & 0.845 & 0.847 & 0.068 \\
\href{https://github.com/ZhengPeng7/GCoNet_plus}{GCoNet+}~\cite{GCoNet+}	& TPAMI 2023 | 1 & 0.786 & 0.691 & 0.574 & 0.113 & 0.881 & 0.828 & 0.807 & 0.068 & 0.917 & 0.875 & 0.876 & 0.054 \\
\hline

\href{https://www.ijcai.org/proceedings/2017/0424.pdf}{GWD}~\cite{GWD} & IJCAI 2017 | 2 & 0.701 & 0.602 & 0.408 & 0.166 & 0.777 & 0.716 & 0.649 & 0.147 & 0.802 & 0.744 & 0.706 & 0.148 \\
\href{https://www.ijcai.org/proceedings/2019/0115.pdf}{RCAN}~\cite{RCAN} & IJCAI 2019 | 2 & 0.702 & 0.616 & 0.422 & 0.160 & 0.808 & 0.744 & 0.688 & 0.130 & 0.842 & 0.779 & 0.764 & 0.126 \\
\href{https://github.com/blanclist/ICNet}{ICNet}~\cite{ICNet} & NeurIPS 2020 | 2 & 0.698 & 0.651 & 0.506 & 0.148 & 0.832 & 0.780 & 0.743 & 0.097 & 0.900 & 0.856 & 0.855 & 0.058 \\
\href{https://github.com/siyueyu/DCFM}{DCFM}~\cite{DCFM} & CVPR 2022 | 2 & 0.783 & 0.710 & 0.598 & 0.085 & 0.874 & 0.810 & 0.805 & 0.067 & 0.892 & 0.838 & 0.856 & 0.067 \\ 
\href{https://github.com/ZhengPeng7/GCoNet_plus}{GCoNet+}~\cite{GCoNet+}	& TPAMI 2023 | 2 & 0.798 & 0.717 & 0.605 & 0.098 & 0.877 & 0.819 & 0.796 & 0.075 & 0.902 & 0.853 & 0.857 & 0.073 \\
\hline

\href{https://github.com/ltp1995/GCAGC-CVPR2020}{GCAGC}~\cite{GCAGC} & CVPR 2020 | 3 & 0.754 & 0.669 & 0.523 & 0.111 & 0.816 & 0.785 & 0.740 & 0.100 & 0.866 & 0.817 & 0.813 & 0.085 \\
\href{https://openaccess.thecvf.com/content/CVPR2021/papers/Zhang_DeepACG_Co-Saliency_Detection_via_Semantic-Aware_Contrast_Gromov-Wasserstein_Distance_CVPR_2021_paper.pdf}{DeepACG}~\cite{DeepACG} & CVPR 2021 | 3 & 0.771 & 0.688 & 0.552 & 0.102 & 0.838 & 0.792 & 0.756 & 0.089 & 0.892 & 0.854 & 0.842 & 0.064 \\
\href{https://github.com/suyukun666/UFO}{UFO}~\cite{UFO} & TMM 2023 | 3 & 0.782 & 0.697 & 0.571 & 0.095 & 0.874 & 0.819 & 0.797 & 0.073 & 0.906 & 0.860 & 0.865 & 0.064 \\ 
\href{https://github.com/ZhengPeng7/GCoNet_plus}{GCoNet+}~\cite{GCoNet+}	& TPAMI 2023 | 3 & 0.787 & 0.712 & 0.602 & 0.100 & 0.875 & 0.820 & 0.793 & 0.075 & 0.899 & 0.853 & 0.852 & 0.071 \\
\hline

\href{https://github.com/nnizhang/CADC}{CADC}~\cite{CADC} & ICCV 2021 | 1,2 & 0.744 & 0.681 & 0.548 & 0.132 & 0.840 & 0.801 & 0.859 & 0.096 & 0.906 & 0.866 & 0.862 & 0.064 \\ 
\href{https://github.com/ZhengPeng7/GCoNet_plus}{GCoNet+}~\cite{GCoNet+}	& TPAMI 2023 | 1,2 & 0.808 & 0.734 & 0.626 & 0.088 & 0.894 & 0.839 & 0.822 & 0.065 & 0.919 & 0.876 & 0.880 & 0.058 \\
\hline

\href{https://github.com/KeeganZQJ/CoSOD-CoADNet}{CoADNet}~\cite{CoADNet} & NeurIPS 2020 | 1,3 & - & - & - & - & 0.878 & 0.824 & 0.791 & 0.076 & 0.914 & 0.861 & 0.858 & 0.064 \\
\href{https://github.com/ZhengPeng7/MCCL}{MCCL}~\cite{MCCL} & AAAI 2023 | 1,3 & 0.796 & 0.714 & 0.590 & 0.103 & 0.903 & 0.858 & 0.837 & 0.061 & 0.927 & 0.890 & 0.891 & 0.051  \\
\href{https://github.com/ZhengPeng7/GCoNet_plus}{GCoNet+}~\cite{GCoNet+}	& TPAMI 2023 | 1,3 & 0.814 & 0.738 & 0.637 & \textbf{0.081} & 0.901 & 0.843 & 0.834 & 0.062 & 0.924 & 0.881 & 0.891 & 0.056 \\
\hline

\href{https://github.com/ZhengPeng7/MCCL}{MCCL}~\cite{MCCL} & AAAI 2023 |~\ourdataset{} & 0.787 &  0.726 & 0.605 & 0.090 & 0.939 & 0.892 & 0.889 & 0.040 & 0.952 & 0.908 & 0.921 & 0.035  \\
\href{https://github.com/ZhengPeng7/GCoNet_plus}{GCoNet+}~\cite{GCoNet+}	& TPAMI 2023 |~\ourdataset{} & 0.811 & 0.725 & 0.645 & 0.108 & 0.912 & 0.853 & 0.854 & 0.061 & 0.930 & 0.876 & 0.896 & 0.053 \\
\rowcolor{mygray}
\href{https://github.com/ZhengPeng7/CoSINe}{\ourmodel{}} & Sub. 2023 |~\ourdataset{} & \textbf{0.820} & \textbf{0.752} & \textbf{0.647} & 0.086 & \textbf{0.945} & \textbf{0.904} & \textbf{0.903} & \textbf{0.036} & \textbf{0.957} & \textbf{0.917} & \textbf{0.927} & \textbf{0.030}  \\
\hline
\end{tabular}
\end{center}
\end{table*}

Since not all CoSOD models are publicly available, only the proposed~\ourmodel{}~with one representative traditional algorithm CBCS~\cite{CBCS} and~\NumDLCoSODComp{} deep learning-based CoSOD models are compared, including all up-to-date models,~\ie{}, GWD~\cite{GWD}, RCAN~\cite{RCAN}, CSMG~\cite{CSMG}, GCAGC~\cite{GCAGC}, GICD~\cite{GICD}, ICNet~\cite{ICNet}, CoADNet~\cite{CoADNet}, CoEGNet~\cite{CoSOD3k}, DeepACG~\cite{DeepACG}, CADC~\cite{CADC}, UFO~\cite{UFO}, DCFM~\cite{DCFM}, GCoNet+~\cite{GCoNet+}, and MCCL~\cite{MCCL}. With the rapid development of the CoSOD method, they have performed much better than the single-SOD methods in recent years. Thus, single SOD methods are not taken into account in the comparison. Due to the limit of computing resources and poor reproducibility of some methods,  the latest methods are also trained with open-source codes~\cite{MCCL,GCoNet+} on the proposed training set~\ourdataset{} for a more comprehensive comparison.

\textbf{Quantitative Results.}~\tabref{tab:sota} shows the quantitative results of the proposed~\ourmodel{} and previous state-of-the-art methods. With~\ourdataset{} as the training set, the proposed~\ourmodel{} outperforms all the previous methods in the widely used metrics. Among these three test sets, CoCA~\cite{GICD} is the latest and most challenging dataset due to the more surrounding objects and the diverse backgrounds. It puts more emphasis on finding common objects. CoSal2015~\cite{CoSal2015} is a relatively easier test set, which puts more attention on finding the salient objects. CoSOD3k~\cite{CoSOD3k} is a more balanced test set in terms of these two aspects. As the results show, the proposed~\ourmodel{} has a much better performance not only on CoCA, but also on CoSal2015 and CoSOD3k. It means that the proposed~\ourmodel{} has a more powerful ability to find common objects and detect salient objects.

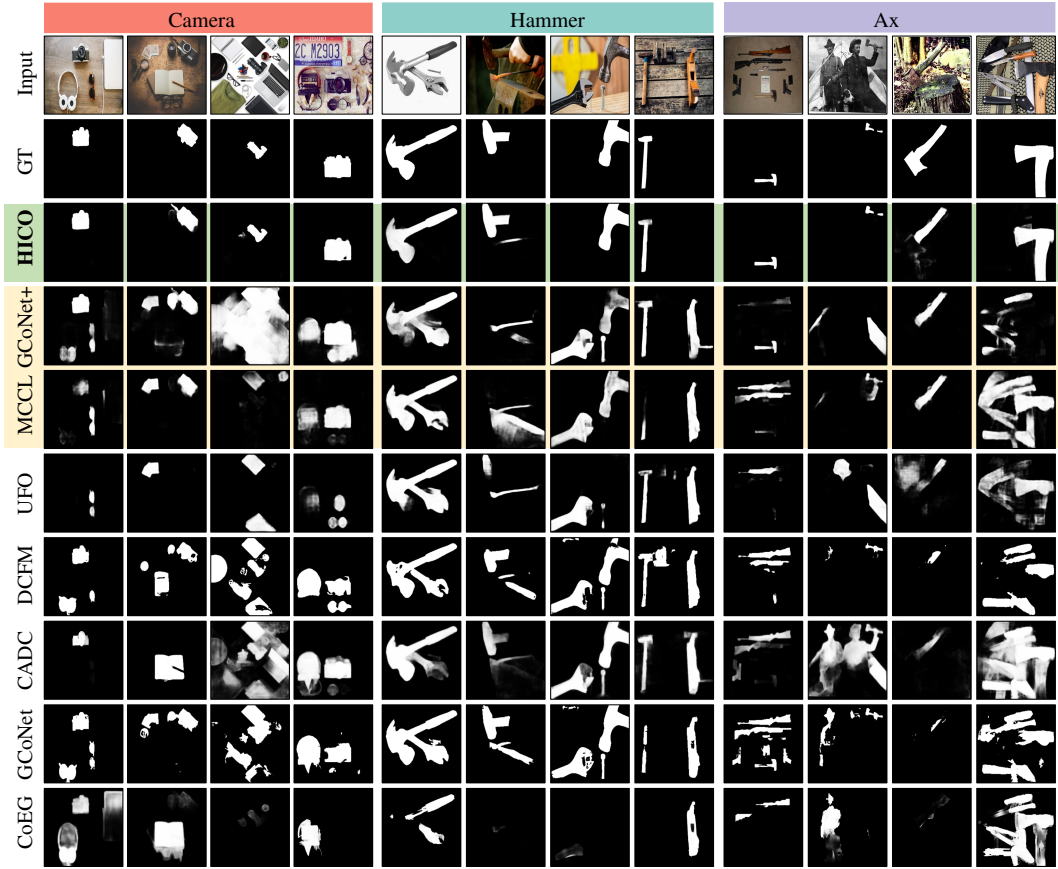
\begin{figure*}[t!]
    \centering
    \scriptsize{}
    \begin{overpic}[width=.99\textwidth]{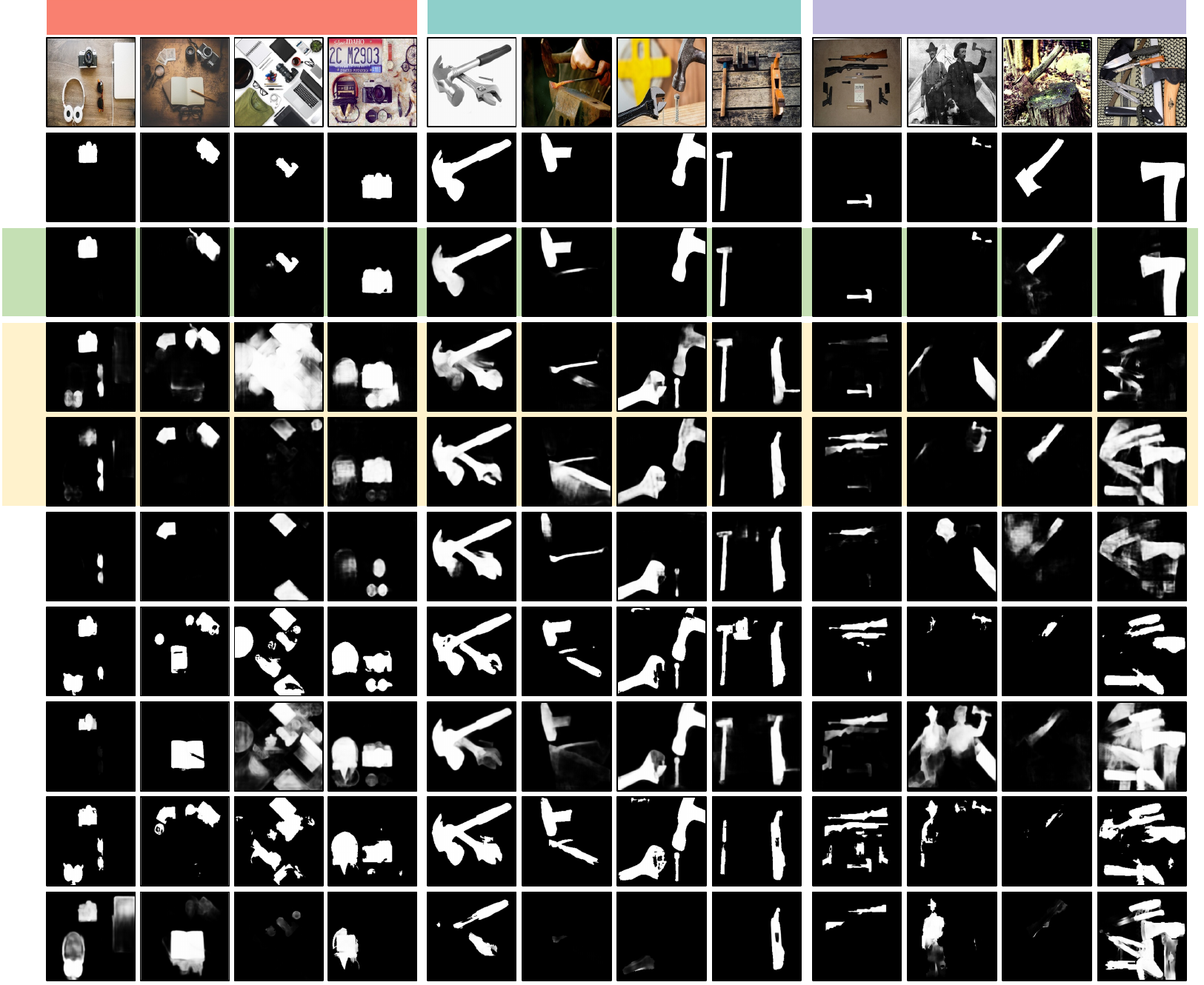}
        \put(15.5,79.5){{Camera}}
        \put(47.5,79.5){{Hammer}}
        \put(81.5,79.5){{Ax}}
        \put(1.4,73){\rotatebox{90}{Input}}
        \put(1.4,65.5){\rotatebox{90}{GT}}
        \put(1.4,57){\rotatebox{90}{\textbf{HICO}}}
        \put(1.4,47.31){\rotatebox{90}{GCoNet+}}
        \put(1.4,41){\rotatebox{90}{MCCL}}
        \put(1.4,33.5){\rotatebox{90}{UFO}}
        \put(1.4,25){\rotatebox{90}{DCFM}}
        \put(1.4,17){\rotatebox{90}{CADC}}
        \put(1.4,8.5){\rotatebox{90}{GCoNet}}
        \put(1.4,2){\rotatebox{90}{CoEG}}
    \end{overpic}
    \caption{\textit{Qualitative comparisons of the proposed~\ourmodel{}~and other methods.} ``GT'' denotes the ground truth. The predictions in the row with the green background are produced by the proposed~\ourmodel{}. The results of Publication I~\cite{GCoNet+} and Publication II~\cite{MCCL} mentioned in~\secref{sec:list_of_publications} are shown on a yellow background.}
    \label{fig:qualitive_res}
\end{figure*}

\textbf{Qualitative Results.}~\figref{fig:qualitive_res} shows the saliency maps produced by the existing most competitive CoSOD models and the proposed~\ourmodel{}. As the results show, the proposed~\ourmodel{} outperforms the previous CoSOD methods in the two perspectives,~\ie{}, the co-segmentation and the accurate detection of the salient object. For example, in the first column of the camera group,~\ourmodel{} can not only accurately segment the salient camera, but can also eliminate the earphone that is salient but not of the right class. On the contrary, other methods neither eliminate objects of wrong classes nor generate highly accurate segmentation maps of the target camera. In addition, the proposed~\ourmodel{} also shows great robustness in defending its prediction from the distraction of surrounding noise objects, which are not salient or not of the right class. For example, the wrench is easily identified by~\ourmodel{} though it is neighboring and similar to the target hammer (see the first example in the hammer group in~\figref{fig:qualitive_res}), the target ax can be accurately segmented, while many surrounding tools (knives, pliers,~\etc{}) are completely excluded (see the last column in the ax group in~\figref{fig:qualitive_res}).

\subsection{Competing Datasets}
\label{sec:competing_datasets}
In order to show the superiority of the proposed~\ourdataset{} dataset for training CoSOD models, experiments are also conducted in which several representative CoSOD models~\cite{GCoNet,GCoNet+,MCCL} and the proposed~\ourmodel{} are trained with different training sets and evaluated on the same test sets. Extensive experiments show that models trained on the proposed~\ourdataset{} dataset can achieve much better performance with fewer training iterations.

\begin{figure*}[t!]
    \centering
    \scriptsize{}
    \begin{overpic}[width=.99\textwidth]{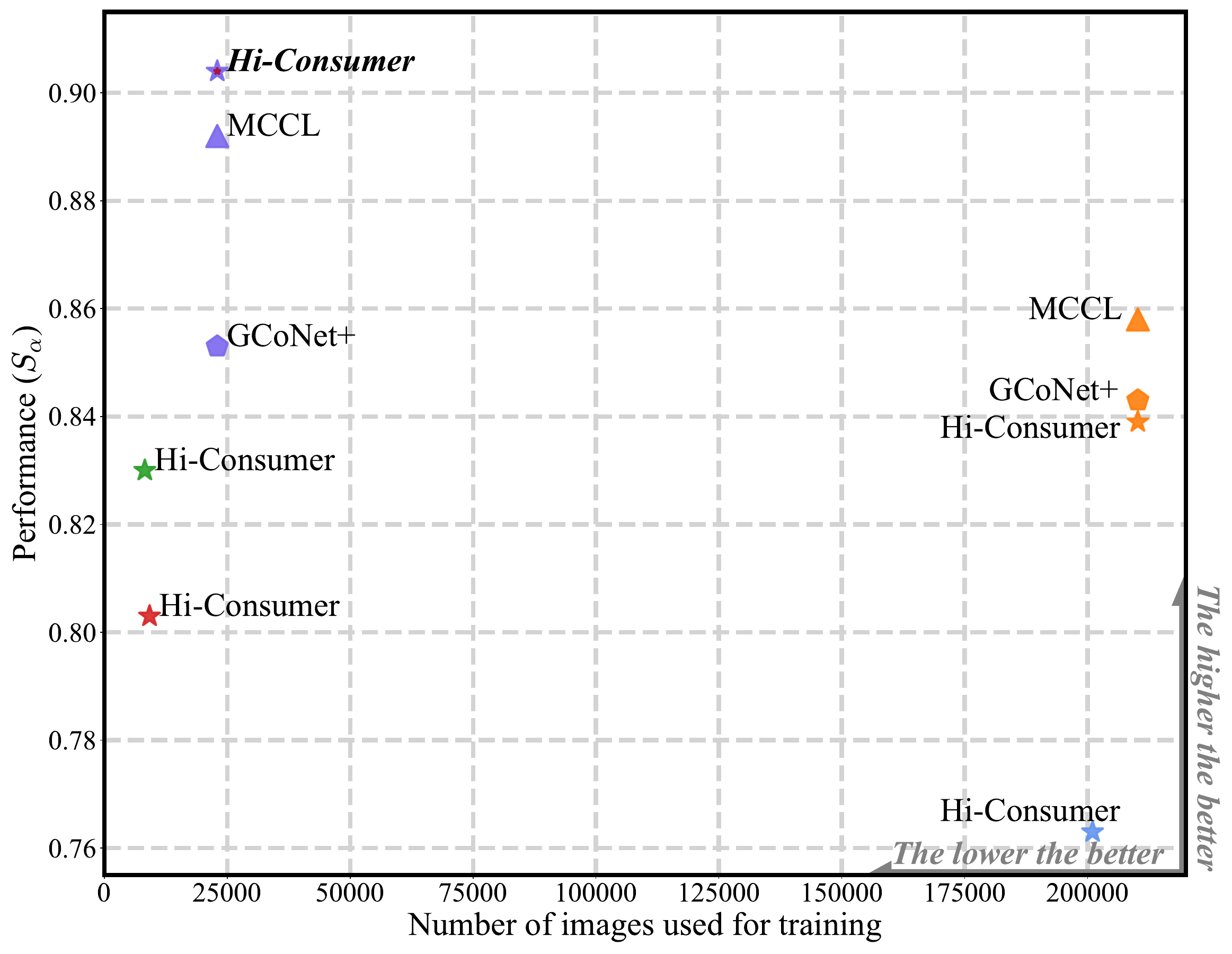}
    \end{overpic}
    \caption{\textit{Quantitative comparisons of the proposed~\ourdataset{}~and other datasets.} The same model trained on different datasets are tagged with the same marker. Different colors represent different datasets, green, red, purple, blue, and orange represent the DUTS\_class, COCO-9k, the proposed~\ourdataset{}, COCO-SEG, and DUTS\_class+COCO-SEG, respectively. Performance is measured by the Smeasure score on the CoSOD3k dataset.}
    \label{fig:dataset_comparison}
\end{figure*}

\begin{table*}[t!]
\begin{center}
\renewcommand{\arraystretch}{1.2}
\renewcommand{\tabcolsep}{1.14mm}
\tiny{}
\caption{\textit{Quantitative comparisons between the proposed~\ourdataset{} dataset and other existing training sets,~\ie{}, DUTS\_class~\cite{GICD}, COCO-9k~\cite{GWD}, and COCO-SEG~\cite{COCO_SEG}.} ``$\uparrow$'' (``$\downarrow$'') means that the higher (lower) is better.} 
\label{tab:dataset_comparison}
\begin{tabular}{r||c|cccc|cccc|cccc}
\hline
 &  & \multicolumn{4}{c|}{CoCA~\cite{GICD}} & \multicolumn{4}{c|}{CoSOD3k~\cite{CoSOD3k}} & \multicolumn{4}{c}{CoSal2015~\cite{CoSal2015}} \\
Method & Training Set & $E_{\xi}^\text{max} \uparrow$ & $S_\alpha \uparrow$ & $F_\beta^\text{ max} \uparrow$ & $\epsilon \downarrow$ & $E_{\xi}^\text{ max} \uparrow$ & $S_\alpha \uparrow$ & $F_\beta^\text{ max} \uparrow$ & $\epsilon \downarrow$ & $E_{\xi}^\text{ max} \uparrow$ & $S_\alpha \uparrow$ & $F_\beta^\text{ max} \uparrow$ & $\epsilon \downarrow$ \\
\hline

\href{https://github.com/ZhengPeng7/MCCL}{MCCL}~\cite{MCCL} & DUTS\_class & 0.763 &  0.697 & 0.574 & 0.120 & 0.881 & 0.831 & 0.817 & 0.079 & 0.895 & 0.890 & 0.901 & 0.64  \\
\href{https://github.com/ZhengPeng7/MCCL}{MCCL}~\cite{MCCL} & COCO-9k & 0.776 &  0.698 & 0.586 & 0.108 & 0.885 & 0.822 & 0.821 & 0.073 & 0.881 & 0.868 & 0.871 & 0.064  \\
\href{https://github.com/ZhengPeng7/MCCL}{MCCL}~\cite{MCCL} & COCO-SEG & 0.785 & 0.708 & 0.585 & 0.113 & 0.893 & 0.842 & 0.825 & 0.071 & 0.886 & 0.879 & 0.882 & 0.058  \\
\rowcolor{mygray}
\href{https://github.com/ZhengPeng7/MCCL}{MCCL}~\cite{MCCL} & ~\ourdataset{} & \textbf{0.787} &  \textbf{0.726} & \textbf{0.605} & \textbf{0.090} & \textbf{0.939} & \textbf{0.892} & \textbf{0.889} & \textbf{0.040} & \textbf{0.952} & \textbf{0.908} & \textbf{0.921} & \textbf{0.035}  \\
\hline

\href{https://github.com/ZhengPeng7/GCoNet_plus}{GCoNet+}~\cite{GCoNet+}	& DUTS\_class & 0.786 & 0.691 & 0.574 & 0.113 & 0.881 & 0.828 & 0.807 & 0.068 & 0.917 & 0.875 & 0.876 & 0.054 \\
\href{https://github.com/ZhengPeng7/GCoNet_plus}{GCoNet+}~\cite{GCoNet+}	& COCO-9k & 0.798 & 0.717 & 0.605 & \textbf{0.098} & 0.877 & 0.819 & 0.796 & 0.075 & 0.902 & 0.853 & 0.857 & 0.073 \\
\href{https://github.com/ZhengPeng7/GCoNet_plus}{GCoNet+}~\cite{GCoNet+}	& COCO-SEG & 0.787 & 0.712 & 0.602 & 0.100 & 0.875 & 0.820 & 0.793 & 0.075 & 0.899 & 0.853 & 0.852 & 0.071 \\
\rowcolor{mygray}
\href{https://github.com/ZhengPeng7/GCoNet_plus}{GCoNet+}~\cite{GCoNet+}	& ~\ourdataset{} & \textbf{0.811} & \textbf{0.725} & \textbf{0.645} & 0.108 & \textbf{0.912} & \textbf{0.853} & \textbf{0.854} & \textbf{0.061} & \textbf{0.930} & \textbf{0.876} & \textbf{0.896} & \textbf{0.053} \\
\hline

\href{https://github.com/ZhengPeng7/CoSINe}{\ourmodel{}} & DUTS\_class & 0.708 & 0.654 & 0.496 & 0.151 & 0.864 & 0.830 & 0.792 & 0.077 & 0.901 & 0.874 & 0.858 & 0.065 \\ 
\href{https://github.com/ZhengPeng7/CoSINe}{\ourmodel{}} & COCO-9k & 0.735 & 0.657 & 0.511 & 0.147 & 0.845 & 0.803 & 0.766 & 0.097 & 0.894 & 0.853 & 0.845 & 0.075 \\ 
\href{https://github.com/ZhengPeng7/CoSINe}{\ourmodel{}} & COCO-SEG & 0.755 & 0.662 & 0.501 & 0.118 & 0.818 & 0.763 & 0.703 & 0.106 & 0.837 & 0.785 & 0.746 & 0.107 \\ 
\rowcolor{mygray}
\href{https://github.com/ZhengPeng7/CoSINe}{\ourmodel{}} & ~\ourdataset{} & \textbf{0.820} & \textbf{0.752} & \textbf{0.647} & \textbf{0.086} & \textbf{0.945} & \textbf{0.904} & \textbf{0.903} & \textbf{0.036} & \textbf{0.957} & \textbf{0.917} & \textbf{0.927} & \textbf{0.030}  \\
\hline
\end{tabular}
\end{center}
\end{table*}

\textbf{Quantitative Results.}

To show the priority of the proposed~\ourdataset{} dataset, several representative methods and the proposed~\ourmodel{} are trained on the~\ourdataset{} and previous training sets, respectively. As shown in~\figref{fig:dataset_comparison}, models trained on~\ourdataset{} can achieve much better results with much fewer images for training, compared to training on previous training sets individually or in combination. This means that the proposed~\ourdataset{} not only can increase the upper bound of the ability of CoSOD models but also can guide them to converge faster.

In addition, the performance of these representative models trained with different training sets is provided. As the results given in~\tabref{tab:dataset_comparison} show, all these models can achieve much better performance on all existing test sets (\ie{}, CoCA~\cite{GICD}, CoSOD3k~\cite{CoSOD3k}, and CoSal2015~\cite{CoSal2015}) when trained with the proposed~\ourdataset{} compared to training with previous training sets.

\textbf{Qualitative Results.}

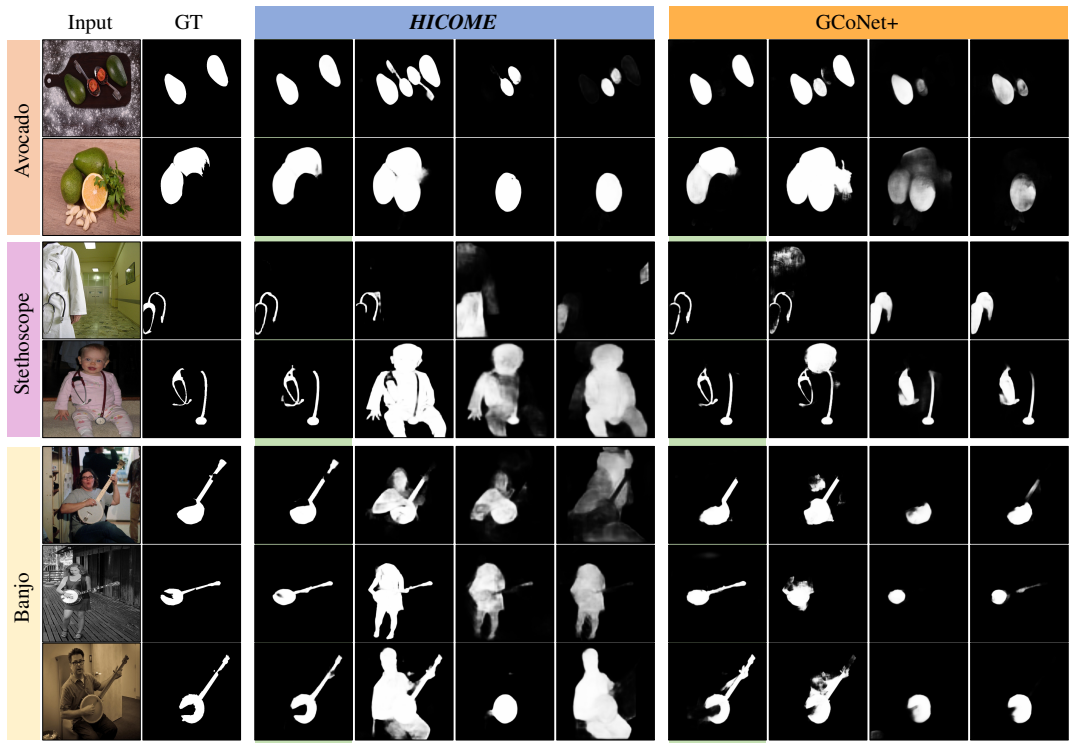
\begin{figure*}[t!]
    \centering
    \scriptsize{}
    \begin{overpic}[width=.99\textwidth]{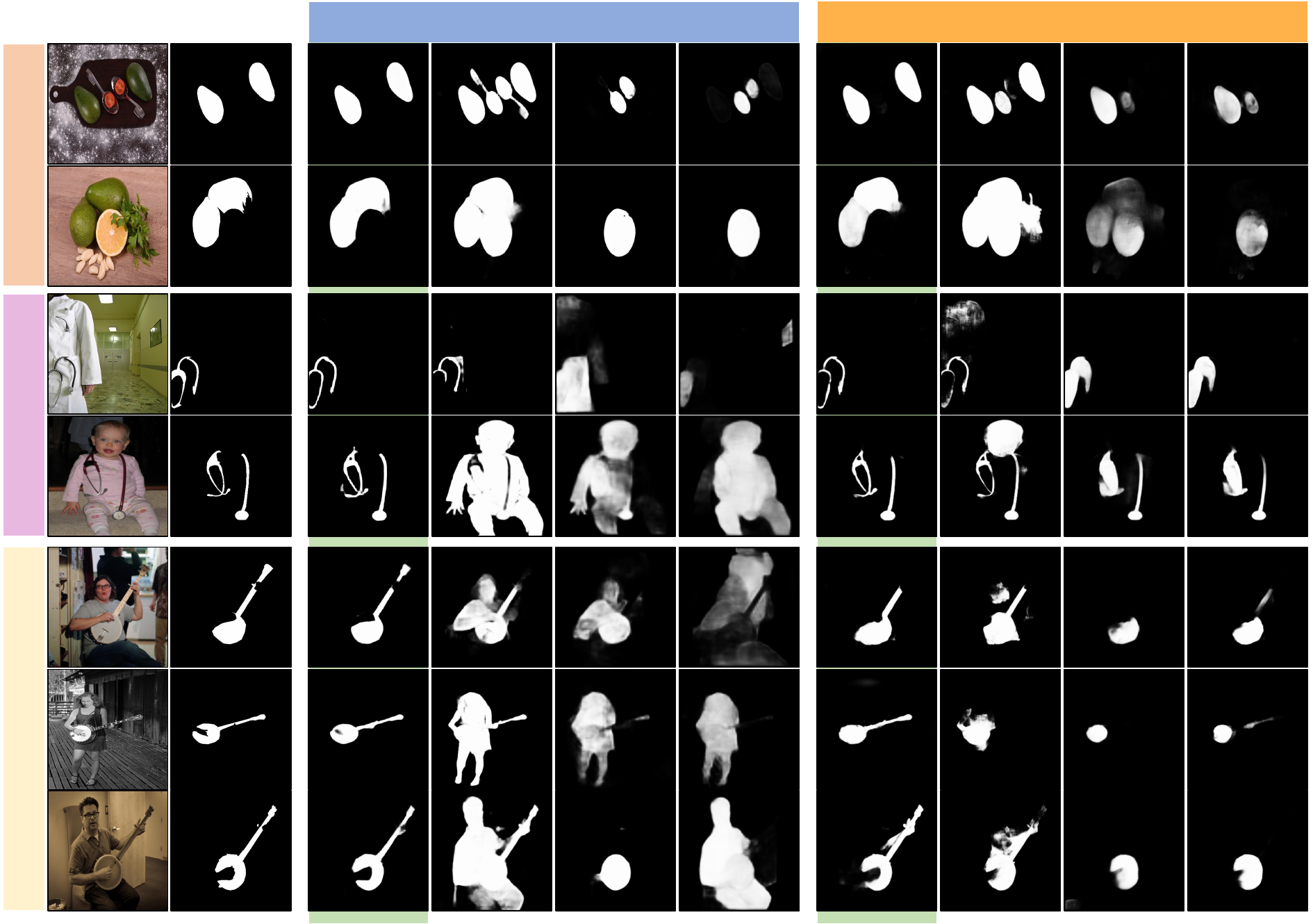}
        \put(38,68){{\textbf{\ourmodel{}}}}
        \put(76,68){{GCoNet+}}
        \put(6,68){{Input}}
        \put(16,68){{GT}}
        \put(1,54){\rotatebox{90}{Avocado}}
        \put(1,33.5){\rotatebox{90}{Stethoscope}}
        \put(1,12){\rotatebox{90}{Banjo}}
    \end{overpic}
    \caption{\textit{The qualitative comparison of the training sets.} Models trained on different training sets,~\ie{},~\ourdataset{}, DUTS\_class, COCO-9k, and COCO-SEG, are given to produce the results, which are listed in the same order in each row.}
    \label{fig:qual_dataset}
\end{figure*}

To compare the proposed~\ourdataset{} dataset with existing training sets and show its superiority, the ablation studies are conducted on the existing SoTA method (GCoNet+~\cite{GCoNet+}) and the proposed~\ourmodel{} trained on a different single dataset,~\ie{}, the proposed~\ourdataset{}, DUTS\_class, COCO-9k, and COCO-SEG, to show how much these models can learn from each dataset.

As shown in~\figref{fig:qual_dataset}, the results produced by the models trained on~\ourdataset{} are much better than others in terms of both the identification of the object category and the quality of the segmentation maps. For example, after training in~\ourdataset{}, ~\ourmodel{} and GCoNet+ can both accurately identify and segment the targets in the avocado group, as shown in the two columns with a green background. On the contrary, when they are trained with the other three datasets, neither eliminate the non-avocado object nor provide the accurate segmentation maps of the target avocados.

In addition, previous training sets often lead CoSOD models to sub-optimal in the CoSOD task, as discussed in~\cite{GCoNet+} in detail. Specifically, on the one hand, models trained on CoSal2015 tend to act better on detecting the salient objects while relatively worse on finding the common class objects. On the other hand, models trained on COCO-9k or COCO-SEG tend to be stronger in detecting the common class objects while weaker in segmenting the salient objects. On the contrary, the proposed~\ourdataset{} can guide CoSOD models to the balance of strong ability to detect both salient objects and common class objects. As shown in~\tabref{tab:dataset_comparison}, models trained on~\ourdataset{} do not show performance degradation in any single test set.

\clearpage

\section{Potential Applications}
\label{sec:applications}
The potential to utilize the extracted co-saliency maps to produce high quality segmentation masks for related downstream image processing tasks is shown as follows:

\begin{figure}[t!]
  \centering
    \begin{overpic}[width=.7\linewidth]{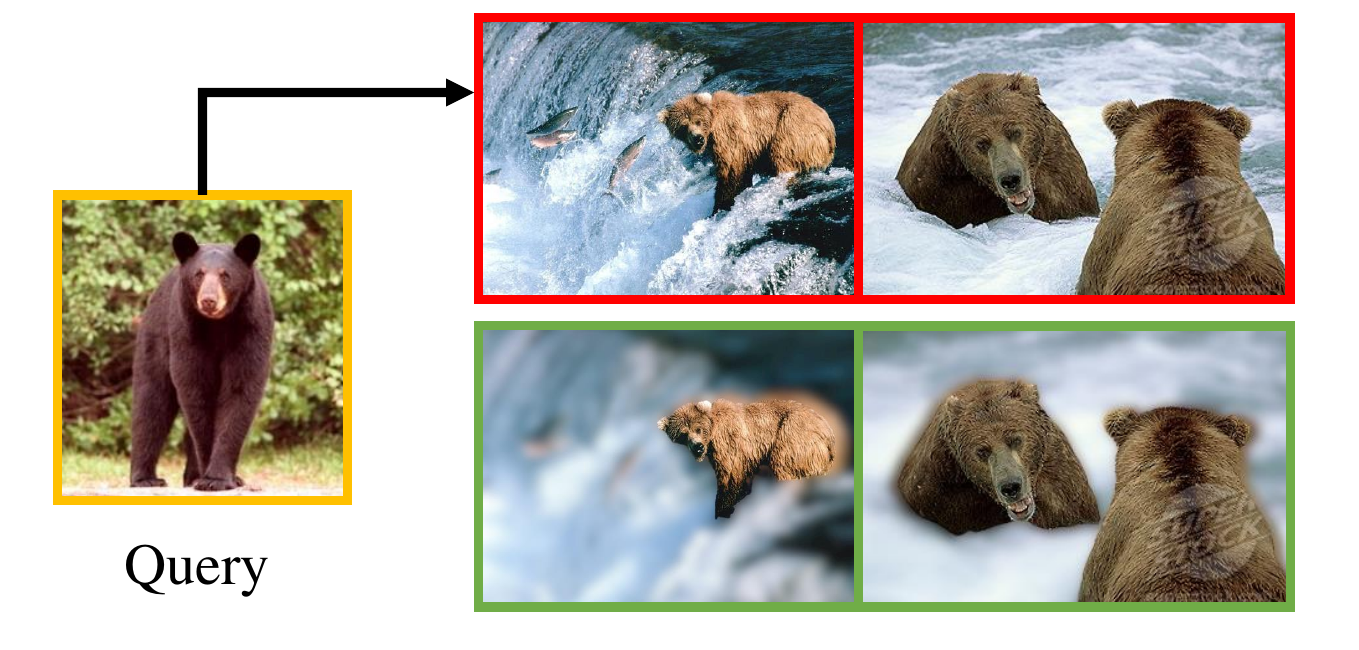}
    \end{overpic}
   \caption{
   \textit{Application \#1: Query-based Object Retrieval.} Objects (``Bear'') in the gallery images are obtained by the proposed~\ourmodel{}.
   }\label{fig:app_1}
\end{figure}

\textbf{Application \#1: Query-based Object Retrieval.} With the goal of finding target objects with certain cues (the query image), the neural network-based application is provided here, which is a good tool for users to search for similar images in the big gallery. As shown in~\figref{fig:app_1}, given the query image in the yellow box, the proposed approach can easily detect objects of the same category in the gallery images surrounded by red boxes. Finally, the pixels of the targets are segmented and the rest of the images are blurred, as shown in the green box.

\begin{figure}[t!]
\begin{center}
\includegraphics[width=.7\linewidth]{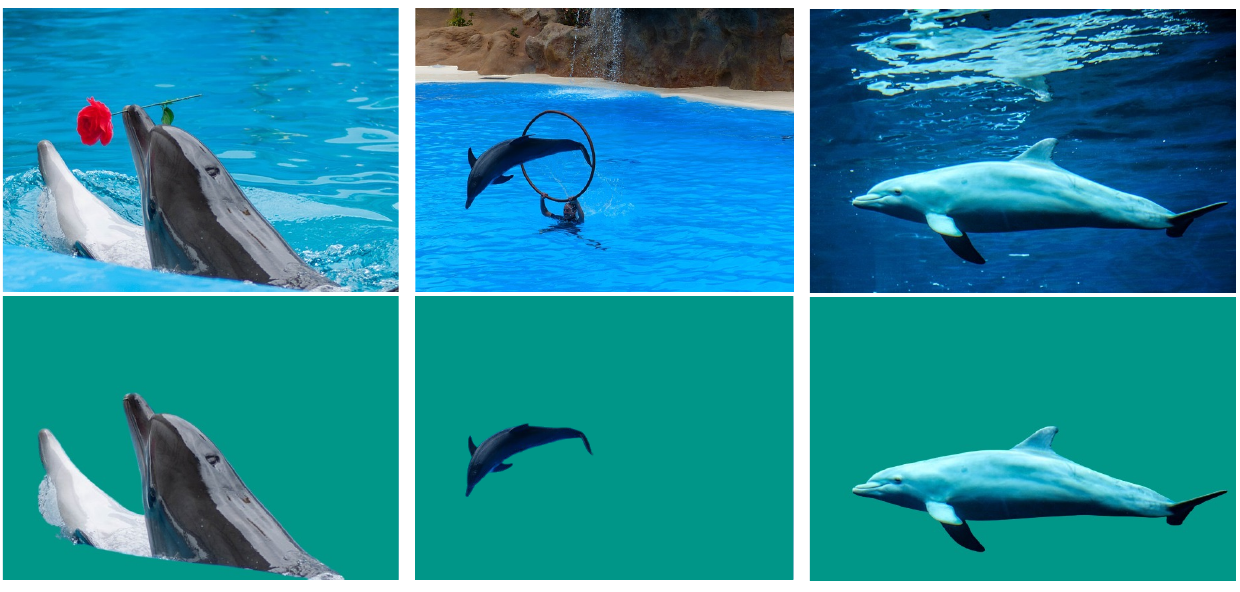}
\end{center}
\vspace{-10pt}
\caption{\textit{Application \#2: Content aware object co-segmentation.} The visual results (``Dolphin'') are obtained by the proposed~\ourmodel{}.}
\label{fig:app_2}
\end{figure}

\textbf{Application \#2: Content-Aware Co-Segmentation.}
Co-saliency maps have been widely used in image pre-processing tasks. Taking the unsupervised object segmentation in the implementation as an example, a group of images is first found by keywords on the Internet. Then, the proposed~\ourmodel{}~is applied to generate co-saliency maps. Finally, the salient objects of the specific group can be extracted with the co-saliency maps. Following~\cite{Cheng2011GlobalCB}, GrabCut~\cite{grabcut} can be used to obtain the final segmentation results. The adaptive threshold~\cite{Peng2014RGBDSO} is chosen here to initialize GrabCut for the binary version of the saliency maps. As shown in~\figref{fig:app_2}, the proposed method works well in the content-aware object co-segmentation task, which should benefit existing E-commerce applications in the background replacement.

\begin{figure}[t!]
\begin{center}
\includegraphics[width=.7\linewidth]{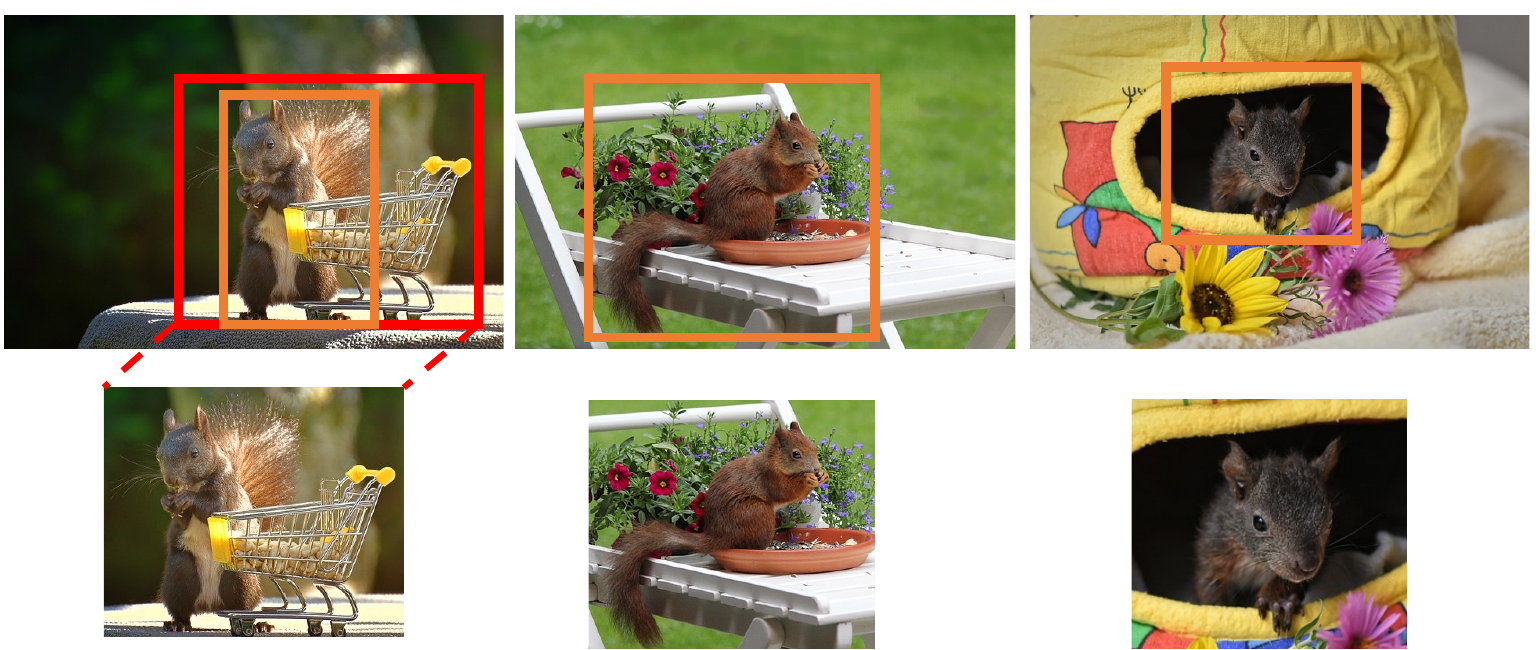}
\end{center}
\vspace{-10pt}
\caption{\textit{Application \#3.} Co-localization based automatic thumbnails (``Squirrel'') produced by the proposed~\ourmodel{}.}
\label{fig:app_3}
\end{figure}

\textbf{Application \#3: Automatic Thumbnails.}
The idea of thumbnails of paired images is from~\cite{CoSOD_trad_align}. With the same goal\footnote{Jacobs~\etal{}’s work~\cite{CoSOD_trad_align} is limited to the case of image pairs}, a CNN-based application of photographic triage is introduced, which is valuable for sharing images on the website. As \figref{fig:app_3} shows, the orange box is generated by the saliency maps obtained from~\ourmodel{}. They can also be scaled with the orange box to get the larger red one. Finally, the collection-aware crop technique~\cite{CoSOD_trad_align} can be adapted to produce the results shown in the second row.

\clearpage

\section{Co-Salient Object Detection in the Future}
\label{sec:CoSOD_future}
Although Co-Salient Object Detection has been driven to develop in recent years, it still has a great potential to be improved with the latest techniques and to improve other investigations. Typical examples among them are listed here to provide further direction for CoSOD in the future.

\begin{figure}[t!]
\begin{center}
\includegraphics[width=.99\linewidth]{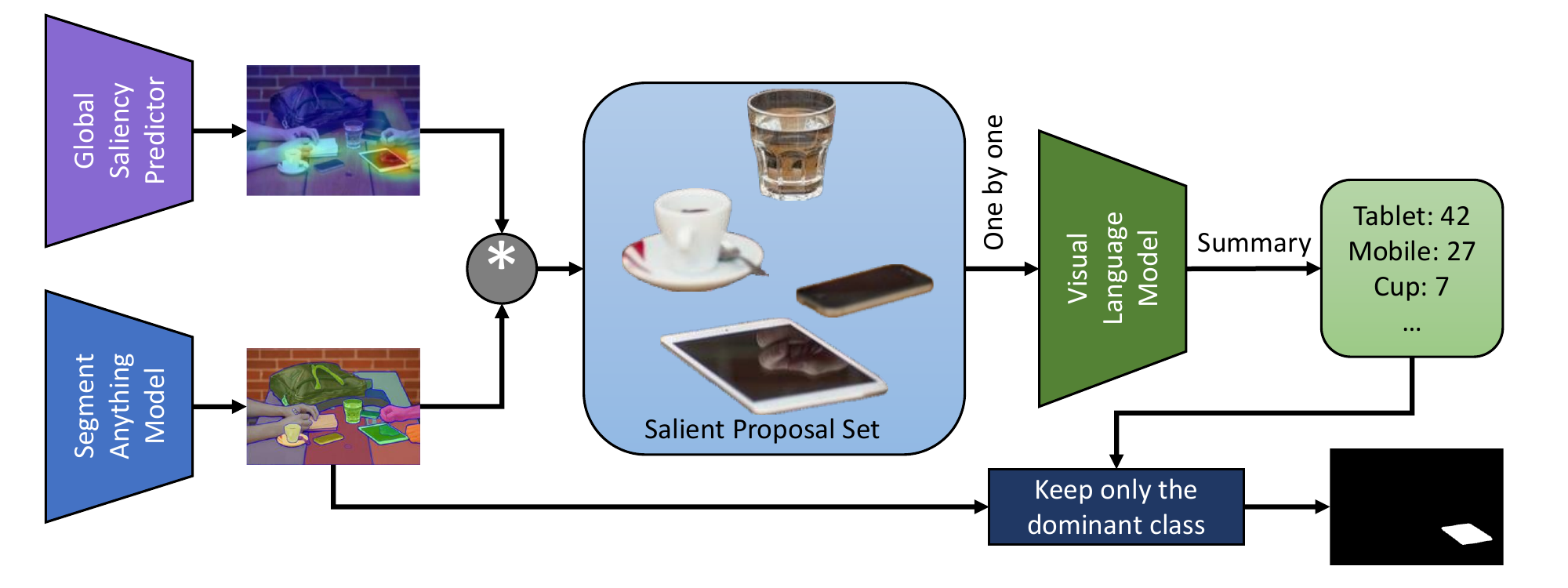}
\end{center}
\caption{\textit{New paradigm of Co-Salient Object Detection.} This new paradigm utilizes the power of SAM and visual language models as the query extractor and classifier, respectively.}
\label{fig:CoSOD_new_paradigm}
\end{figure}

\textbf{Improvement on CoSOD \#1: Query extraction with SAM.}
Segment anything model (SAM)~\cite{SAM} has raised a wide range of interest in the computer vision community since it was released~\cite{SAM_medic,SAM_adapter,SAM_3d}. On the basis of its strong ability of general segmentation, SAM can be used as a class-agnostic query extractor. In the earlier works in the area of Salient Object Detection, splitting the image into computational units is a widely used approach to extract local features (refer to~\secref{sec:CoSOD_methods}). However, the handcrafted units can only cover a small local patch in the object instead of the whole object. On the contrary, SAM can produce masks of objects with different objectiveness levels. With the support of SAM and visual language models (VLMs), CoSOD can be rethought as a new paradigm: 1). Predict the saliency maps of the images; 2). Obtain object proposals with SAM; 3). Multiply the saliency maps and the corresponding proposal maps to eliminate non-salient proposals. 4). Use VLMs (\eg{}, CLIP~\cite{CLIP}, BLIP~\cite{BLIP,BLIP2}) to generate the class names of each proposal and keep the proposals with the dominant class as the final co-saliency maps. 5). Refinement on the maps with information from original images and fine-tuning the architecture with adapters are also possible to make further improvements. The entire paradigm is illustrated as shown in~\figref{fig:CoSOD_new_paradigm} in detail.

\begin{figure}[t!]
\begin{center}
\includegraphics[width=.99\linewidth]{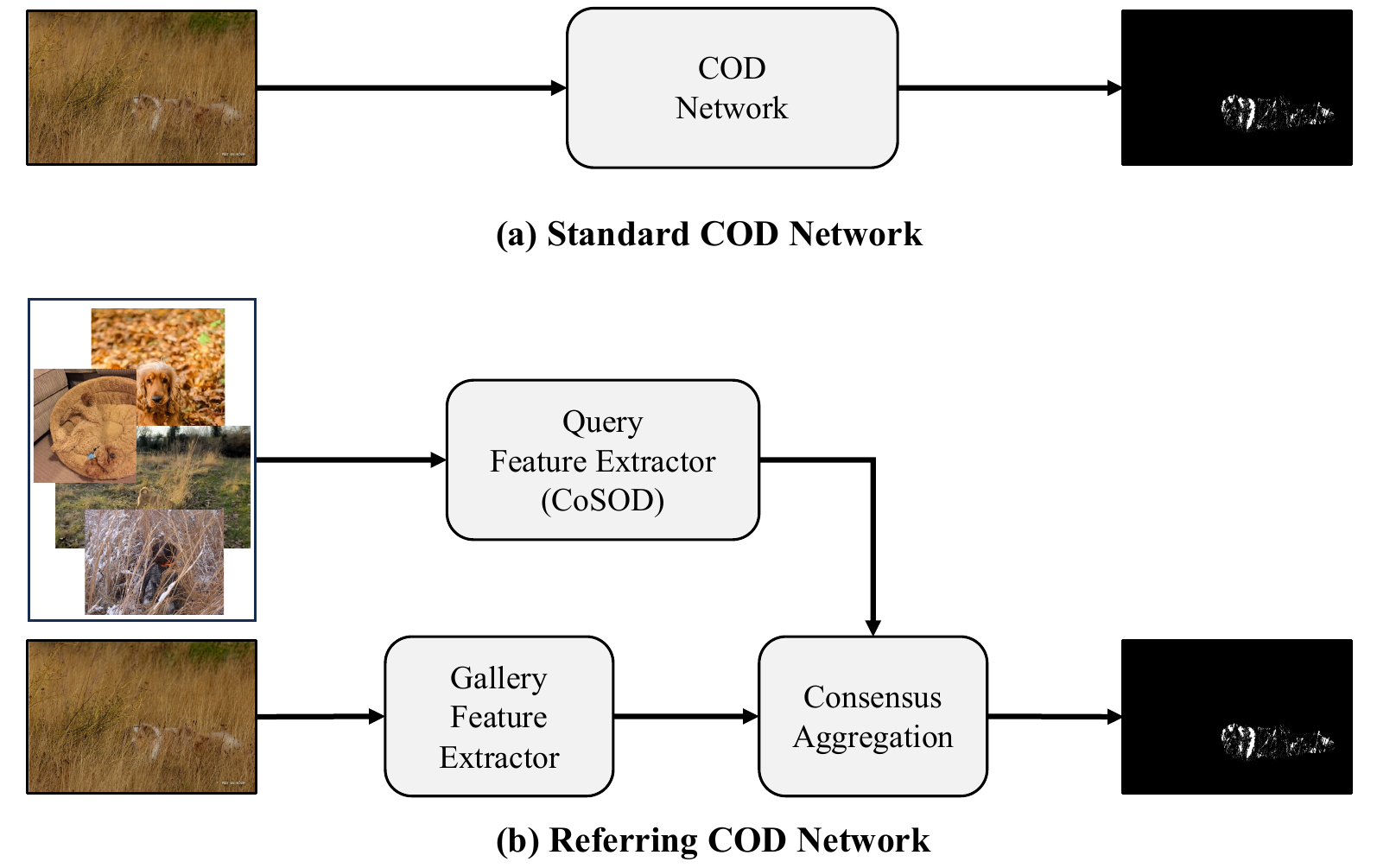}
\vspace{-5mm}
\end{center}
\caption{\textit{The assistance of CoSOD in COD task.} The common information obtained by CoSOD models can be applied to contribute to finding the camouflaged objects with the same class of queries.}
\label{fig:ref_COD}
\end{figure}

\textbf{Improvement by CoSOD \#2: CoSOD-induced Camouflaged Object Detection.}
Due to the stronger performance of the latest CoSOD models in finding objects of common classes, they have also been applied to other higher-difficult segmentation tasks as a clue~\cite{referring_COD}. For example, camouflaged objects are difficult to detect in natural scenes with a vanilla end-to-end style. However, when given a group of query images containing salient objects with the target class, CoSOD can help the COD model discover camouflaged objects in gallery images of the same class, as shown in~\figref{fig:ref_COD}.

\section{Conclusions}
\label{sec:conclusions}
In this thesis, the main remaining problems in the Co-Salient Object Detection (CoSOD) dataset are raised and several methods are explored to investigate the essence of better discriminative features in finding common objects in a given group of images. 

In terms of the dataset, number of groups is much more important than number of images in total in a CoSOD training set. Existing CoSOD training sets only have limited groups (50$~\sim{}$300), which may heavily suppress the effect of metric learning. To overcome this problem, a new CoSOD training set named~\ourdataset{} is introduced in this thesis, which contains 919 groups of images. CoSOD models trained on the proposed~\ourdataset{} can achieve better performance on the three widely used CoSOD benchmarks compared with the same models trained on existing training sets.

In terms of the method, some of the presented metric learning methods tailored for CoSOD have shown the rationality to drive the metric learning methods to learn the more precise intra-group compactness and inter-group separability (\eg{}, the Group-based Symmetry Triplet (GST) loss proposed in Publication I, and the memory-aided contrastive consensus presented in Publication II). Besides, to make full of the large group number in the proposed~\ourdataset{} dataset, this thesis proposes a novel approach for detecting co-salient objects named~\ourmodel{}. The proposed~\ourmodel{} is a lightweight model with three original components organically combined together,~\ie{}, Hierarchical Consensus Fusion (\textbf{HCF}), Spatial Increment Attention (\textbf{SIA}), and Instance-Aware Contrastive Consensus Learning (\textbf{IACCL}).

Qualitative and quantitative experiments demonstrate the state-of-the-art performance of the proposed~\ourmodel{} and the superiority of the~\ourdataset{} dataset.


In addition, well-trained~\ourmodel{} is shown to be easily used in many relevant CV tasks and practical applications, such as co-segmentation and image retrieval.

Finally, this thesis solves an urgent problem in CoSOD (class insufficiency) and proposes a novel network focusing on this problem. Further research on CoSOD and research with CoSOD has also been given for more inspiration. Hope this work encourages other researchers to rethink tackling the CoSOD task in a better direction.

\clearpage

{
\bibliographystyle{IEEEtran}

}

\clearpage

\end{document}

%% file: cover.tex
\degreeprogram{ICT Innovation}

\major{Visual Computing and Communication}

\univdegree{MSc}

\thesisauthor{Peng Zheng}

\thesistitle{Discriminative Consensus Mining with A Thousand Groups for More Accurate Co-Salient Object Detection}
\thesissubtitle{}

\place{Espoo}

\date{25 Sep 2023}

\supervisor{Senior University Lecture\ Jorma Laaksonen}

\advisor{Senior University Lecture\ Jorma Laaksonen}

\uselogo{''}
\copyrighttext{\noexpand\textcopyright\ \number\year. This work is 
	licensed under a Creative Commons "Attribution-NonCommercial-ShareAlike 4.0 
	International" (BY-NC-SA 4.0) license.}{\noindent\textcopyright\ \number
	\year \ \doclicenseThis}

%% file: head.tex

\thesisabstract{%
The abstract is a short description of the essential contents of the thesis:
what was studied and how and what were the main findings. For a Finnish thesis,
the abstract should be written in both Finnish and English; for a Swedish
thesis, in Swedish and English. The abstracts for English theses written by
Finnish or Swedish speakers should be written in English and either in Finnish
or in Swedish, depending on the student’s language of basic education. Students
educated in languages other than Finnish or Swedish write the abstract only in
English. Students may include a second or third abstract in their native
language, if they wish. 
The abstract text of this thesis is written on the readable abstract page as
well as into the pdf file's metadata via the thesisabstract macro (see the 
comment in the TeX file). Write here the text that goes into the metadata. The 
metadata cannot contain special characters, linebreak or paragraph break 
characters, so these must not be used here. If your abstract does not contain 
special characters and it does not require paragraphs, you may take advantage of
the abstracttext macro (see the comment in the TeX file below). Otherwise, the 
metadata abstract text must be identical to the text on the abstract page.
}



\begin{abstractpage}[english]
Co-Salient Object Detection (CoSOD) is a rapidly growing task, extended from Salient Object Detection (SOD) and Common Object Segmentation (Co-Segmentation). It is aimed at detecting the co-occurring salient object in the given image group. Many effective approaches have been proposed on the basis of existing datasets. However, there is still no standard and efficient training set in CoSOD, which makes it chaotic to choose training sets in the recently proposed CoSOD methods. First, the drawbacks of existing training sets in CoSOD are analyzed in a comprehensive way, and potential improvements are provided to solve existing problems to some extent. In particular, in this thesis, a new CoSOD training set is introduced, named Co-Saliency of ImageNet (\ourdataset{}) dataset. The proposed~\ourdataset{} is the largest number of groups among all existing CoSOD datasets. The images obtained here span a wide variety in terms of categories, object sizes,~\etc{}. In experiments, models trained on~\ourdataset{} can achieve significantly better performance with fewer images compared to all existing datasets. Second, to make the most of the proposed~\ourdataset{}, a novel CoSOD approach named Hierarchical Instance-aware COnsensus MinEr (~\ourmodel{}) is proposed, which efficiently mines the consensus feature from different feature levels and discriminates objects of different classes in an object-aware contrastive way. As extensive experiments show, the proposed~\ourmodel{} achieves SoTA performance on all the existing CoSOD test sets. Several useful training tricks suitable for training CoSOD models are also provided. Third, practical applications are given using the CoSOD technique to show the effectiveness. Finally, the remaining challenges and potential improvements of CoSOD are discussed to inspire related work in the future. The source code, the dataset, and the online demo will be publicly available at \url{https://github.com/ZhengPeng7/CoSINe}.
\end{abstractpage}

\dothesispagenumbering{}

\mysection{Preface}
The work presented in this thesis is a follow-up work of my previous investigation on the Co-Salient Object Detection task, which is part of my works at the Inception Institute of Artificial Intelligence (IIAI) from April 2021 to December 2021 and at the Mohamed bin Zayed University of Artificial Intelligence (NBZUAI) from January 2022 to September 2022.

First, I would like to express my greatest gratitude to my supervisor, Jorma Laaksonen, who has given me quite a lot of patience for my procrastination and valuable suggestions on how to do this thesis and make it better. Thanks a lot for your support and guidance. I am also grateful to the coordinator of the EIT program at Aalto University (Päivi and Anna) and Professor Nicu Sebe at the University of Trento. I experienced a lot of mental stress when everything went bad during the COVID-19 period in Europe and was very resistant to continuing my school studies. But all of you really helped me to get back to finish my studies.

During my years of research, I was very lucky to meet some really wonderful friends and mentors who have given me selfless and priceless guidance, chance, and support in my growth. In particular, I would like to thank Prof. Xiaogang Cheng, who provided me with a door for entering computer vision. I also want to give my sincere thanks to Prof. Jie Qin, who has helped me not only in my academic work, but also in my life in Abu Dhabi. I want to give many thanks to Dr. Deng-Ping Fan, who has helped me in doing my academic work and living a more wonderful life. I would like to thank a number of my friends, elder brothers, and mentors on my journey from Espoo to Abu Dhabi to Hangzhou to Shanghai in alphabetical order: Dehong Gao, Geng Chen, Huan Xiong, Nian Liu, Tian-Zhu Xiang, Yichao Yan. Without you, I cannot do what I have done.

I also want to thank my family for their unconditional love, who gave me full support in trying everything I wanted and provided me with a suggestion instead of asking me to do it or not. I also want to thank my girlfriend, Leying Li, for her company during my blue time.

\vspace{2cm}
Shanghai, 25 September 2023\\

\vspace{5mm}
{\hfill Peng Zheng \hspace{1cm}}

\newpage

\thesistableofcontents

\mysection{Symbols and Abbreviations}

\subsection*{Symbols}

\begin{tabular}{ll}
$\mathcal{F}$        & Feature  \\
$\mathcal{G}$        & Original images \\
$\mathcal{M}$        & Predicted maps \\
$\uparrow$          & The larger is better \\
$\downarrow$          & The lower is better \\
\end{tabular}



\subsection*{Abbreviations}

\begin{tabular}{ll}
BCE         & Binary Cross Entropy \\
CNN         & Convolution Neural Network \\
COD         & Camouflaged Object Detection \\
CoSeg       & Common Object Segmentation \\
CoSOD       & Co-Salient Object Detection \\
CV          & Computer Vision \\
FPN         & Feature Pyramid Network \\
GT          & Ground Truth \\
HCF         & Hierarchical Consensus Fusion \\
MHA         & Multi-Head Attention \\
IACCL       & Instance-Aware Contrastive Consensus Learning \\
SIA         & Spatial Increment Attention \\
SOD         & Salient Object Detection \\
SVM         & Support Vector Machine \\
\end{tabular}

\cleardoublepage


%% file: main.bbl
\begin{thebibliography}{100}
\providecommand{\url}[1]{#1}
\csname url@samestyle\endcsname
\providecommand{\newblock}{\relax}
\providecommand{\bibinfo}[2]{#2}
\providecommand{\BIBentrySTDinterwordspacing}{\spaceskip=0pt\relax}
\providecommand{\BIBentryALTinterwordstretchfactor}{4}
\providecommand{\BIBentryALTinterwordspacing}{\spaceskip=\fontdimen2\font plus
\BIBentryALTinterwordstretchfactor\fontdimen3\font minus \fontdimen4\font\relax}
\providecommand{\BIBforeignlanguage}[2]{{%
\expandafter\ifx\csname l@#1\endcsname\relax
\typeout{** WARNING: IEEEtran.bst: No hyphenation pattern has been}%
\typeout{** loaded for the language `#1'. Using the pattern for}%
\typeout{** the default language instead.}%
\else
\language=\csname l@#1\endcsname
\fi
#2}}
\providecommand{\BIBdecl}{\relax}
\BIBdecl

\bibitem{GCoNet}
Q.~Fan, D.-P. Fan, H.~Fu, C.-K. Tang, L.~Shao, and Y.-W. Tai, ``Group collaborative learning for co-salient object detection,'' in \emph{{IEEE / CVF Computer Vision and Pattern Recognition Conference}}, 2021, pp. 12\,283--12\,293.

\bibitem{GCoNet+}
P.~Zheng, H.~Fu, D.-P. Fan, Q.~Fan, J.~Qin, Y.-W. Tai, C.-K. Tang, and L.~Van~Gool, ``{GCoNet+}: A stronger group collaborative co-salient object detector,'' \emph{IEEE Transactions on Pattern Analysis and Machine Intelligence}, pp. 1--18, 2023.

\bibitem{CoSOD3k}
D.-P. Fan, T.~Li, Z.~Lin, G.-P. Ji, D.~Zhang, M.-M. Cheng, H.~Fu, and J.~Shen, ``Re-thinking co-salient object detection,'' \emph{{IEEE Transactions on Pattern Analysis and Machine Intelligence}}, pp. 1--1, 2021.

\bibitem{MCCL}
P.~Zheng, J.~Qin, S.~Wang, T.-Z. Xiang, and H.~Xiong, ``Memory-aided contrastive consensus learning for co-salient object detection,'' in \emph{{AAAI Conference on Artificial Intelligence}}, 2023.

\bibitem{CBCS}
H.~Fu, X.~Cao, and Z.~Tu, ``Cluster-based co-saliency detection,'' \emph{{IEEE Transactions on Image Process.}}, vol.~22, no.~10, pp. 3766--3778, 2013.

\bibitem{CoSOD_trad_align}
D.~E. Jacobs, D.~B. Goldman, and E.~Shechtman, ``{Cosaliency: Where people look when comparing images},'' in \emph{ACM symposium on User interface software and technology}, 2010, pp. 219--228.

\bibitem{CoSal2015}
D.~Zhang, J.~Han, C.~Li, J.~Wang, and X.~Li, ``Detection of co-salient objects by looking deep and wide,'' \emph{{International Journal of Computer Vision}}, vol. 120, no.~2, pp. 215--232, 2016.

\bibitem{GICD}
Z.~Zhang, W.~Jin, J.~Xu, and M.-M. Cheng, ``Gradient-induced co-saliency detection,'' in \emph{{European Conference on Computer Vision}}, 2020, pp. 455--472.

\bibitem{hsu2019deepco3}
K.-J. Hsu, Y.-Y. Lin, and Y.-Y. Chuang, ``{DeepCO3}: Deep instance co-segmentation by co-peak search and co-saliency detection,'' in \emph{{IEEE / CVF Computer Vision and Pattern Recognition Conference}}, 2019, pp. 8846--8855.

\bibitem{wang2019no}
X.~Wang, X.~Liang, B.~Yang, and F.~W. Li, ``No-reference synthetic image quality assessment with convolutional neural network and local image saliency,'' \emph{Computational Visual Media}, vol.~5, no.~2, pp. 193--208, 2019.

\bibitem{DUTS}
L.~Wang, H.~Lu, Y.~Wang, M.~Feng, D.~Wang, B.~Yin, and X.~Ruan, ``Learning to detect salient objects with image-level supervision,'' in \emph{{IEEE / CVF Computer Vision and Pattern Recognition Conference}}, 2017, pp. 3796--3805.

\bibitem{COCO_SEG}
C.~Wang, Z.~Zha, D.~Liu, and H.~Xie, ``Robust deep co-saliency detection with group semantic,'' in \emph{{AAAI Conference on Artificial Intelligence}}, 2019, pp. 8917--8924.

\bibitem{GWD}
L.~Wei, S.~Zhao, O.~E.~F. Bourahla, X.~Li, and F.~Wu, ``Group-wise deep co-saliency detection,'' in \emph{{International Joint Conference on Artificial Intelligence}}, 2017, pp. 3041--3047.

\bibitem{COCO}
T.-Y. Lin, M.~Maire, S.~J. Belongie, J.~Hays, P.~Perona, D.~Ramanan, P.~Doll{\'a}r, and C.~L. Zitnick, ``Microsoft {COCO}: Common objects in context,'' in \emph{{European Conference on Computer Vision}}, 2014, pp. 740--755.

\bibitem{CoADNet}
Q.~Zhang, R.~Cong, J.~Hou, C.~Li, and Y.~Zhao, ``{CoADNet}: Collaborative aggregation-and-distribution networks for co-salient object detection,'' \emph{{Advances in Neural Information Processing Systems}}, pp. 6959--6970, 2020.

\bibitem{CADC}
N.~Zhang, J.~Han, N.~Liu, and L.~Shao, ``Summarize and search: Learning consensus-aware dynamic convolution for co-saliency detection,'' in \emph{{IEEE / CVF International Conference on Computer Vision}}, 2021, pp. 4167--4176.

\bibitem{Cheng2011GlobalCB}
M.-M. Cheng, G.-X. Zhang, N.~J. Mitra, X.~Huang, and S.~Hu, ``Global contrast based salient region detection,'' in \emph{{IEEE / CVF Computer Vision and Pattern Recognition Conference}}, 2011, pp. 409--416.

\bibitem{SOD_att1}
X.~Zhang, T.~Wang, J.~Qi, H.~Lu, and G.~Wang, ``Progressive attention guided recurrent network for salient object detection,'' in \emph{{IEEE / CVF Computer Vision and Pattern Recognition Conference}}, 2018, pp. 714--722.

\bibitem{SOD_trad1}
H.~Jiang, Z.~Yuan, M.-M. Cheng, Y.~Gong, N.~Zheng, and J.~Wang, ``Salient object detection: A discriminative regional feature integration approach,'' \emph{{International Journal of Computer Vision}}, vol. 123, pp. 251--268, 2013.

\bibitem{SOD_op1}
L.~Wang, H.~Lu, X.~Ruan, and M.-H. Yang, ``Deep networks for saliency detection via local estimation and global search,'' in \emph{{IEEE / CVF Computer Vision and Pattern Recognition Conference}}, 2015, pp. 3183--3192.

\bibitem{VST}
N.~Liu, N.~Zhang, K.~Wan, J.~Han, and L.~Shao, ``Visual saliency transformer,'' in \emph{{IEEE / CVF International Conference on Computer Vision}}, 2021, pp. 4702--4712.

\bibitem{SOD_RGB-D}
D.~P. Kingma and J.~Ba, ``Promoting saliency from depth: Deep unsupervised rgb-d saliency detection,'' in \emph{{International Conference on Learning Representations}}, 2022.

\bibitem{SOD_patch1}
G.~Li and Y.~Yu, ``Visual saliency based on multiscale deep features,'' in \emph{{IEEE / CVF Computer Vision and Pattern Recognition Conference}}, 2015, pp. 5455--5463.

\bibitem{SOD_patch2}
R.~Zhao, W.~Ouyang, H.~Li, and X.~Wang, ``Saliency detection by multi-context deep learning,'' in \emph{{IEEE / CVF Computer Vision and Pattern Recognition Conference}}, 2015, pp. 1265--1274.

\bibitem{SOD_patch3}
G.~Lee, Y.-W. Tai, and J.~Kim, ``Deep saliency with encoded low level distance map and high level features,'' in \emph{{IEEE / CVF Computer Vision and Pattern Recognition Conference}}, 2016, pp. 660--668.

\bibitem{SOD_patch4}
S.~He, R.~W.~H. Lau, W.~Liu, Z.~Huang, and Q.~Yang, ``{SuperCNN}: A superpixelwise convolutional neural network for salient object detection,'' \emph{{International Journal of Computer Vision}}, vol. 115, pp. 330--344, 2015.

\bibitem{SOD_op2}
J.~Zhang, S.~Sclaroff, Z.~L. Lin, X.~Shen, B.~L. Price, and R.~Mech, ``Unconstrained salient object detection via proposal subset optimization,'' in \emph{{IEEE / CVF Computer Vision and Pattern Recognition Conference}}, 2016, pp. 5733--5742.

\bibitem{SOD_op3}
J.~Kim and V.~Pavlovic, ``A shape-based approach for salient object detection using deep learning,'' in \emph{{European Conference on Computer Vision}}, 2016, pp. 455--470.

\bibitem{SOD_clustering}
G.~Shin, S.~Albanie, and W.~Xie, ``Unsupervised salient object detection with spectral cluster voting,'' in \emph{{IEEE / CVF Computer Vision and Pattern Recognition Conference Workshop}}, 2022, pp. 3971--3980.

\bibitem{SOD_trad3}
F.~Perazzi, P.~Kr{\"a}henb{\"u}hl, Y.~Pritch, and A.~Sorkine-Hornung, ``Saliency filters: Contrast based filtering for salient region detection,'' in \emph{{IEEE / CVF Computer Vision and Pattern Recognition Conference}}, 2012, pp. 733--740.

\bibitem{grabcut}
C.~Rother, V.~Kolmogorov, and A.~Blake, ``{"GrabCut"}: interactive foreground extraction using iterated graph cuts,'' \emph{{ACM Transactions on Graphics}}, vol.~23, no.~3, pp. 309--314, 2004.

\bibitem{SOD_trad2}
X.~Li, H.~Lu, L.~Zhang, X.~Ruan, and M.-H. Yang, ``Saliency detection via dense and sparse reconstruction,'' in \emph{{IEEE / CVF International Conference on Computer Vision}}, 2013, pp. 2976--2983.

\bibitem{FCN}
J.~Long, E.~Shelhamer, and T.~Darrell, ``Fully convolutional networks for semantic segmentation,'' in \emph{{IEEE / CVF Computer Vision and Pattern Recognition Conference}}, 2015, pp. 3431--3440.

\bibitem{EGNet}
J.-X. Zhao, J.~Liu, D.-P. Fan, Y.~Cao, J.~Yang, and M.-M. Cheng, ``{EGNet}: Edge guidance network for salient object detection,'' in \emph{{IEEE / CVF International Conference on Computer Vision}}, 2019, pp. 8778--8787.

\bibitem{BASNet}
X.~Qin, Z.~Zhang, C.~Huang, C.~Gao, M.~Dehghan, and M.~Jagersand, ``{BASNet}: Boundary-aware salient object detection,'' in \emph{{IEEE / CVF Computer Vision and Pattern Recognition Conference}}, 2019, pp. 7479--7489.

\bibitem{PoolNet}
J.~Liu, Q.~Hou, M.-M. Cheng, J.~Feng, and J.~Jiang, ``A simple pooling-based design for real-time salient object detection,'' in \emph{{IEEE / CVF Computer Vision and Pattern Recognition Conference}}, 2019, pp. 3912--3921.

\bibitem{UNet}
O.~Ronneberger, P.~Fischer, and T.~Brox, ``{U-Net}: Convolutional networks for biomedical image segmentation,'' in \emph{{Medical Image Computing and Computer Assisted Interventions}}, 2015, pp. 234--241.

\bibitem{FPN}
T.-Y. Lin, P.~Doll{\'a}r, R.~B. Girshick, K.~He, B.~Hariharan, and S.~J. Belongie, ``Feature pyramid networks for object detection,'' in \emph{{IEEE / CVF Computer Vision and Pattern Recognition Conference}}, 2017, pp. 936--944.

\bibitem{SOD1}
N.~Liu and J.~Han, ``{DHSNet}: Deep hierarchical saliency network for salient object detection,'' in \emph{{IEEE / CVF Computer Vision and Pattern Recognition Conference}}, 2016, pp. 678--686.

\bibitem{ViT}
A.~Dosovitskiy, L.~Beyer, A.~Kolesnikov, D.~Weissenborn, X.~Zhai, T.~Unterthiner, M.~Dehghani, M.~Minderer, G.~Heigold, S.~Gelly, J.~Uszkoreit, and N.~Houlsby, ``An image is worth 16x16 words: Transformers for image recognition at scale,'' in \emph{{International Conference on Learning Representations}}, 2021.

\bibitem{swin_v1}
Z.~Liu, Y.~Lin, Y.~Cao, H.~Hu, Y.~Wei, Z.~Zhang, S.~Lin, and B.~Guo, ``Swin transformer: Hierarchical vision transformer using shifted windows,'' in \emph{{IEEE / CVF International Conference on Computer Vision}}, 2021, pp. 9992--10\,002.

\bibitem{PVTv1}
W.~Wang, E.~Xie, X.~Li, D.-P. Fan, K.~Song, D.~Liang, T.~Lu, P.~Luo, and L.~Shao, ``Pyramid vision transformer: A versatile backbone for dense prediction without convolutions,'' in \emph{{IEEE / CVF International Conference on Computer Vision}}, 2021, pp. 548--558.

\bibitem{integrity}
M.~Zhuge, D.-P. Fan, N.~Liu, D.~Zhang, D.~Xu, and L.~Shao, ``Salient object detection via integrity learning,'' \emph{{IEEE Transactions on Pattern Analysis and Machine Intelligence}}, 2022.

\bibitem{SOD_gradient}
L.~Zhang, Y.~Zhang, H.~Yan, Y.~Gao, and W.~Wei, ``Salient object detection in hyperspectral imagery using multi-scale spectral-spatial gradient,'' \emph{Neurocomputing}, vol. 291, pp. 215--225, 2018.

\bibitem{SENet}
J.~Hu, L.~Shen, and G.~Sun, ``Squeeze-and-excitation networks,'' in \emph{{IEEE / CVF Computer Vision and Pattern Recognition Conference}}, 2018, pp. 7132--7141.

\bibitem{CBAM}
S.~Woo, J.~Park, J.-Y. Lee, and I.~S. Kweon, ``Cbam: Convolutional block attention module,'' in \emph{Proceedings of the European conference on computer vision (ECCV)}, 2018, pp. 3--19.

\bibitem{DANet}
J.~Fu, J.~Liu, H.~Tian, Y.~Li, Y.~Bao, Z.~Fang, and H.~Lu, ``Dual attention network for scene segmentation,'' in \emph{{IEEE / CVF Computer Vision and Pattern Recognition Conference}}, 2019, pp. 3146--3154.

\bibitem{SOD_att2}
N.~Liu, J.~Han, and M.-H. Yang, ``{PiCANet}: Learning pixel-wise contextual attention for saliency detection,'' in \emph{{IEEE / CVF Computer Vision and Pattern Recognition Conference}}, 2018, pp. 3089--3098.

\bibitem{SOD_att3}
T.~Zhao and X.~Wu, ``Pyramid feature attention network for saliency detection,'' in \emph{{IEEE / CVF Computer Vision and Pattern Recognition Conference}}, 2019, pp. 3080--3089.

\bibitem{SOD_att4}
S.~Chen, X.~Tan, B.~Wang, and X.~Hu, ``Reverse attention for salient object detection,'' in \emph{{European Conference on Computer Vision}}, 2018, pp. 236--252.

\bibitem{SOD_RGB-D1}
L.~Qu, S.~He, J.~Zhang, J.~Tian, Y.~Tang, and Q.~Yang, ``Rgbd salient object detection via deep fusion,'' \emph{{IEEE Transactions on Image Process.}}, vol.~26, no.~5, pp. 2274--2285, 2017.

\bibitem{SOD_RGB-D2}
K.~Fu, D.-P. Fan, G.-P. Ji, Q.~Zhao, J.~Shen, and C.~Zhu, ``Siamese network for rgb-d salient object detection and beyond,'' \emph{{IEEE Transactions on Pattern Analysis and Machine Intelligence}}, vol.~44, no.~9, pp. 5541--5559, 2021.

\bibitem{SOD_RGB-T1}
W.~Zhou, Q.~Guo, J.~Lei, L.~Yu, and J.-N. Hwang, ``Ecffnet: Effective and consistent feature fusion network for rgb-t salient object detection,'' \emph{{IEEE Transactions on Circuits and Systems for Video Technology}}, vol.~32, no.~3, pp. 1224--1235, 2021.

\bibitem{SOD_RGB-T2}
H.~Bi, R.~Wu, Z.~Liu, J.~Zhang, C.~Zhang, T.-Z. Xiang, and X.~Wang, ``Psnet: Parallel symmetric network for rgb-t salient object detection,'' \emph{Neurocomputing}, vol. 511, pp. 410--425, 2022.

\bibitem{SOD_RGB-LF1}
M.~Zhang, J.~Li, J.~Wei, Y.~Piao, and H.~Lu, ``Memory-oriented decoder for light field salient object detection,'' \emph{{Advances in Neural Information Processing Systems}}, vol.~32, 2019.

\bibitem{SOD_RGB-LF2}
Y.~Piao, Y.~Jiang, M.~Zhang, J.~Wang, and H.~Lu, ``Panet: Patch-aware network for light field salient object detection,'' \emph{IEEE Transactions on Cybernetics}, vol.~53, no.~1, pp. 379--391, 2023.

\bibitem{SOD_semi1}
Y.~Zhou, S.~Huo, W.~Xiang, C.~Hou, and S.-Y. Kung, ``Semi-supervised salient object detection using a linear feedback control system model,'' \emph{IEEE Transactions on Cybernetics}, vol.~49, no.~4, pp. 1173--1185, 2018.

\bibitem{SOD_semi2}
Y.~Lv, B.~Liu, J.~Zhang, Y.~Dai, A.~Li, and T.~Zhang, ``Semi-supervised active salient object detection,'' \emph{{Pattern Recognition}}, vol. 123, p. 108364, 2022.

\bibitem{SOD_self1}
Y.~Wang, X.~Shen, S.~X. Hu, Y.~Yuan, J.~L. Crowley, and D.~Vaufreydaz, ``Self-supervised transformers for unsupervised object discovery using normalized cut,'' in \emph{{IEEE / CVF Computer Vision and Pattern Recognition Conference}}, 2022, pp. 14\,543--14\,553.

\bibitem{SOD_self2}
X.~Zhao, Y.~Pang, L.~Zhang, H.~Lu, and X.~Ruan, ``Self-supervised pretraining for rgb-d salient object detection,'' in \emph{{AAAI Conference on Artificial Intelligence}}, 2022, pp. 3463--3471.

\bibitem{SOD_unsupervised1}
D.~Zhang, J.~Han, and Y.~Zhang, ``Supervision by fusion: Towards unsupervised learning of deep salient object detector,'' in \emph{{IEEE / CVF International Conference on Computer Vision}}, 2017, pp. 4048--4056.

\bibitem{coseg_fss1}
W.~Liu, C.~Zhang, G.~Lin, and F.~Liu, ``{CRNet}: Cross-reference networks for few-shot segmentation,'' in \emph{{IEEE / CVF Computer Vision and Pattern Recognition Conference}}, 2020, pp. 4164--4172.

\bibitem{coseg_fss2}
M.~Siam, N.~Doraiswamy, B.~N. Oreshkin, H.~Yao, and M.~J{\"a}gersand, ``Weakly supervised few-shot object segmentation using co-attention with visual and semantic embeddings,'' in \emph{{International Joint Conference on Artificial Intelligence}}, 2020.

\bibitem{coseg_ss1}
T.-W. Ke, J.-J. Hwang, Y.~Guo, X.~Wang, and S.~X. Yu, ``Unsupervised hierarchical semantic segmentation with multiview cosegmentation and clustering transformers,'' in \emph{{IEEE / CVF Computer Vision and Pattern Recognition Conference}}, 2022, pp. 2571--2581.

\bibitem{coseg_ss2}
H.~Zhang, H.~Zhang, C.~Wang, and J.~Xie, ``Co-occurrent features in semantic segmentation,'' in \emph{{IEEE / CVF Computer Vision and Pattern Recognition Conference}}, 2019, pp. 548--557.

\bibitem{iCoseg}
D.~Batra, A.~Kowdle, D.~Parikh, J.~Luo, and T.~Chen, ``{iCoseg}: Interactive co-segmentation with intelligent scribble guidance,'' in \emph{{IEEE / CVF Computer Vision and Pattern Recognition Conference}}, 2010, pp. 3169--3176.

\bibitem{coseg_rw1}
K.-Y. Chang, T.-L. Liu, and S.-H. Lai, ``From co-saliency to co-segmentation: An efficient and fully unsupervised energy minimization model,'' in \emph{{IEEE / CVF Computer Vision and Pattern Recognition Conference}}, 2011, pp. 2129--2136.

\bibitem{coseg_rw2}
C.~Rother, T.~Minka, A.~Blake, and V.~Kolmogorov, ``Cosegmentation of image pairs by histogram matching-incorporating a global constraint into mrfs,'' in \emph{{IEEE / CVF Computer Vision and Pattern Recognition Conference}}, 2006, pp. 993--1000.

\bibitem{coseg2}
W.~Li, O.~Hosseini~Jafari, and C.~Rother, ``Deep object co-segmentation,'' in \emph{{Asian Conference on Computer Vision}}, 2018, pp. 638--653.

\bibitem{coseg_lstm1}
C.~Zhang, G.~Li, G.~Lin, Q.~Wu, and R.~Yao, ``Cyclesegnet: Object co-segmentation with cycle refinement and region correspondence,'' \emph{{IEEE Transactions on Image Process.}}, vol.~30, pp. 5652--5664, 2021.

\bibitem{coseg_lstm2}
B.~Li, Z.~Sun, Q.~Li, Y.~Wu, and A.~Hu, ``Group-wise deep object co-segmentation with co-attention recurrent neural network,'' in \emph{{IEEE / CVF International Conference on Computer Vision}}, 2019, pp. 8519--8528.

\bibitem{lstm}
S.~Hochreiter and J.~Schmidhuber, ``Long short-term memory,'' \emph{Neural computation}, vol.~9, no.~8, pp. 1735--1780, 1997.

\bibitem{coseg_weakly}
X.~Wang, S.~You, X.~Li, and H.~Ma, ``Weakly-supervised semantic segmentation by iteratively mining common object features,'' in \emph{{IEEE / CVF Computer Vision and Pattern Recognition Conference}}, 2018, pp. 1354--1362.

\bibitem{SLIC}
R.~Achanta, A.~Shaji, K.~Smith, A.~Lucchi, P.~Fua, and S.~S{\"u}sstrunk, ``Slic superpixels compared to state-of-the-art superpixel methods,'' \emph{{IEEE Transactions on Pattern Analysis and Machine Intelligence}}, vol.~34, no.~11, pp. 2274--2282, 2012.

\bibitem{trad_CoSOD_contour}
Z.~Liu, W.~Zou, L.~Li, L.~Shen, and O.~Le~Meur, ``Co-saliency detection based on hierarchical segmentation,'' \emph{IEEE Signal Processing Letters}, vol.~21, no.~1, pp. 88--92, 2014.

\bibitem{MetricCoSOD}
J.~Han, G.~Cheng, Z.~Li, and D.~Zhang, ``A unified metric learning-based framework for co-saliency detection,'' \emph{{IEEE Transactions on Circuits and Systems for Video Technology}}, vol.~28, no.~10, pp. 2473--2483, 2018.

\bibitem{MetricCoSOD1}
D.~Zhang, D.~Meng, and J.~Han, ``Co-saliency detection via a self-paced multiple-instance learning framework,'' \emph{{IEEE Transactions on Pattern Analysis and Machine Intelligence}}, vol.~39, no.~5, pp. 865--878, 2017.

\bibitem{ren2022adaptive}
G.~Ren, T.~Dai, and T.~Stathaki, ``Adaptive intra-group aggregation for co-saliency detection,'' in \emph{{International Conference on Acoustics, Speech, and Signal Processing}}, 2022, pp. 2520--2524.

\bibitem{VGG}
K.~Simonyan and A.~Zisserman, ``Very deep convolutional networks for large-scale image recognition,'' in \emph{{International Conference on Learning Representations}}, 2015.

\bibitem{ResNet}
K.~He, X.~Zhang, S.~Ren, and J.~Sun, ``Deep residual learning for image recognition,'' in \emph{{IEEE / CVF Computer Vision and Pattern Recognition Conference}}, 2016, pp. 770--778.

\bibitem{Inceptionv2v3}
C.~Szegedy, V.~Vanhoucke, S.~Ioffe, J.~Shlens, and Z.~Wojna, ``Rethinking the inception architecture for computer vision,'' in \emph{{IEEE / CVF Computer Vision and Pattern Recognition Conference}}, 2016, pp. 2818--2826.

\bibitem{swin_v2}
Z.~Liu, H.~Hu, Y.~Lin, Z.~Yao, Z.~Xie, Y.~Wei, J.~Ning, Y.~Cao, Z.~Zhang, L.~Dong, F.~Wei, and B.~Guo, ``Swin transformer v2: Scaling up capacity and resolution,'' in \emph{{IEEE / CVF Computer Vision and Pattern Recognition Conference}}, 2021, pp. 11\,999--12\,009.

\bibitem{PVTv2}
W.~\vspace{0mm}Wang, E.~Xie, X.~Li, D.-P. Fan, K.~Song, D.~Liang, T.~Lu, P.~Luo, and L.~Shao, ``Pvtv2: Improved baselines with pyramid vision transformer,'' \emph{Computational Visual Media}, vol.~8, no.~3, pp. 1--10, 2022.

\bibitem{CoSformer}
L.~Tang, ``{CoSformer}: Detecting co-salient object with transformers,'' \emph{arXiv preprint arXiv:2104.14729}, 2021.

\bibitem{FASS}
X.~Zheng, Z.~Zha, and L.~Zhuang, ``A feature-adaptive semi-supervised framework for co-saliency detection,'' in \emph{{ACM International Conference on Multimedia}}, 2018, pp. 959--966.

\bibitem{qian2022co}
X.~Qian, Y.~Zeng, W.~Wang, and Q.~Zhang, ``Co-saliency detection guided by group weakly supervised learning,'' \emph{{IEEE Transactions on Multimedia}}, pp. 1--1, 2022.

\bibitem{GONet}
K.-J. Hsu, C.-C. Tsai, Y.-Y. Lin, X.~Qian, and Y.-Y. Chuang, ``Unsupervised cnn-based co-saliency detection with graphical optimization,'' in \emph{{European Conference on Computer Vision}}, 2018, pp. 502--518.

\bibitem{hsu2018co}
K.-J. Hsu, Y.-Y. Lin, and Y.-Y. Chuang, ``Co-attention cnns for unsupervised object co-segmentation,'' in \emph{{International Joint Conference on Artificial Intelligence}}, 2018, pp. 748--756.

\bibitem{zeng2019joint}
Y.~Zeng, Y.~Zhuge, H.~Lu, and L.~Zhang, ``Joint learning of saliency detection and weakly supervised semantic segmentation,'' in \emph{{IEEE / CVF International Conference on Computer Vision}}, 2019, pp. 7223--7233.

\bibitem{MSRC}
J.~Winn, A.~Criminisi, and T.~Minka, ``Object categorization by learned universal visual dictionary,'' in \emph{{IEEE / CVF International Conference on Computer Vision}}, vol.~2, 2005, pp. 1800--1807.

\bibitem{li2011co}
H.~Li and K.~N. Ngan, ``A co-saliency model of image pairs,'' \emph{{IEEE Transactions on Image Process.}}, vol.~20, no.~12, pp. 3365--3375, 2011.

\bibitem{TripletLoss}
F.~Schroff, D.~Kalenichenko, and J.~Philbin, ``{FaceNet}: A unified embedding for face recognition and clustering,'' in \emph{{IEEE / CVF Computer Vision and Pattern Recognition Conference}}, 2015, pp. 815--823.

\bibitem{DCFM}
S.~Yu, J.~Xiao, B.~Zhang, and E.~G. Lim, ``Democracy does matter: Comprehensive feature mining for co-salient object detection,'' in \emph{{IEEE / CVF Computer Vision and Pattern Recognition Conference}}, 2022.

\bibitem{CoSOD_group_exchange}
Y.~Wu, H.~Song, B.~Liu, K.~Zhang, and D.~Liu, ``Co-salient object detection with uncertainty-aware group exchange-masking,'' in \emph{{IEEE / CVF Computer Vision and Pattern Recognition Conference}}, 2023, pp. 19\,639--19\,648.

\bibitem{ImageNet}
J.~Deng, W.~Dong, R.~Socher, L.-J. Li, K.~Li, and L.~Fei-Fei, ``Imagenet: A large-scale hierarchical image database,'' in \emph{{IEEE / CVF Computer Vision and Pattern Recognition Conference}}, 2009, pp. 248--255.

\bibitem{ImageNet-S}
S.~Gao, Z.-Y. Li, M.-H. Yang, M.-M. Cheng, J.~Han, and P.~Torr, ``Large-scale unsupervised semantic segmentation,'' \emph{{IEEE Transactions on Pattern Analysis and Machine Intelligence}}, vol.~45, no.~6, pp. 7457--7476, 2023.

\bibitem{ICNet}
W.-D. Jin, J.~Xu, M.-M. Cheng, Y.~Zhang, and W.~Guo, ``{ICNet}: Intra-saliency correlation network for co-saliency detection,'' \emph{{Advances in Neural Information Processing Systems}}, pp. 18\,749--18\,759, 2020.

\bibitem{Smeasure}
D.-P. Fan, M.-M. Cheng, Y.~Liu, T.~Li, and A.~Borji, ``{Structure-Measure}: A new way to evaluate foreground maps,'' in \emph{{IEEE / CVF International Conference on Computer Vision}}, 2017, pp. 4558--4567.

\bibitem{Fmeasure}
R.~Achanta, S.~Hemami, F.~Estrada, and S.~Susstrunk, ``Frequency-tuned salient region detection,'' in \emph{{IEEE / CVF Computer Vision and Pattern Recognition Conference}}, 2009, pp. 1597--1604.

\bibitem{Emeasure}
D.-P. Fan, C.~Gong, Y.~Cao, B.~Ren, M.-M. Cheng, and A.~Borji, ``Enhanced-alignment measure for binary foreground map evaluation,'' in \emph{{International Joint Conference on Artificial Intelligence}}, 2018, pp. 698--704.

\bibitem{SOD_review1}
A.~Borji, M.-M. Cheng, H.~Jiang, and J.~Li, ``Salient object detection: A benchmark,'' \emph{{IEEE Transactions on Image Process.}}, vol.~24, pp. 5706--5722, 2015.

\bibitem{Adam}
D.~P. Kingma and J.~Ba, ``Adam: {A} method for stochastic optimization,'' in \emph{{International Conference on Learning Representations}}, 2015.

\bibitem{PyTorch}
A.~Paszke, S.~Gross, F.~Massa, A.~Lerer, J.~Bradbury, G.~Chanan, T.~Killeen, Z.~Lin, N.~Gimelshein, L.~Antiga \emph{et~al.}, ``{PyTorch}: An imperative style, high-performance deep learning library,'' \emph{{Advances in Neural Information Processing Systems}}, vol.~32, 2019.

\bibitem{UFO}
Y.~Su, J.~Deng, R.~Sun, G.~Lin, H.~Su, and Q.~Wu, ``A unified transformer framework for group-based segmentation: Co-segmentation, co-saliency detection and video salient object detection,'' \emph{IEEE Transactions on Multimedia}, pp. 1--13, 2023.

\bibitem{RCAN}
B.~Li, Z.~Sun, L.~Tang, Y.~Sun, and J.~Shi, ``Detecting robust co-saliency with recurrent co-attention neural network,'' in \emph{{International Joint Conference on Artificial Intelligence}}, 2019, pp. 818--825.

\bibitem{GCAGC}
K.~Zhang, T.~Li, S.~Shen, B.~Liu, J.~Chen, and Q.~Liu, ``Adaptive graph convolutional network with attention graph clustering for co-saliency detection,'' in \emph{{IEEE / CVF Computer Vision and Pattern Recognition Conference}}, 2020, pp. 9047--9056.

\bibitem{DeepACG}
K.~Zhang, M.~Dong, B.~Liu, X.-T. Yuan, and Q.~Liu, ``{DeepACG}: Co-saliency detection via semantic-aware contrast gromov-wasserstein distance,'' in \emph{{IEEE / CVF Computer Vision and Pattern Recognition Conference}}, 2021, pp. 13\,698--13\,707.

\bibitem{CSMG}
K.~Zhang, T.~Li, B.~Liu, and Q.~Liu, ``Co-saliency detection via mask-guided fully convolutional networks with multi-scale label smoothing,'' in \emph{{IEEE / CVF Computer Vision and Pattern Recognition Conference}}, 2019, pp. 3090--3099.

\bibitem{Peng2014RGBDSO}
H.~Peng, B.~Li, W.~Xiong, W.~Hu, and R.~Ji, ``Rgbd salient object detection: A benchmark and algorithms,'' in \emph{{European Conference on Computer Vision}}, 2014, pp. 92--109.

\bibitem{SAM}
A.~Kirillov, E.~Mintun, N.~Ravi, H.~Mao, C.~Rolland, L.~Gustafson, T.~Xiao, S.~Whitehead, A.~C. Berg, W.-Y. Lo, P.~Doll{\'a}r, and R.~Girshick, ``Segment anything,'' in \emph{{IEEE / CVF International Conference on Computer Vision}}, 2023.

\bibitem{SAM_medic}
S.~He, R.~Bao, J.~Li, P.~E. Grant, and Y.~Ou, ``Accuracy of segment-anything model (sam) in medical image segmentation tasks,'' \emph{arXiv preprint arXiv:2304.09324}, 2023.

\bibitem{SAM_adapter}
T.~Chen, L.~Zhu, C.~Ding, R.~Cao, S.~Zhang, Y.~Wang, Z.~Li, L.~Sun, P.~Mao, and Y.~Zang, ``Sam fails to segment anything?--sam-adapter: Adapting sam in underperformed scenes: Camouflage, shadow, and more,'' \emph{arXiv preprint arXiv:2304.09148}, 2023.

\bibitem{SAM_3d}
Q.~Shen, X.~Yang, and X.~Wang, ``Anything-3d: Towards single-view anything reconstruction in the wild,'' \emph{arXiv preprint arXiv:2304.10261}, 2023.

\bibitem{CLIP}
A.~Radford, J.~W. Kim, C.~Hallacy, A.~Ramesh, G.~Goh, S.~Agarwal, G.~Sastry, A.~Askell, P.~Mishkin, J.~Clark \emph{et~al.}, ``Learning transferable visual models from natural language supervision,'' in \emph{{International Conference on Machine Learning}}, 2021, pp. 8748--8763.

\bibitem{BLIP}
J.~Li, D.~Li, C.~Xiong, and S.~Hoi, ``Blip: Bootstrapping language-image pre-training for unified vision-language understanding and generation,'' in \emph{International Conference on Machine Learning}.\hskip 1em plus 0.5em minus 0.4em\relax PMLR, 2022, pp. 12\,888--12\,900.

\bibitem{BLIP2}
J.~Li, D.~Li, S.~Savarese, and S.~Hoi, ``Blip-2: Bootstrapping language-image pre-training with frozen image encoders and large language models,'' \emph{{International Conference on Machine Learning}}, 2023.

\bibitem{referring_COD}
X.~Zhang, B.~Yin, Z.~Lin, Q.~Hou, D.-P. Fan, and M.-M. Cheng, ``Referring camouflaged object detection,'' \emph{arXiv preprint arXiv:2306.07532}, 2023.

\end{thebibliography}
